\documentclass[final,3p,times]{elsarticle}
\usepackage{lineno,hyperref}
\modulolinenumbers[5]
\usepackage[utf8]{inputenc}
\usepackage[T1]{fontenc}
\usepackage{verbatim}
\usepackage[figuresright]{rotating}
\usepackage{float}
\usepackage{booktabs}
\usepackage{lineno}
\usepackage{longtable}
\usepackage[namelimits]{amsmath} 
\usepackage{amssymb}             
\usepackage{amsfonts}            
\usepackage{mathrsfs}            
\usepackage{supertabular}        
\usepackage{caption}
\usepackage{epsfig} 
\usepackage{graphicx}
\usepackage{ulem}
\usepackage{threeparttable}
\usepackage{subfigure}
\usepackage{natbib}
\usepackage[numbers]{natbib}
\usepackage{multicol}
\usepackage{multirow}
\usepackage{enumitem}
\usepackage{hyperref}
\hypersetup{ colorlinks=true, linkcolor=blue, anchorcolor=blue, citecolor=blue}
\usepackage{bookmark}
\usepackage{algorithm}
\usepackage{algorithmic}

\usepackage{float}  
\usepackage{lipsum}

\makeatletter
 \newenvironment{breakablealgorithm}
{
	\begin{center}
		\refstepcounter{algorithm}
		\hrule height.8pt depth0pt \kern2pt
		\renewcommand{\caption}[2][\relax]{
			{\raggedright\textbf{\ALG@name~\thealgorithm} ##2\par}%
			\ifx\relax##1\relax 
			\addcontentsline{loa}{algorithm}{\protect\numberline{\thealgorithm}##2}%
			\else 
			\addcontentsline{loa}{algorithm}{\protect\numberline{\thealgorithm}##1}%
			\fi
			\kern2pt\hrule\kern2pt
		}
	}{
		\kern2pt\hrule\relax
	\end{center}
}
 
 \makeatother
 \usepackage[font=small,labelfont=bf,labelsep=none]{caption}  
 \newcommand{\figref}[1]{Fig.~\ref{#1}}
 \newcommand{\tabref}[1]{Table.~\ref{#1}}

\begin{document}

\begin{frontmatter}

\title{A Tent Lévy Flying Sparrow Search Algorithm for Feature Selection: A COVID-19 Case Study \tnoteref{mytitlenote}}

\author[]{Qinwen Yang\fnref{fn1,fn3,fn4}}
\ead{yangqq7160@gmail.com}
\author[]{Yuelin Gao\corref{cor1}\fnref{fn1,fn2,fn3}}
\ead{gaoyuelin@263.net}
\author[]{Yanjie Song\fnref{fn5}}
\ead{songyanjie17@nudt.edu.cn}

\cortext[cor1]{Corresponding author}
\fntext[fn1]{School of Computer Science and Engineering, North Minzu University, Yinchuan, 750021, China}
\fntext[fn2]{School of Mathematics and Information Science, North Minzu University, Yinchuan, 750021, China}
\fntext[fn3]{Ningxia Key Laboratory of Intelligent Information and Big Data Processing, Yinchuan, 750021, China}
\fntext[fn4]{Ningxia Province Cooperative Innovation Center of Scientific Computing and Intelligent in Formation Processing,North Minzu University, Yinchuan, 750021, China}
\fntext[fn5]{College of Systems Engineering, National University of Defense Technology, Changsha, 410073, China}




\begin{abstract}
The "Curse of Dimensionality" induced by the rapid development of information science, might have a negative impact when dealing with big datasets. In this paper, we propose a variant of the sparrow search algorithm (SSA), called Tent Lévy flying sparrow search algorithm (TFSSA), and use it to select the best subset of features in the packing pattern for classification purposes. SSA is a recently proposed algorithm that has not been systematically applied to feature selection problems. After verification by the CEC2020 benchmark function, TFSSA is used to select the best feature combination to maximize classification accuracy and minimize the number of selected features. The proposed TFSSA is compared with nine algorithms in the literature. Nine evaluation metrics are used to properly evaluate and compare the performance of these algorithms on twenty-one datasets from the UCI repository. Furthermore, the approach is applied to the coronavirus disease (COVID-19) dataset, yielding the best average classification accuracy and the average number of feature selections, respectively, of 93.47\% and 2.1. Experimental results confirm the advantages of the proposed algorithm in improving classification accuracy and reducing the number of selected features compared to other wrapper-based algorithms.
\end{abstract}


\end{frontmatter}
\section{Introduction}
An iterative series of task sequences, data pretreatment, data mining, pattern evaluation, and knowledge presentation make up knowledge discovery \cite{too2021hyper, frawley1992knowledge, cios1998data}. The main objective of data preprocessing, as the initial stage in knowledge discovery, is to prepare datasets for use by data mining algorithms \cite{gandomi2012krill}. However, as information science progresses, the dimensionality of datasets increases dramatically, affecting the performance of clustering and classification approaches\cite{garcia2016big, alasadi2017review, mishra2020new}. High-dimensional datasets also have data redundancy, performance deterioration, and a longer period to build models\cite{kamiran2012data, luengo2020big, shen2021two}. The negative is that data analysis is becoming increasingly difficult. To solve this problem, feature selection (FS) is frequently used as a preprocessing approach in the data mining process to determine the best subset of features from all available feature sets\cite{fu2020composite, di2021supervised, kashef2018multilabel}. It tries to eliminate irrelevant and redundant features, simplify clustering and classification, and enhance accuracy \cite{zheng2020full, li2021dual}. 

There are many ways to solve the FS problem, which can generally be divided into the following three categories, namely filters, wrappers, and embedded methods \cite{guyon2003introduction, zhang2020multi}. For the filtering method, the features in a given feature set are first sorted according to a series of criteria. Then the features with a higher ranking are formed into feature subsets \cite{xue2014particle}. Although the obtained feature subset is not necessarily the optimal subset, its calculation  efficiency is very high, so this method is often used for high-dimensional FS problems. Representative filtering methods include minimum redundant F-score criteria \cite{diao2015nature}, maximum correlation (mRMR) \cite{peng2005feature}, Gini index \cite{park2015sequential}, and correlation coefficient \cite{oh2004hybrid}. Wrapper approaches often use a predetermined learning process that is evaluated using a subset of features\cite{du2020joint}. In most circumstances, wrapper methods outperform filter approaches that aren't dependent on any learning mechanism. The embedded methods is embedded in the learning algorithm, and a subset of features can be obtained when the training process of the classification algorithm is completed\cite{zhao2021faster}. The embedded method can solve the problem of excessive redundancy in the results of the filter methods based on feature sorting, and can also solve the problem of excessive time complexity of the wrapper methods, which is a compromise between filter and wrapper\cite{zhao2020cloud, bi2022survey}.

Various methods for discovering optimal feature subsets have arisen in the wake of the wrapper-based method, including heuristic search, complete search, greedy search, and random search, to mention a few\cite{xu2022online, chen2022software, xu2021feature, jain2021feature}. However, the majority of these approaches have local optima and/or are computationally expensive \cite{jin2020deep, emary2016binary}.

Therefore, searching for (close to) optimal subsets from the original set is a challenging problem. Over the past three decades, Evolutionary Algorithms (EA) have been very reliable in solving various optimization problems, such as image processing \cite{abdel2020hsma_woa}, intrusion detection \cite{hosseini2020new}, path planning \cite{pan2020hybrid, wu2021hybrid}, particle filtering \cite{moghaddasi2020hybrid}, production scheduling \cite{diao2015nature}, support vector machines \cite{hamdi2018accurate}, wireless sensors \cite{harizan2020evolutionary}, neural network models\cite{mirjalili2019evolutionary}.

Due to its capabilities in seeking competitive solutions employing tactics that perform well in exploration, EA has recently gained a lot of attention in tackling FS challenges 
\cite{kamath2012evolutionary,abd2014review,jadhav2018information}. These approaches include Genetic Algorithm (GA)  \cite{ghamisi2014feature}, Memetic Algorithm (MA) \cite{kudo2000comparison},  Ant Colony Optimization (ACO) \cite{kashef2015advanced}, Gravitational Search Algorithm (GSA) \cite{han2014feature}, Flower Pollination Algorithm (FPA) \cite{yang2012flower}, Bat Algorithm (BA) \cite{rodrigues2014wrapper}, Differential Evolution (DE) \cite{zorarpaci2016hybrid}, and Particle Swarm Optimization (PSO) \cite{wang2007feature}. A comprehensive review of nature-inspired FS techniques can be found in \cite{tang2014feature}, and a detailed analysis of EA to FS can be found in \cite{xue2015survey}. Here are some examples. 

Based on GA, the K-Nearest-Neighbors approach for diagnosing patient diseases was proposed in \cite{maleki2021k}. It used a hybrid genetic algorithm to perform efficient feature selection. The K-NN algorithm was utilized to diagnose lung cancer after an experimental technique was employed to determine the ideal value of K. Experiments using large cancer databases indicated 100\% accuracy.  In \cite{zhou2021problem}, by minimizing the numerous objectives of the FS, a non-dominated sorting genetic algorithm (NSGA) is employed to solve the multi-objective optimization problem (MOP). Results from 15 real-world high-dimensional datasets show that NSGA can achieve competitive classification accuracy with a reduced selection of features. 
Recently, \cite{xue2021adaptive} proposes a multi-objective binary genetic algorithm called MOBGA-AOS with five crossover operators. The experimental results showed that MOBGA-AOS could remove a large number of features while ensuring a small classification error. Furthermore, it obtains outstanding advantages on large-scale datasets, indicating that MOBGA-AOS is competent to solve high-dimensional FS problems. 

Based on PSO, \cite{song2021feature} proposes a K-NN and mutual information-based bare-bones PSO (BBPSO) feature selection algorithm. The adaptive flip mutation operator is intended to assist particles in breaking out from the local optimal solution. The results of the studies showed that the proposed method may yield a higher performance feature subset.
In \cite{song2021fast}, a high-dimensional FS problem is solved using a multi-stage hybrid FS algorithm (HFS-C-P) based on PSO. The technique produced great feature subsets at the lowest computational cost, according to the results. 
Recently, \cite{li2021improved} propose an improved Sticky Binary PSO (ISBPSO) algorithm for FS. ISBPSO achieves superior accuracy with fewer features in most cases, according to results from 12 UCI datasets. 

Grey Wolf Optimizer (GWO) and SSA are other EA which were also investigated for solving the FS problem. \cite{jangir2018new} proposed non-dominated sorting GWO (NSGWO) is used to perform FS to order to improve the categorization of cervical lesions by reducing the number of textural features while increasing the classification accuracy. 
Recently,  \cite{sathiyabhama2021novel} proposed a FS framework based on GWO and a rough set method called GWORS for finding salient features from extracted mammograms. In \cite{almazini2021grey}, a FS method called the enhanced binary GWO (EBGWO) algorithm for outliers in detecting network intrusions. Based on SSA, \cite{chen2021feature} proposed a spark-based improved sparrow search algorithm (SPISSA) that is used to search feature subsets on intrusion detection datasets. Experiments show that SPISSA can effectively locate the optimal subset while minimizing the algorithm's computational time cost.

In addition, applying FS technology based on EA in detecting COVID-19 patients is also becoming more extensive. \cite{da2020forecasting} combining single models such as Stereo Regression, Quantile Random Forest, k-Nearest Neighbors, Bayesian Regression Neural Network, and Support Vector Regression with Variational Mode Decomposition (VMD) to create a hybrid model to forecast COVID-19 cases in Brazil and the United States. The VMD-based model proved to be one of the most effective strategies for forecasting COVID-19 cases five days in advance. 
\cite{dey2021mrfgro} presented a hybrid model. To begin, extract several characteristics from the COVID-19-affected lungs. The Manta-Ray Foraging-based Golden Ratio Optimizer (MRFGRO), a hybrid meta-heuristic FS technique, is then presented to pick the most critical subset of characteristics. Although the findings show that the proposed strategy is quite effective, the model is only tested on the CT scan dataset.
\cite{shaban2021accurate} proposes Distance Biased Naive Bayes (DBNB), a new approach for detecting COVID-19 infected patients. Through a novel FS technique called Advanced Particle Swarm Optimization, DBNB picks the most informative characteristics for identifying COVID-19 patients (APSO). APSO is a hybrid strategy that uses filter and wrapper approaches to offer accurate but crucial classification features. Because it introduces the highest accuracy rating with minimal time loss, DBNB surpasses top-of-the-line COVID-19 diagnostic techniques, according to experimental results. 


In conclusion, in the realm of FS, numerous EA have been used. Nonetheless, the No Free Lunch (NFL) \cite{adam2019no} theorem states that existing procedures can always be improved. The SSA was recently created by Jiankai Xue for global optimization, and there is currently insufficient research in the literature to solve the FS problem using the sparrow search algorithm, motivating us to suggest a variant of SSA for FS in the \autoref{sec:The proposed algorithm}. 
The Sparrow Search Algorithm (SSA) is a new and well-organized EA that can be used in different areas for solving optimization problems, such as brain tumor detection \cite{liu2021optimal}, parameter identification \cite{zhu2021optimal}, configuration network \cite{zhang2021stochastic}, and fault diagnosis of wind turbines \cite{tuerxun2021fault}. However, SSA still has the problem of easy to fall into the local optimum and so far the application of the SSA for FS is very scarce\cite{gad2022improved}. Motivated by the above analysis, we propose a Tent Lévy Flying Sparrow Search Algorithm (TFSSA) in this paper to increase the capability of SSA in confronting FS challenges, where the main contributions are summarized below.
\begin{itemize}
\item A TFSSA is proposed for feature selection problems, and it is utilized to solve a COVID-19 case study.
\item An improved Tent chaos strategy, LFs mechanism, and Self-adaptive hyper-parameters are integrated into TFSSA to improve SSA's exploratory behavior and perform well in the CEC2020 benchmark function.
\item A comprehensive comparison of TFSSA and nine different algorithms for feature selection problems is undertaken in nine aspects.
\item The proposed TFSSA's improved searching capabilities are tested on 21 well-known feature selection datasets, with excellent results.
\end{itemize}
The remainder of this paper is organized as follows: 
\autoref{SSA} presents the background of the sparrow search algorithm (SSA). The proposed tent lévy flying sparrow search algorithm (TFSSA) is described in \autoref{sec:The proposed algorithm}. \autoref{sec:TFSSA for FS} presents the proposed TFSSA algorithms for FS, while the experimental results with discussions are reported in \autoref{sec:Experimental}. \autoref{Realworld} demonstrates the adoption of the proposed TFSSA in a COVID-19 application. Finally, conclusions is stated in \autoref{Conclusions}. 
\section{Classical sparrow search algorithm (SSA)}\label{SSA} 
The academic performance of animal and insect populations provides a fascinating field of research for different researchers. Researchers can draw inspiration from the collective actions of animal communities or insect societies to plan algorithms or distributed problem-solving mechanisms. In 2020, Xue Jiankai proposed the SSA to enhance optimization technology and decode the complexity involved in the process.
\subsection{Rule design}
The classical SSA is primarily motivated by sparrow population foraging behavior. It's a search algorithm with high optimization and efficiency capabilities \cite{xue2020novel}. For simplicity, the biology of sparrow populations during foraging is idealized and normalized as the following behaviors.
\begin{enumerate}
    \item [(1)] Producers (leaders) have access to plentiful food sources and are responsible for ensuring that all scroungers (followers) have access to foraging sites. 
    \item [(2)] Some sparrows will be chosen as patrollers. When patrollers come across a predator, they will sound an alarm. When the safety threshold is exceeded, the producer must direct the scroungers (followers) to other safe regions. 
    \item [(3)] Sparrows that can discover a better food source earn more energy and are promoted to producers (leaders), while hungry scroungers (followers) are more likely to fly elsewhere to forage in the hopes of gaining more energy, and the producer-to-forager ratio remains steady. 
    \item [(4)] Scroungers (followers) hunt for food after the finest producers (leaders). Simultaneously, certain predators may observe producers (leaders) and steal food. 
    \item [(5)] When threatened, sparrows near the flock's edge moved swiftly to a safe region, while sparrows in the center of the flock moved randomly to approach other sparrows in the safe area. 
\end{enumerate}
\subsection{Algorithm design}
The following matrix can be used to depict the position of sparrows:

\begin{equation}
X=\left[\begin{array}{ccccccc}
x_{11} & x_{12} & x_{13} & \ldots & x_{1 j} & \ldots & x_{1 D} \\
x_{21} & x_{22} & x_{23} & \ldots & x_{2 j} & \ldots & x_{2 D} \\
x_{31} & x_{32} & x_{33} & \ldots & x_{3 j} & \ldots & x_{3 D} \\
\vdots & \vdots & \vdots & \ddots & \vdots & \ddots & \vdots \\
x_{i 1} & x_{i 2} & x_{i 3} & \ldots & x_{i j} & \ldots & x_{i D} \\
\vdots & \vdots & \vdots & \ddots & \vdots & \ddots & \vdots \\
x_{N 1} & x_{N 2} & x_{N 3} & \ldots & x_{N j} & \ldots & x_{N D}\label{eq1}
\end{array}\right],
\end{equation}
where $N$ is the number of sparrows and $D$ is the dimension of the variables to optimize. Then, the fitness values of all sparrows are represented by vectors as follows:
\begin{equation}
F_{X}=\left[\begin{array}{ccccccc}
f\left(\left[x_{11}\right.\right. & x_{12} & x_{13} & \ldots & x_{1 j} & \ldots &\left.\left. x_{1 D}\right]\right) \\
f\left(\left[x_{21}\right.\right. & x_{22} & x_{23} & \ldots & x_{2 j} & \ldots &\left.\left. x_{2 D}\right]\right) \\
f\left(\left[x_{31}\right.\right. & x_{32} & x_{33} & \ldots & x_{3 j} & \ldots &\left.\left. x_{3 D}\right]\right) \\
\vdots & \vdots & \vdots & \ddots & \vdots & \ddots & \vdots \\
f\left(\left[x_{i1}\right.\right. & x_{i2} & x_{i3} & \ldots & x_{i j} & \ldots &\left.\left. x_{i D}\right]\right) \\
\vdots & \vdots & \vdots & \ddots & \vdots & \ddots & \vdots \\
f\left(\left[x_{N1}\right.\right. & x_{N2} & x_{N3} & \ldots & x_{N j} & \ldots &\left.\left. x_{N D}\right]\right) \\\label{eq2}
\end{array}\right],
\end{equation}
where the value of each row in $F_{X}$ denotes the individual's fitness. The location of the producer is updated according to the rules $(1)$ and $(2)$ in each iteration:
\begin{equation}
X_{i}^{t+1}=\begin{cases}
X_{i}^{t} \cdot \exp (-\frac{i}{\lambda  \cdot T\_max}),& \text { if } R_{2}<S T \\
X_{i}^{t}+L \cdot Q,& \text { if } R_{2} \geq S T\end{cases}\label{eq3},
\end{equation}
where \textit{T\_max} is the maximum number of iterations, $t$ indicates the current iteration, $X_{i}^{t}$ represents the value of the $i$ th sparrow at iteration $t$. $S T \in[0.5,1]$ represent the safety threshold and $ R_{2} \in[0,1]$ represent the warning value. $\lambda \in(0,1]$ is a random number. $L$ shows a vector of $1 \cdot D$. $Q$ is a random number and $Q \sim N\left(0, 1\right)$.


According to the rules (3)$\sim$(4), some followers keep a closer eye on leaders(producers). When the followers spots the producers who have located food, they will promptly leave their current place to collect the food. The scrounger's position update formula is as follows if they don't grab it: 
\begin{equation}
X_{i}^{t+1}=\begin{cases}
Q \cdot \exp \left(\frac{X_{\text{worst }}-X_{i}^{t}}{i^{2}}\right),& \text { if } i>N / 2 \\
X_{P}^{t+1}+\left|X_{i}-X_{P}^{t+1}\right| \cdot L \cdot (A^{T})^ {2 }\cdot A,& \text { otherwise }\label{eq4}
\end{cases},
\end{equation}
where $A$ represents a vector of $1\cdot D$, where each element is randomly assigned $\pm1$. $X_{\text {worst }}$ represents the current global worst position. $X_{P}$ is the current optimal position of the producer. $A$ represents a vector of $1\cdot D$, where each element is randomly assigned $\pm1$. When $i>N / 2$, it indicates that the $i$th scroungers(followers) with poor fitness value is most likely to starve.

We hypothesized that these patrollers made up one-tenth to one-fifth of the population in our simulations. These sparrows' initial placements are generated at random. The mathematical model is expressed as follows according to rule (5):
\begin{equation}
X_{i}^{t+1}=\begin{cases}
X_{b e s t}^{t}+\beta \cdot\left|X_{i}^{t}-X_{b e s t}^{t}\right|,& \text { if } f_{i}>f_{g} \\
X_{i}^{t}+K \cdot\left(\frac{\mid X_{i}^{t}-X_{w o r s t}^{t}}{\left(f_{i}-f_{w}\right)+\varepsilon}\right),& \text { if } f_{i}=f_{g}\end{cases}\label{eq5},
\end{equation}
where $X_{best}$ represents the current global optimal position. $\beta$ is a random number that obeys the standard Gaussian distribution. $\varepsilon$ is the smallest constant to avoid division by zero errors. $K \in[-1,1]$ is a random number. The current fitness value of the sparrow is $f_{i}$. The current global best and worst fitness values are $f_{g}$ and $f_{w}$, respectively. 

For simplicity, when $f_{i}>f_{g}$ represents the sparrow at the edge of the group. $X_{best}$ represents the center of the population location, around which it is safe. $f_{i}=f_{g}$ indicates that sparrows in the middle of the population need to approach other sparrows because they are aware of the danger. $K$ is the step size control coefficient, which indicates the direction in which the sparrow moves. Algorithm \ref{A1} demonstrates the algorithmic structure of the classic SSA.
\begin{breakablealgorithm}
\caption{Sparrow Search Algorithm} \label{A1} 

\begin{algorithmic}[1] 
\begin{footnotesize} 
\REQUIRE ~~\\ 
$N$: the amount of sparrows \\
$PD$: the amount of Producers(leaders) \\
$SD$: the amount of Patrollers \\
$R_{2}$: the warning value \\
\textit{T\_max}: the maximum iterations \\
\ENSURE ~~\\ 
$X_{best}$: the best position\\
$f_{g}$: the best solution\\
\STATE $t$ $\gets$ 0;\\
\WHILE {($t$ $<$ \textit{T\_max})} 
\STATE Calculate the $F_{X}$, $f_{g}$ and $f_{w}$.\\
Update the $R_{2}$.\\
\FOR {each leaders $i \in [1,PD]$}
\STATE The location of leaders(producers) is updated using Eq.(\ref{eq3});
\ENDFOR
\FOR {each followers $i \in [$PD$+1,$N$]$}
\STATE The location of followers(scroungers) is updated using Eq.(\ref{eq4});
\ENDFOR
\FOR {each patrollers $i \in [1,SD]$} 
\STATE The location of patrollers is updated using Eq.(\ref{eq5});
\ENDFOR
\STATE Find the current new location $X_{i}^{t+1}$; 
\STATE If the new location is better than before, update it; 
\STATE Rank the $F_{X}$;
\STATE $t$ $\gets$ $t$ + 1; 
\ENDWHILE
\RETURN $X_{best}$, $f_{g}$. 
\end{footnotesize}
\end{algorithmic}
\end{breakablealgorithm}
\section{The proposed algorithm}\label{sec:The proposed algorithm}
\subsection{Motivations}
The NFL theorem shows that no learning algorithm can consistently produce the most accurate learner in any field. No matter what learning algorithm is used, there is at least one objective function, making the random guess algorithm a better algorithm. Although the SSA has the advantages of faster convergence and stronger optimization-seeking abilities, the original SSA, like other traditional EA, is more subject to the population's poor diversity and falls into a local optimum. The placements of the sparrows in the solution space are randomly distributed, and a random walk method \cite{zhu2021optimal} is used when no nearby sparrows are surrounding the current individual. For a limited number of iterations, this mode delays the convergence trend and reduces convergence accuracy. We propose a tent lévy flying sparrow search algorithm (TFSSA) to improve the complete optimization performance of SSA and to address these shortcomings. 

In subsection 3.2, the improved Tent chaotic map is utilized to initialize the population to boost the diversity of the starting population in SSA. In order to improve the global optimization ability, prevent the algorithm from falling into local optimization, and improve the population diversity in the middle and late stages of the algorithm, the LFs mechanism is introduced in Section 3.3. In order to expand the search scope of producers and improve the global search capability. The adaptive control factor is introduced in Section 3.4. An adaptive updating formula for the number of patrollers is proposed to improve the algorithm's performance by controlling the number of patrollers. When all sparrows find the optimal solution, the enhanced tent chaos and roulette strategy are used to improve the global convergence accuracy further to mutate the optimal sparrow individuals in subsection 3.5.

\subsection{Initialized population by \texorpdfstring{$\psi$-T} Tent chaos map}
Initialization is a severe step in the meta-heuristic algorithm and seriously affects the convergence speed and solution accuracy. The primary motivation of the most advanced initialization methods is to cover the search space as evenly as possible based on generating a small initial population. However, these methods are affected by the dimension disaster, high computational cost, and sensitivity to parameters, which ultimately reduce the convergence speed of the algorithm.

The efficiency of EA is greatly affected by chaotic mapping, which has the advantages of uniform ergodicity, sensitivity to initial values, and fast search speed. Using the randomicity, ergodicity, and regularity of chaotic variables to solve optimization problems can make the algorithm jump out of local optimization, maintain population diversity, and improve the global search ability to a certain extent. However, different chaotic maps significantly impact the chaotic optimization process. Various scholars have introduced chaos mapping and chaos search into EA, trying to improve the problem of falling into local optimums in the latter period and improve the convergence speed and accuracy of the algorithm. The chaotic map used more in the literature is the Logistic map. Still, its value probability is high in the two ranges [0,0.1] and [0.9,1], and the optimization speed is affected by the uneven Logistic traversal, so the algorithm's efficiency will be significantly reduced. Many papers have pointed out that the tent map has better ergodic uniformity and faster convergence speed than the logistic map and have further proved that the tent map can be used as a chaotic sequence to generate optimization algorithms through strict mathematical reasoning.The Tent mapping expression is shown Eq. (\ref{eq61}).
\begin{equation}
	x_{i+1}= \begin{cases}\frac{x_{i}}{a} ,& 0 \leq x \leq 1/2 \\  \frac{1-x_{i}}{1-a}, & 1/2 <x \leq 1\end{cases}\label{eq61}.
\end{equation}

Eq. (\ref{eq61}) after Bernoulli shift transformation is as follows:

\begin{equation}
	x_{i+1}=\left(2 x_i\right) \bmod 1.
\end{equation}

Tent mapping has the advantages of randomness, consistency, and orderliness, and it is a standard method for scholars to find the optimal solution\cite{kuang2014artificial}.On the other hand, a chaotic tent map has flaws such as a short period and unstable period points\cite{shan2005chaotic}. Therefore, to avoid falling into a small period or unstable periodic point, the Tent chaos map is improved by the $\psi$ as shown in Eq. (\ref{eq6}) \cite{zhang2020gravitational}.
\begin{equation}
x_{i+1}= \begin{cases}\frac{x_{i}}{a}+\psi & 0 \leq x \leq a \\  \frac{1-x_{i}}{1-a}+\psi & a<x \leq 1\end{cases}\label{eq6},
\end{equation}
where  $a=0.7$ in current experiments, $\psi=rand(0,1)$ $\times$ $1/N$, $N$ is the population number of sparrows. Eq. (\ref{eq6}) after Bernoulli shift transformation is as follows:
\begin{equation}
	x_{i+1}=\left(2 x_i\right) \bmod 1 +\psi  \label{eq611}.
\end{equation}

Therefore, in TFSSA, the Eq.(\ref{eq1}) and Eq.(\ref{eq611}) in SSA is replaced by the Eqs. (\ref{eq6}) to increase the population diversity. At this time, the improved Tent chaotic sequence is introduced based on the original SSA to initialize the sparrow population. Although the algorithm not only retains the randomness of the initial individuals, but also improves the population diversity at the initial stage, it still cannot guarantee that the population diversity will still have the same degree later. On the contrary, in the algorithm's later stages, the population's diversity is not well guaranteed, and the scavengers constantly hop around the producers, making the algorithm fall into local optimization to a large extent. In this case, the algorithm performance needs to be further improved. We consider introducing the LFs mechanism.

\subsection{ LFs mechanism}
It can be seen from the SSA rule design (3) that when the producer's food does not have enough temptation, hungry scavengers may fly to other places to look for food. However, according to the SSA rule design (4), scavengers mainly search for food in accordance with the producer and go elsewhere to look for food. Generally, they only search for food within a relatively close range of the producer. Therefore, most sparrows may only move around areas with poor solution quality. On the other hand, for each iteration, the individual sparrow will move indiscriminately to the sparrow (producer) whose food is better than his own. This situation increases the algorithm's complexity and leads to low convergence accuracy and a higher possibility of falling into the optimal local solution. Random numbers obeying the Lévy distribution have the characteristics of short-distance walking and long-distance jumping, which will significantly improve the disadvantage of hungry sparrows (scavengers) that only search for food within a relatively close range of producers.

The French mathematician Paul·Lévy (1886–1971) first proposed Lévy Flights (LFs) in 1937. LFs, try to strengthen the optimization process with diversity and universality, which helps the algorithm find the search location effectively and avoid local minima. Therefore, LFs is embedded in the SSA  mechanism to improve the overall optimization efficiency. The foraging activities of most animals also have the characteristics of Lévy flight. For example, most foraging time is spent around known food sources, and sometimes a long flight is required to find other food sources.

Therefore, this part combines the LFs strategy and the inertia weight factor into the classic SSA to improve its ability to expand the search scope and avoid local optimization. In this way, TFSSA can locate the optimal global solution more effectively. The following equation describes this mechanism. Eq. (\ref{eq10}) can be used to express the Lévy distribution \cite{yang2010firefly}.
\begin{equation}
L\acute{e}vy\left ({\alpha }\right)\sim \mu=e^{-1-\alpha },0 < \alpha \le 2 \label{eq10},
\end{equation}
where $\alpha$ is a stability index, and $\alpha =1.5$, the $\mu$ is a Gaussian distribution. The inertia weighting factor $\sigma$ is expressed by Eq. (\ref{eq13}).
\begin{equation}
\sigma =1-t/T\_{}max \label{eq13},
\end{equation}

Then, the sparrow's position $ x_{iD}^{t}$ is mutated by the random roulette strategy in Eq. (\ref{eq14}). 

If rand $>$ $\sigma$,
\begin{equation}
 x_{iD}^{t'}= x_{iD}^{t}+L\left ({\alpha }\right)\cdot \left ({x_{iD}^{t}-x_{best}^{t} }\right)\label{eq14}.
\end{equation}

Else the $x_{iD}^{best'}$ is also changed by Eq. (\ref{eq15}).
\begin{equation}
 x_{iD}^{best'}=x_{iD}^{best}\cdot \left ({1+L\left ({\alpha }\right)}\right)\label{eq15},
\end{equation}
where \textit{L($\alpha$)} is a randomly distributed number drawn from Lévy distribution.This part mainly combines the LFs strategy with classic SSA and uses LFs characteristics to improve its ability to expand the search scope and avoid local optimization. LFs can optimize the diversity of search agents, enabling the algorithm to explore search locations and avoid local minima effectively. The combination of LFs and the SSA algorithm improves the population diversity to a certain extent and enhances the robustness and global optimization capability of the SSA algorithm. However, in many experiments, it is found that the occasional long-distance jump of LFs has no significant impact on the final performance of the algorithm as expected. Because its performance in the CEC2020 benchmark function is not very strong, it still has some effect. This paper continues to consider further improving the algorithm from the manufacturer's location formula. We use super adaptive parameters in the next section to update the producer location and improve global search capability.
\subsection{Self-adaptive hyper-parameters}
In the rule design of SSA, SSA divides the sparrow population into producers (leaders) and scavengers (followers). Producers need more search space to find food sources, while scavengers mainly follow producers to find food. Therefore, the optimization ability of SSA is highly related to the search scope of producers.

In the Eq.(\ref{eq3}), when $R_ {2} <ST$means that there are no predators at present, and the producer (leader) opens the wide area search mode.In this mode, the location update of producers (leaders) is mainly affected by $\exp (-\frac{i}{\alpha  \cdot max\_iteration})$. When $\alpha$ in the Eq.(\ref{eq3}) gets a large random value, the value of $\exp (-\frac{i}{\alpha  \cdot max\_iteration})$ will gradually decrease from (0,1) to (0,0.4) as $i$ becomes larger. Therefore, we introduce adaptive control factors, such as Eq.(\ref{eqzi}), to expand the search scope of producers to Eq.(\ref{eqz}).

\begin{equation}
	w=w_0 \times c^t \label{eqzi},
\end{equation}
where $w_ 0=1 $ is the initial weight, $c $ is the adaptive factor of $w $, which can be set according to the actual problem, and $t $ is the current iteration number. Among them, $c $ is set to 0.9 to keep $w $ at a low value, so as to expand the search scope of producers and improve the global search capability.

\begin{equation}
	X_{i}^{t+1}=\begin{cases}
		X_{i}^{t} \cdot \exp (-\frac{i}{w\cdot \alpha  \cdot max\_iteration}),& \text { if } R_{2}<S T \\
		X_{i}^{t}+Q \cdot L,& \text { if } R_{2} \geq S T\end{cases}\label{eqz}.
\end{equation}

In addition, according to SSA's rule design (2), to avoid predators in the process of foraging, 10\%–20\% of sparrows are selected as patrollers. The presence of patrolmen can help the sparrow population obtain better SSA solutions. When the number of patrollers(PN) is large, improving the sparrow's global optimization ability is beneficial. However, when the number of patrolmen is small, accelerating the SSA algorithm's convergence is helpful. Therefore, as shown in Eq.(\ref{eqp}), an adaptive updating formula for the number of patrolmen is proposed, which can be reduced nonlinearly in the iterative process.
\begin{equation}
	P N=P N_{\max }- \textit{Round} \left[\left(P N_{\max }-P N_{\min }\right) \times \frac{t}{max\_iteration}\right]\label{eqp},
\end{equation}
where $P N_ { max} $ is the maximum number of Patrollers, $ PN_ { min} $ represents the minimum number of Patrollers, the $round$ function is used to round numbers, $t$ represents the current iteration $max\_ iteration$ represents the maximum number of iterations. When all sparrows find the optimal solution, we still cannot guarantee that the improved strategy has significantly improved the local optimal, so we consider further mutation of the optimal sparrow individual.
\subsection{Optimal individual mutation by \texorpdfstring{$\psi$-T}Tent chaos}
The position of the optimal sparrow is mutated using enhanced tent chaos and roulette strategy to improve global convergence accuracy even more \cite{cao2020adaptive}. Thus, in TFSSA, the optimal sparrow individual was changed suddenly by Eqs. (\ref{eq8}) and (\ref{eq9}).
\begin{equation}
r=\frac{e^{2 \cdot\left(1-k / \operatorname{\textit{T\_max})}\right.}-e^{-2 \cdot\left(1-k / \operatorname{\textit{T\_max})}\right.}}{e^{2 \cdot\left(1-k / \operatorname{\textit{T\_max}}\right)}+e^{-2 \cdot\left(1-k / \operatorname{\textit{T\_max}}\right)}}\label{eq8}.
\end{equation}

If $rand$ $<$ $r$, the optimal sparrow position $x_{i D}^{b e s t^{\prime}}$ was mutated by Eq. (\ref{eq9}).
\begin{equation}
x_{i D}^{b e s t^{\prime}}=x_{i D}^{b e s t} \cdot\left(1+\text {\textit{ $\psi$-Tent} }\left(x_{i D}^{b e s t}\right)\right)\label{eq9},
\end{equation}
where \textit{$\psi$-Tent} $\left(x_{i D}^{\text {best }}\right)$ can be calculated by Eq. (\ref{eq6}). The basic steps of TFSSA can be summarized in Algorithm \ref{A2}. 

\begin{figure}[htbp]
\centering
\begin{minipage}[t]{0.48\textwidth}
\centering
\includegraphics[width=7.5cm]{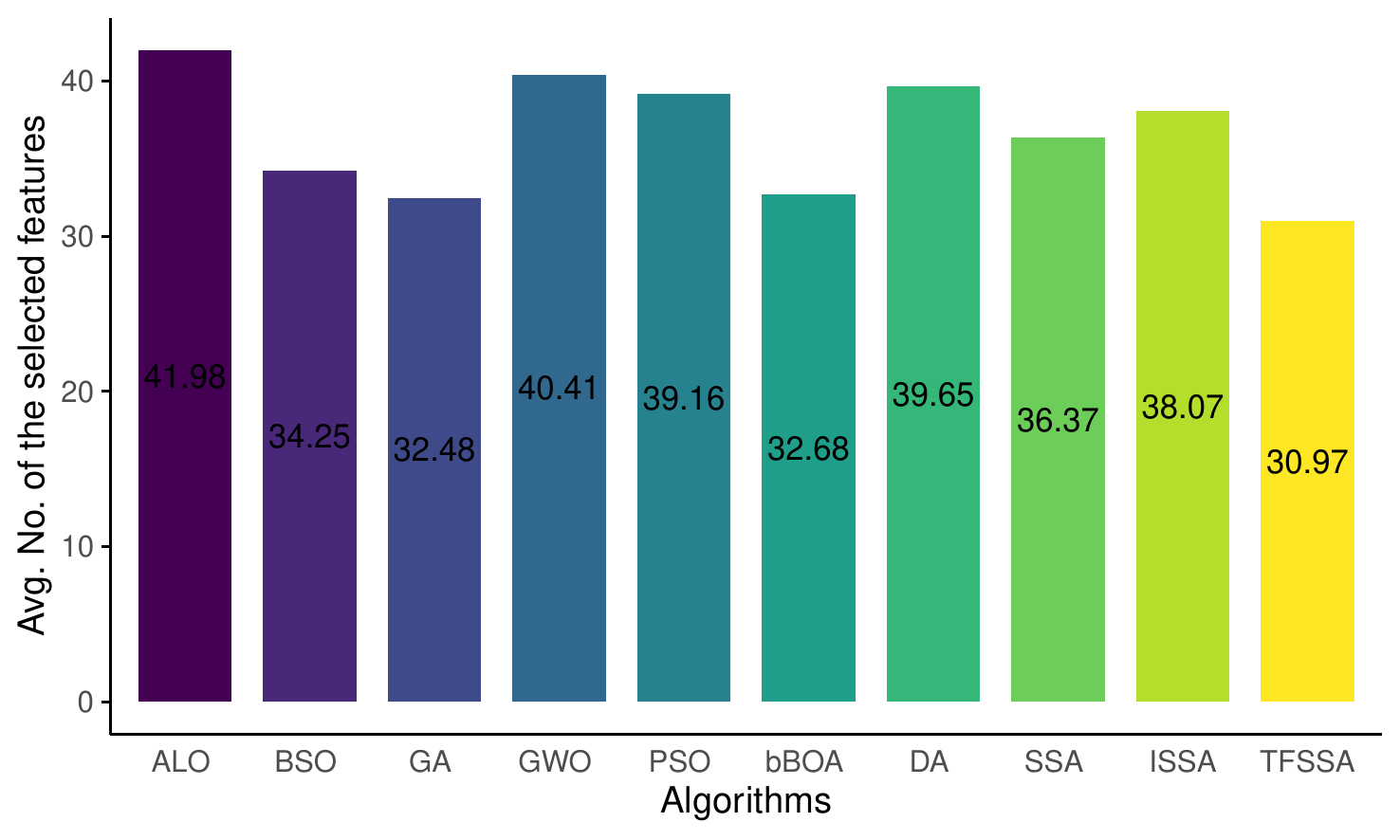}
\end{minipage}
\begin{minipage}[t]{0.48\textwidth}
\centering
\includegraphics[width=7.5cm]{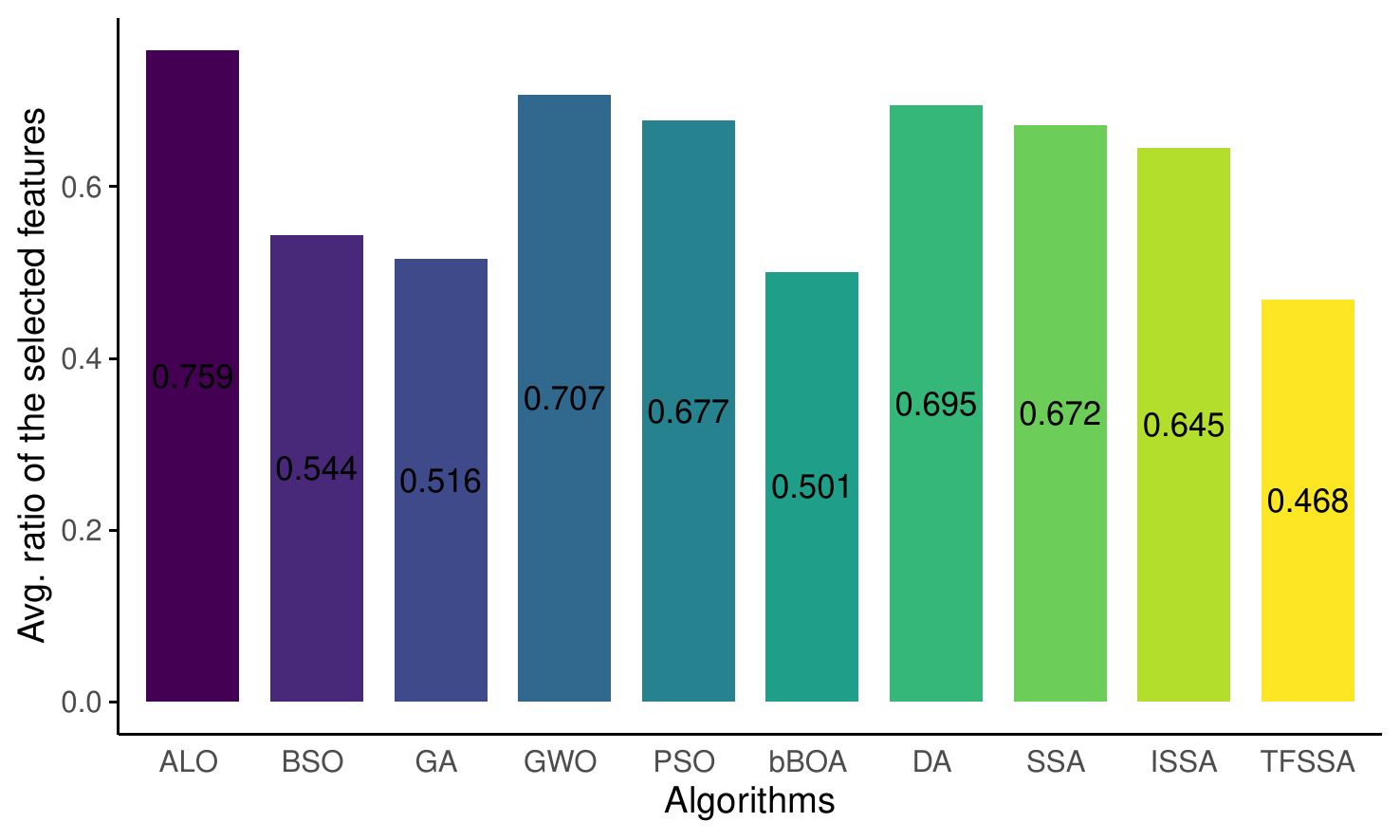}
\end{minipage}
	\caption{Comparison among algorithms total average number of features and the selected feature ratio.}
	\label{fig:feat & ratio}
\end{figure}

\section{TFSSA applied for FS}\label{sec:TFSSA for FS}
\subsection{Initialization}
\begin{figure}
	\centering
	\includegraphics[width=0.7\linewidth]{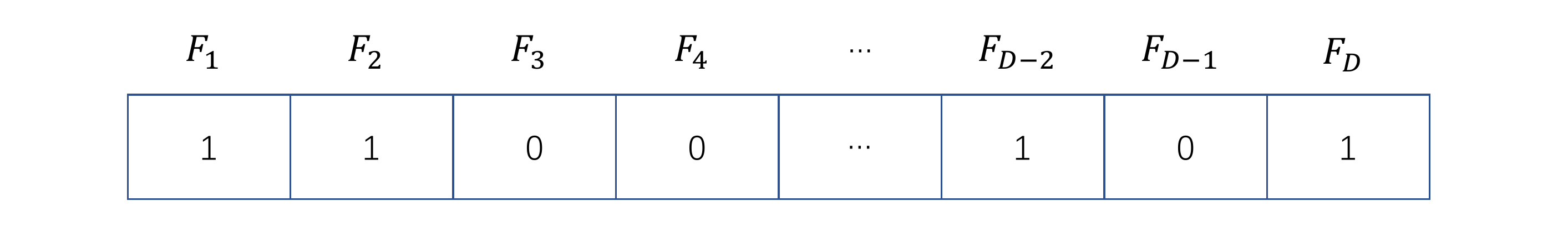}
	\caption{Solution representation}
	\label{fig: initialization}
\end{figure}

The initialization stage is the first step of the population-based algorithm, in which, according to the Eq.(\ref{eq6}) and the Eq.(\ref{eq611}), a sparrow individual of $N$ is generated through the chaotic initialization of  \texorpdfstring {$ \psi$-T} Tent. Each candidate solution $i$ in this study is bound by upper and lower limits, with a range of [-1,1] to enable individuals to conduct more extensive and strict relative searches in the continuous search space domain. Each search agent represents a potential solution with dimension $D$, which in the feature selection example equals the original number of features in any dataset. The FS problem for classification can be stated as picking the least relevant feature subset to maximize average classification accuracy. Therefore, in this study, we try to identify the significant $1$ value and reject the other feature $0$ value. Before starting the fitness evaluation process, according to Eq.(\ref{eq6}) and Eq.(\ref{eq611}) and figure \ref{fig: initialization}, first discretize the initial position of each sparrow in the group to the position on each dimension, that is, $0$ (not selected) or $1$ (selected), and convert it to a random binary value (between 0 and 1).
\subsection{Fitness evaluation}
In this part, the TFSSA is exploited in FS for classification problems. The different feature combinations for a feature vector of size $\eta$ would be $2^{\eta}$, which is a massive space of features to be searched thoroughly. As a result, TFSSA is utilized to choose the optimal feature subset's feature space. Eq. (\ref{eq16}) shows the fitness function utilized in TFSSA to evaluate individual sparrow placements.

\begin{equation}
\text { $Fitness$ }=\lambda E_{R}(D)+\mu \frac{|S|}{|T|} \label{eq16},
\end{equation}
where $E_{R}(D)$ is the error rate for the classifier of condition attribute set, $|S|/|T|$ denotes the ratio of chosen features to total features, 
$\lambda \in [0, 1]$ and $\mu$ = 1-$\lambda$.
%

K-Nearest Neighbors (K-NN) \cite{altman1992introduction} is a popular classification method that may be used to evaluate fitness functions as a simple candidate classifier. The smallest distance between the query instance and the training examples determines the K-NN classifier.
A crucial characteristic of wrapper techniques in FS is the use of the classifier as a guide to the FS activity. The following three primary items can be used to classify wrapper-based feature selection:
\begin{itemize}
    \item [(1)] Method of classification.
    \item [(2)] Criteria for evaluating features.
    \item [(3)] Search method. 
\end{itemize}
%

As demonstrated in Eq. (\ref{eq16}), TFSSA is employed as a search strategy that may adaptively explore the feature space to maximize the feature evaluation criterion. Sparrow's location in the search space reflects one feature combination or solution since each dimension represents a different feature combination or solution.
\subsection{Position updating}
The position of sparrow in TFSSA is updated according to the Eq.(\ref{eqz}), Eq.(\ref{eq4}) and Eq.( \ref{eq5}). If the current sparrow is a producer, update its position with the Eq. (\ref{eqz}); Otherwise, apply the Eq.(\ref{eq4}) to update the location of the scavenger. At the same time, Eq.(\ref{eq5}) will be used to update the Patrollers. It should be noted that the continuous value of the position vector is preserved after each iteration for future use in the continuous position update of the entire continuous iteration. These values are discretized to evaluate the fitness value of the generated binary solution according to the classification error rate obtained by the involved classifier using the features selected by TFSSA. Next, the process iterates until the stop criterion is met, that is, the maximum number of iterations in this study (which has been proved to be sufficient to quantify the quality of the TFSSA). Finally, the basic steps of TFSSA can be summarized in Algorithm \ref{A2}. 

\begin{breakablealgorithm}
\caption{Tent Lévy Flying Sparrow Search Algorithm} \label{A2} 
\begin{algorithmic}[1] 
\begin{footnotesize} 
\REQUIRE ~~\\ 
$N$: the amount of sparrows \\
$PD$: the amount of Producers(Leaders) \\
$SD$: the amount of Patrollers \\
$ST$: the safety threshold\\
$R_{2}$: the warning  value \\
\textit{T\_max}: the maximum iterations \\
Initialize a flock of sparrows' location $X$  // Pretreatment using $\psi$-Tent chaos map by Eq. (\ref{eq6}) and Eq. (\ref{eq611}).\\
\ENSURE ~~\\ 
$X_{best}$: the best position so far\\
$f_{g}$: the best solution so far\\
\STATE $t$ $\gets$ 0;\\
\WHILE {($t$ $<$ \textit{T\_max})} 
\STATE Rank the fitness vaule $F_{X}$ using Eq.(\ref{eq2}); \\
Find the $f_{g}$ and $f_{w}$;\\
Update the $R_{2}$ $\gets$ a random value in [0,1], and calculate the $\sigma$ using Eq. (\ref{eq8}).\\
\FOR {each leaders $i \in [1,PD]$}
\STATE The location of leaders(producers) is updated using Eq.(\ref{eqz});
\ENDFOR
\FOR {each followers $i \in [$PD$+1,$N$]$}
\STATE The location of followers(scroungers) is updated using Eq.(\ref{eq4});
\ENDFOR
\FOR {each patrollers $i \in [1,SD]$} 
\STATE The location of patrollers is updated using Eq.(\ref{eq5});
\ENDFOR
\STATE Update $X_{best}$ and $f_{g}$.

\FOR{ $m \in [1,N]$}
\IF{($rand$ $>$ $\sigma$)} 
\STATE The $X_{best}$ is updated using Eq.(\ref{eq14}). // $\sigma$ indicates the inertia weighting factor
\ELSE
\STATE The$X_{best}$ is mutated using Eq. (\ref{eq15}).
\ENDIF
\ENDFOR
\STATE Update $X_{best}$ and $f_{g}$.
\STATE Calculate the $r$ using Eq. (\ref{eq8}). // $r$ is generated by a random roulette strategy
\IF{($rand$ $<$ $r$)}
\STATE $X_{best}$ $\gets$ $x_{i D}^{b e s t^{\prime}}$; // The $X_{best}$ is mutated using Eq. (\ref{eq9}).
\ENDIF
\STATE Rearrange all of the population's $F_{X}$ in ascending order.
\STATE $X_{best}$ $\gets$ $x_{best}^{t+1}$; 
\STATE $f_{g}$ $\gets$  $f(X_{best})$;// Update $X_{best}$ and $f_{g}$.
\STATE $t$ $\gets$ $t$ + 1;
\ENDWHILE
\RETURN $X_{best}$, $f_{g}$. 
\end{footnotesize}
\end{algorithmic}
\end{breakablealgorithm}

\begin{table}[]
\centering

\small
\caption{Datasets description.} \label{T1}
 \scalebox{0.95}{
\begin{tabular}{llllll}
\hline No. & Dataset      & \#Feat & \#SMP  & \#CL & Area   \\\hline 
1   & BreastCO & 9         & 699  & 2       & Medical  \\
2   & BreastCWD     & 30        & 569  & 2       & Medical\\
3   & Clean-1       & 166       & 476  & 2       & Physical \\
4   & Clean-2       & 166       & 6598 & 2       & Physical \\
5   & CongressVR   & 16        & 435  & 2       & Social   \\
6   & Exactly-1      & 13        & 1000 & 2       & Biology  \\
7   & Exactly-2     & 13        & 1000 & 2       & Biology \\
8   & StatlogH      & 13        & 270  & 5       & Life      \\
9   & IonosphereVS & 34        & 351  & 2       & Physical   \\
10  & KrvskpEW     & 36        & 3196 & 2       & Game     \\
11  & Lymphography & 18        & 148  & 4       & Medical   \\
12  & M-of-n       & 13        & 1000 & 2       & Biology   \\
13  & Penglung   & 325       & 73   & 2       & Biology   \\
14  & Semeion      & 265       & 1593 & 2       & Computer  \\
15  & SonarMR      & 60        & 208  & 2       & Physical  \\
16  & Spectheart      & 22        & 267  & 2       & Life     \\
17  & 3T Endgame  & 9         & 958  & 2       & Game     \\
18  & Vote         & 16        & 300  & 2       & Life     \\
19  & WaveformV2   & 40        & 5000 & 3       & Physical \\
20  & Wine       & 13        & 178  & 3       & Physical  \\
21  & Zoology          & 16        & 101  & 7       & Life    \\\hline 
\end{tabular}
}
\end{table}

\section{Experimental results} \label{sec:Experimental}
All experiments in this section were carried out on a 64-bit Windows 10 educational computer with an Intel (R) Xeon (R) E-2224 CPU @ 3.40GHz and 16.0 GB of memory. All experiments in this chapter are based on the MATLAB (version 9.11.01769968 (R2021b)) platform to test the performance of the proposed algorithm.
\subsection{Evaluation of TFSSA}
In order to verify the effectiveness and superiority of the proposed algorithm TFSSA, the CEC2020 benchmark function was selected to test the performance of TFSSA and compared with 7 algorithms which are ABC, PSO, CSO, DE, SSA, OFA and SHADE.
\subsubsection{Benchmark functions}
CEC benchmarks are the most widely used benchmark problems and have been used by a huge number of research scientists to test their algorithms. The most popular set of benchmark functions in the last dozen years include CEC2008, CEC2010, CEC2013, CEC2014, CEC2017, CEC2019, and CEC2020. The improved methods and problems sometimes need to update the traditional test standards, so this paper uses the classical and newer standard test function set CEC2020 to test the comprehensive performance of the proposed algorithm TFSSA\cite{mohamed2020evaluating}. CEC2020 includes one unimodal function (CEC2020 \_F1), three basic functions (CEC2020 \_F2 – CEC2020 \_F4), and three mixed functions (CEC2020 \_F5
– CEC2020\_ F7) and three synthesis functions (CEC2020 \_F8 – CEC2020 \_F10), as shown in the Table\ref{tab:CEC 2020}.

\subsubsection{Parameter setting}
Artifical Bee Colony algorithm(ABC)  \cite{karaboga2009comparative}, PSO, Competitive Swarm Optimizer(CSO)  \cite{cheng2014competitive}, DE  \cite{liu2005fuzzy}, SSA, Optimal Foraging Algorithm(OFA)  \cite{zhu2017optimal}, Success History based Adaptive Differential Evolution(SHADE)  \cite{viktorin2016success} is used as the benchmark algorithm for comparison.
The number of trials for releasing a food resource of ABC is 20, the inertia weight of PSO is 0.4, the social factor of CSO is 0.1, the conjugate constant of DE is 0.9, the mutation factor of DE is 0.5, the chaos disturbance factor parameter is 0.7, and the Lévy flight parameter is 1.5. The number of populations of all algorithms is set to
100, the number of runs is set to 30. Each algorithm repeats the experiment 30 times independently to obtain statistical results.
In order to verify the optimization performance of the proposed algorithm for different dimension problems, this paper conducts experiments on the solution schemes of 10, 15, and 20 dimensions (D = 10, 15, and 20). The maximum number of function evaluations is set to 10000 * D.
\subsubsection{Statistical test}

The significance level is used to compare whether the two algorithms have a significant difference in performance. 
Wilcoxon rank sum test with $\alpha$=0.05 \cite{li2022self}. The original assumption is that the performance of TFSSA and the comparison algorithm is independent. When rejecting the original hypothesis, this paper uses three symbols to indicate whether there is a significant difference in the performance between TFSSA and the comparison algorithm.

(1) $+$: TFSSA performs significantly better than the comparison algorithm.

(2) $=$: The performance of TFSSA is not significantly related to the performance of the comparison algorithm.

(3) $-$: TFSSA's performance is not significantly better than the comparison algorithm.
\begin{figure}[]
	\begin{subfigure}{}
		\centering
		\includegraphics[width=1\linewidth]{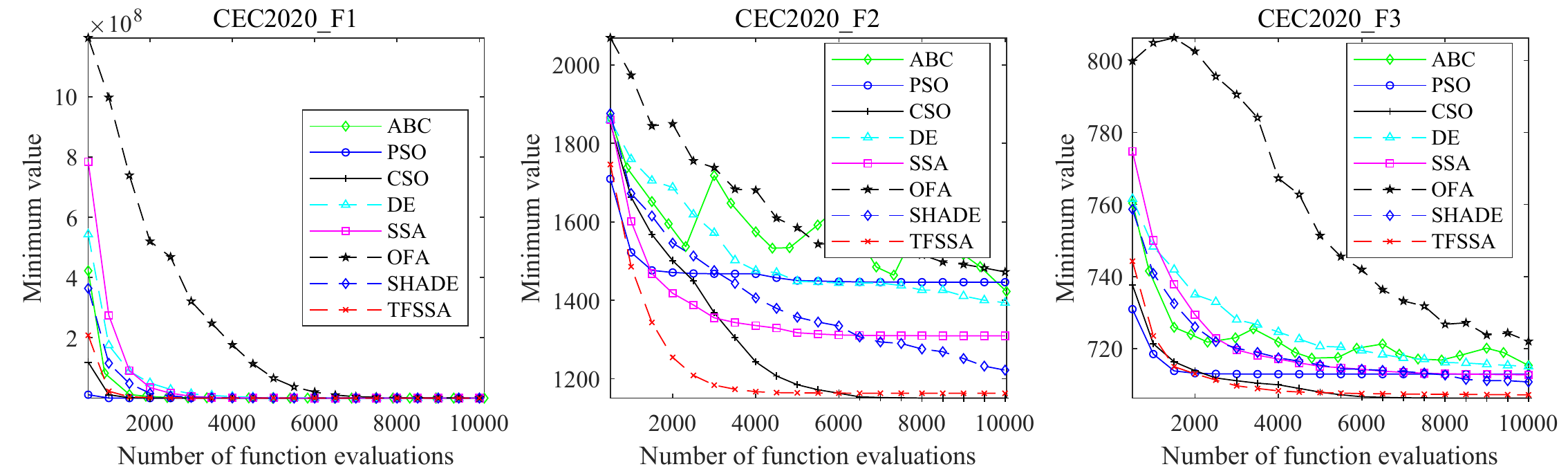}
	\end{subfigure}
	\begin{subfigure}{}
		\centering
		\includegraphics[width=1\linewidth]{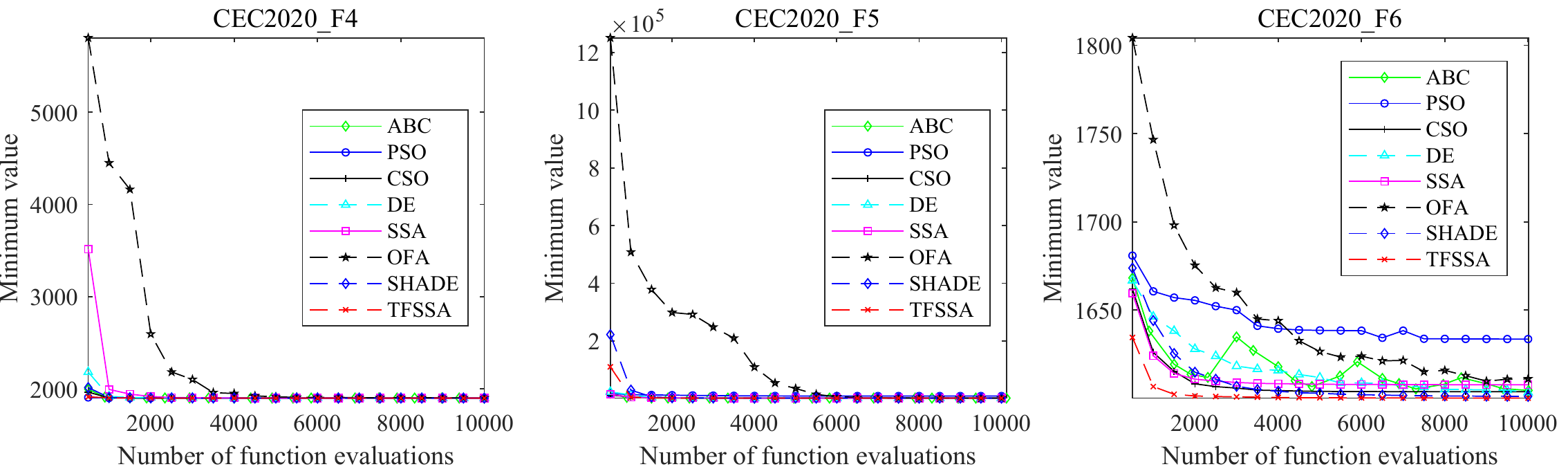}
	\end{subfigure}
	\begin{subfigure}{}
		\centering
		\includegraphics[width=1\linewidth]{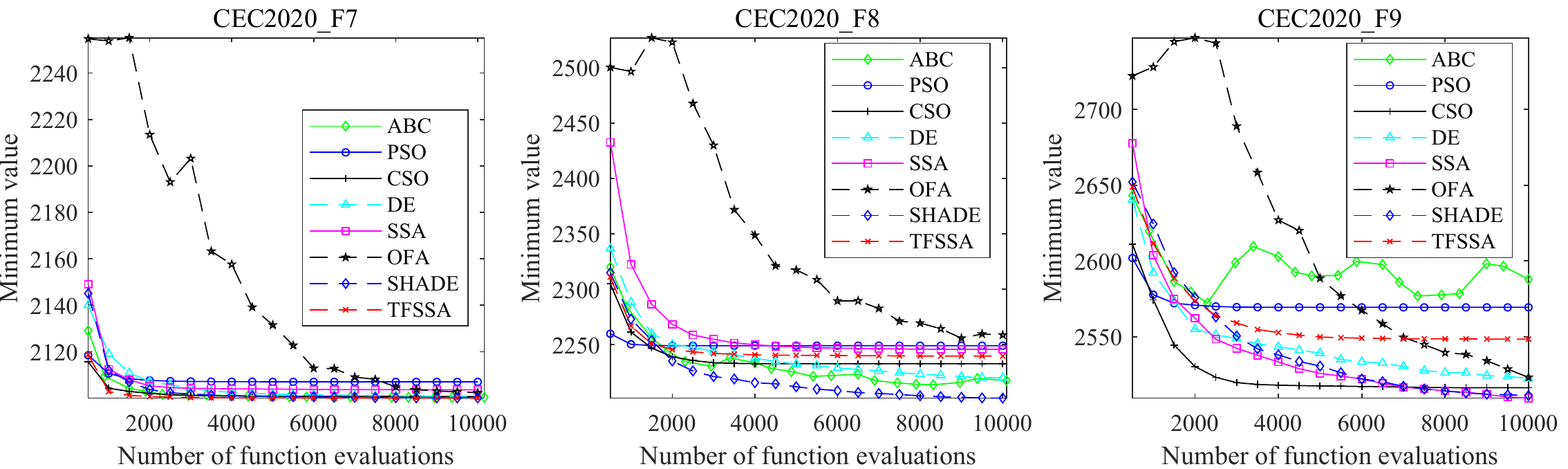}
	\end{subfigure}
\caption{Convergence curves of different algorithms on CEC2020 F1-F9 D=10}
	\label{D1 1-9}
\end{figure}
\begin{figure}[]
	\begin{subfigure}{}
		\centering
		\includegraphics[width=1\linewidth]{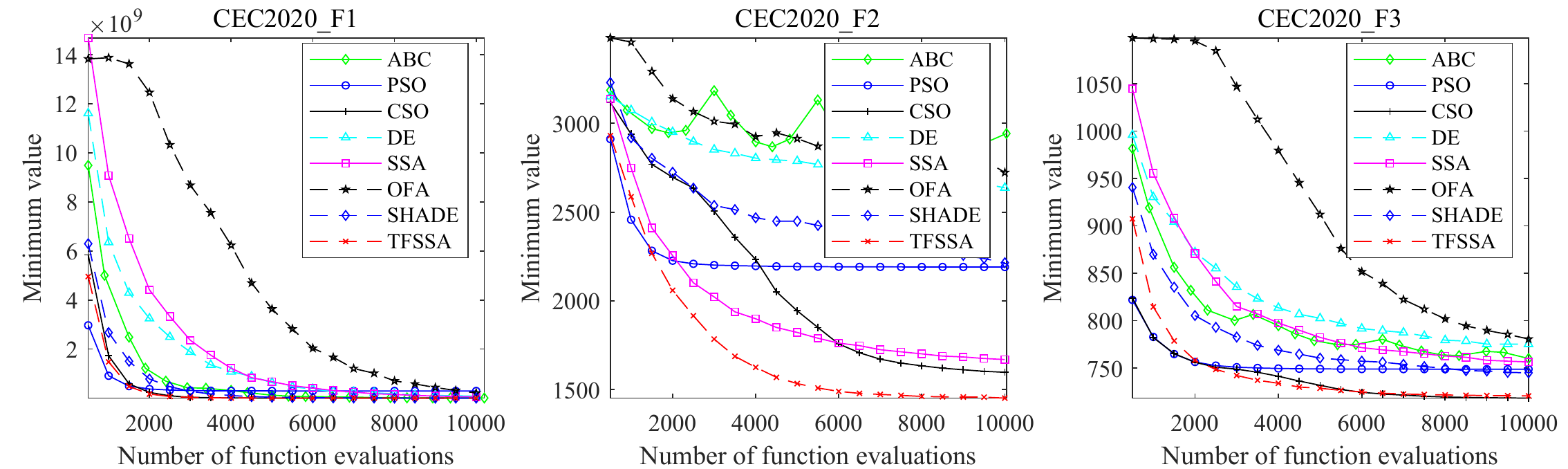}
	\end{subfigure}
	\begin{subfigure}{}
		\centering
		\includegraphics[width=1\linewidth]{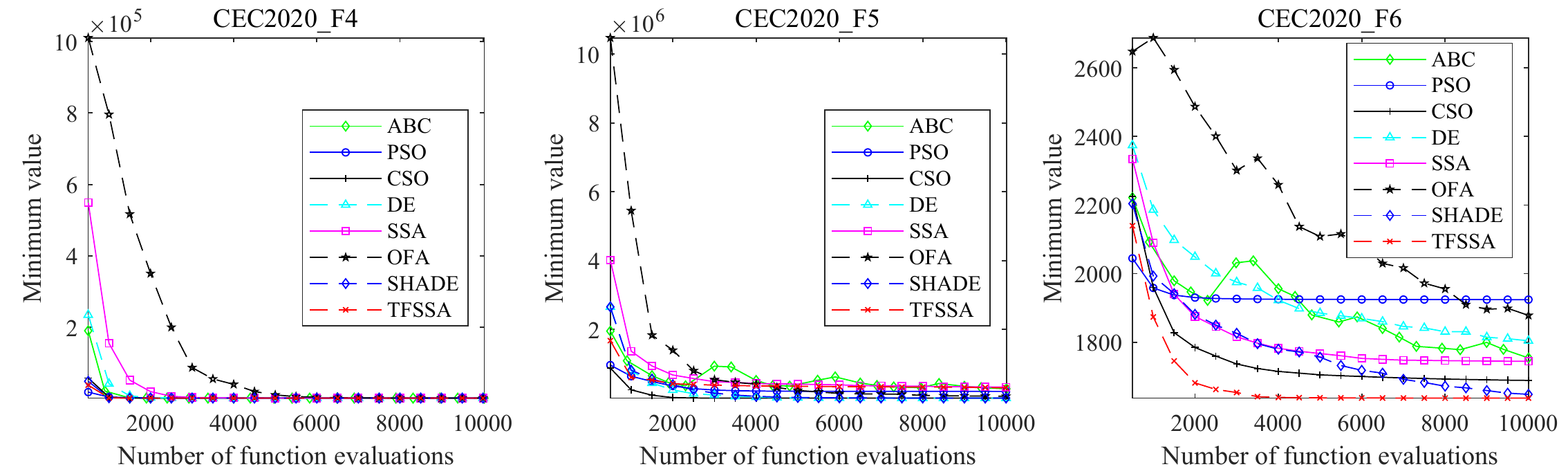}
	\end{subfigure}
	\begin{subfigure}{}
		\centering
		\includegraphics[width=1\linewidth]{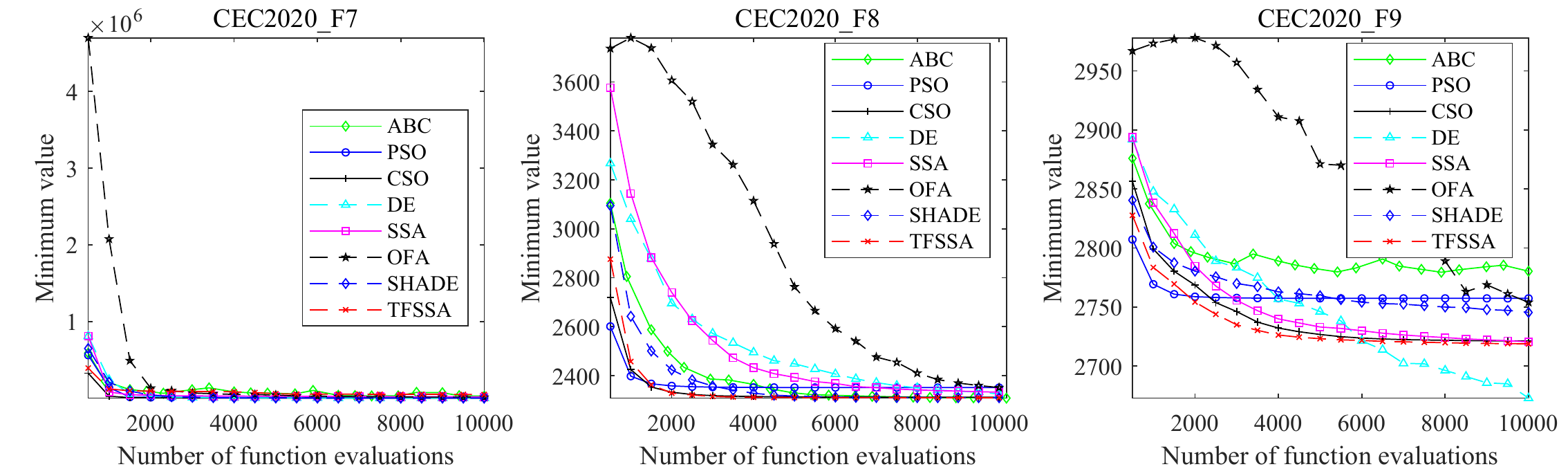}
	\end{subfigure}
		\caption{Convergence curves of different algorithms on CEC2020 F1-F9 D=15}
	\label{D2 1-9}
\end{figure}
\begin{figure}[]
	\begin{subfigure}{}
		\centering
		\includegraphics[width=1\linewidth]{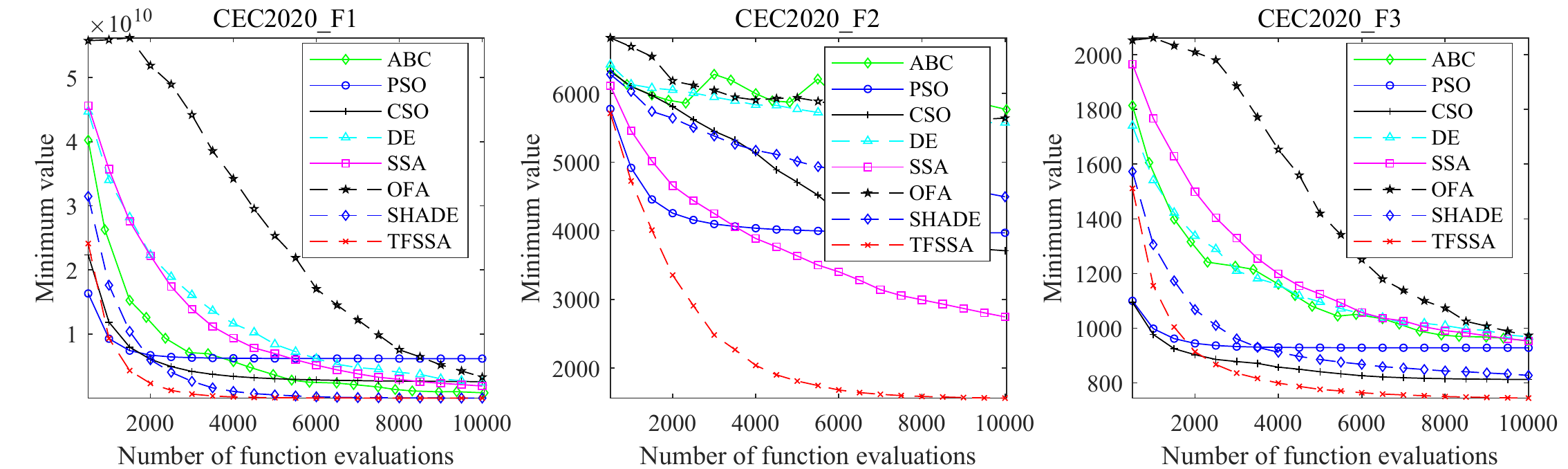}
	\end{subfigure}
	\begin{subfigure}{}
		\centering
		\includegraphics[width=1\linewidth]{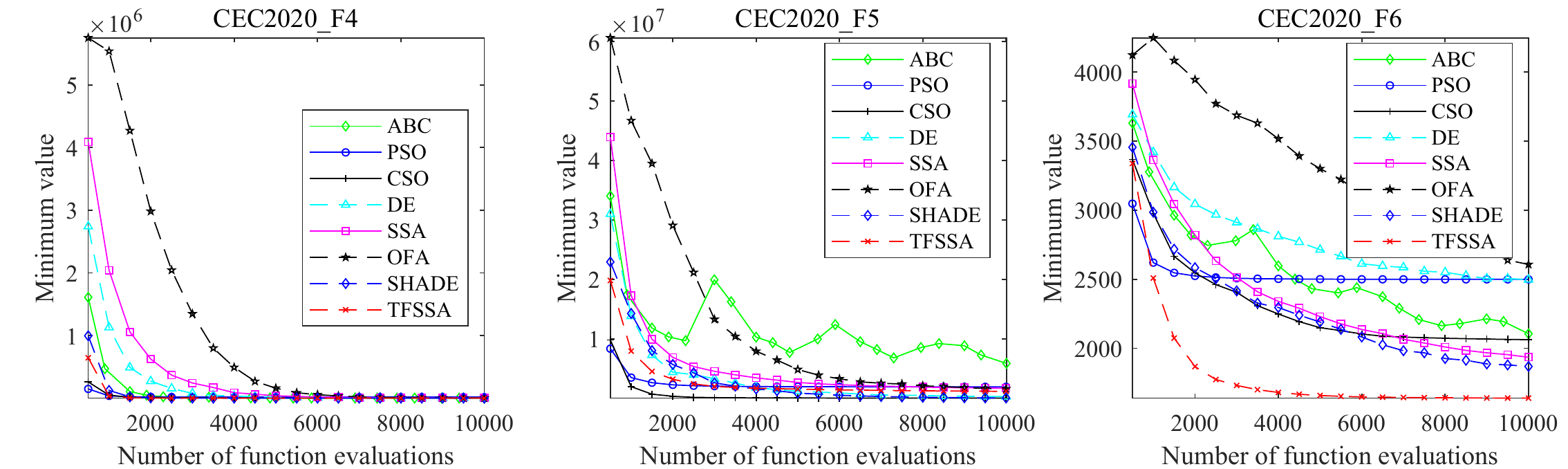}
	\end{subfigure}
	\begin{subfigure}{}
		\centering
		\includegraphics[width=1\linewidth]{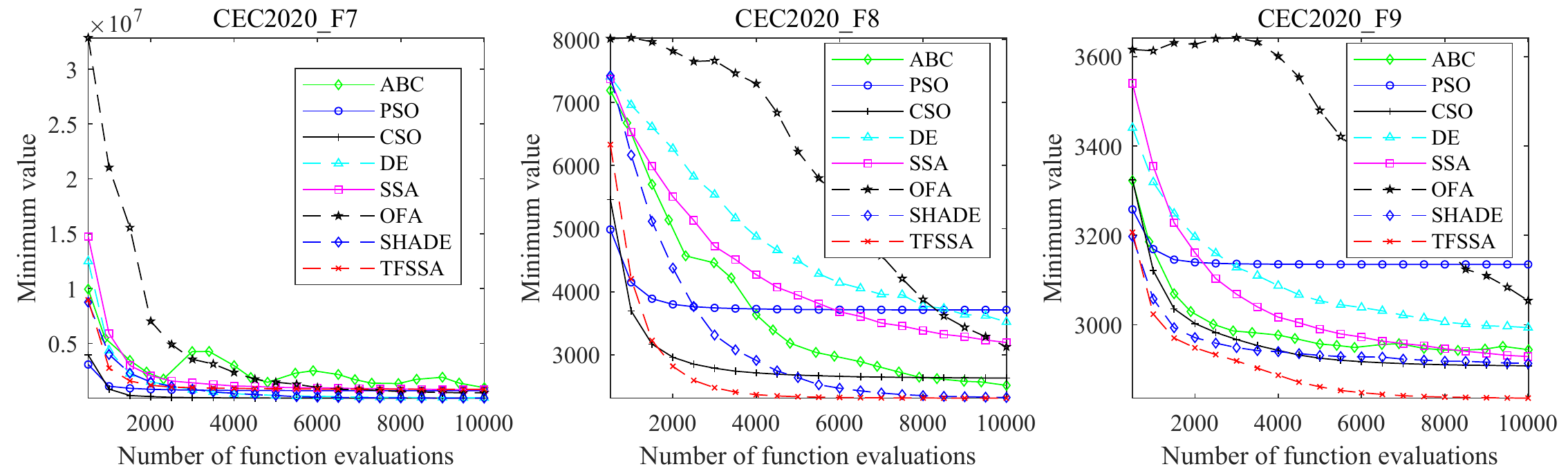}
	\end{subfigure}
	
		\caption{Convergence curves of different algorithms on CEC2020 F1-F9 D=20}
	\label{D3 1-9}
\end{figure}

\begin{table}[]
	\caption{CEC 2020 test suite.}
	\label{tab:CEC 2020}
	\begin{tabular}{|clll|ll}
		\cline{1-4}
		\multicolumn{1}{|l|}{} &
		\multicolumn{1}{c|}{\textbf{No.}} &
		\multicolumn{1}{c|}{\textbf{Functions}} &
		\textbf{${F_i}^{*}$= ${F_i}{(x^{*})}$} &
	 &
	\\ \cline{1-4}
	\multicolumn{1}{|l|}{Unimodal Function} &
	\multicolumn{1}{l|}{1} &
	\multicolumn{1}{l|}{Shifted and Rotated Bent Cigar Function(CEC 2017 F1)} &
	100 &
	&
	\\ \cline{1-4}
	\multicolumn{1}{|c|}{\multirow{3}{*}{\begin{tabular}[c]{@{}c@{}}Basic \\ Functions\end{tabular}}} &
	\multicolumn{1}{l|}{2} &
	\multicolumn{1}{l|}{Shifted and Rotated Schwefel’s Function(CEC 2014 F11)} &
	1100 &
	&
	\\ \cline{2-4}
	\multicolumn{1}{|c|}{} &
	\multicolumn{1}{l|}{3} &
	\multicolumn{1}{l|}{Shifted and Rotated Lunacek bi-RastriginFunction (CEC 2017 F7)} &
	700 &
	&
	\\ \cline{2-4}
	\multicolumn{1}{|c|}{} &
	\multicolumn{1}{l|}{4} &
	\multicolumn{1}{l|}{Expanded Rosenbrock’s plus Griewangk’s Function (CEC 2017 F19)} &
	1900 &
	&
	\\ \cline{1-4}
	\multicolumn{1}{|c|}{\multirow{3}{*}{\begin{tabular}[c]{@{}c@{}}Hybrid \\ Functions\end{tabular}}} &
	\multicolumn{1}{l|}{5} &
	\multicolumn{1}{l|}{Hybrid Function 1 (N= 3) (CEC 2014 F17)} &
	1700 &
	&
	\\ \cline{2-4}
	\multicolumn{1}{|c|}{} &
	\multicolumn{1}{l|}{6} &
	\multicolumn{1}{l|}{Hybrid Function 2 (N= 4) (CEC 2017 F16)} &
	1600 &
	&
	\\ \cline{2-4}
	\multicolumn{1}{|c|}{} &
	\multicolumn{1}{l|}{7} &
	\multicolumn{1}{l|}{Hybrid Function 3 (N= 5) (CEC 2014 F21)} &
	2100 &
	&
	\\ \cline{1-4}
	\multicolumn{1}{|c|}{\multirow{3}{*}{\begin{tabular}[c]{@{}c@{}}Composition \\ Functions\end{tabular}}} &
	\multicolumn{1}{l|}{8} &
	\multicolumn{1}{l|}{Composition Function 1 (N=3) (CEC 2017 F22)} &
	2200 &
	&
	\\ \cline{2-4}
	\multicolumn{1}{|c|}{} &
	\multicolumn{1}{l|}{9} &
	\multicolumn{1}{l|}{Composition Function 2 (N = 4) (CEC 2017 F24)} &
	2400 &
	&
	\\ \cline{2-4}
	\multicolumn{1}{|c|}{} &
	\multicolumn{1}{l|}{10} &
	\multicolumn{1}{l|}{Composition Function 3 (N = 5) (CEC 2017 F25)} &
	2500 &
	&
		\\ \cline{1-4}
		\multicolumn{4}{|c|}{Search Range =${[}-100, 100{]}^{D}$} &
		&
		\\ \cline{1-4}
	\end{tabular}
\end{table}
\subsubsection{Solution accuracy analysis}
This section displays the average value (Mean), standard deviation (Std), and Wilcoxon rank sum test results produced by various algorithms on CEC 2020 for each test function. The best results from all experiments are highlighted in bold. The following is a complete description and analysis of the experimental results:

(1) Unimodal function (F1): The F1 in tables \ref{TD1},  \ref{TD2}, and \ref{TD3} show the optimization results of unimodal function obtained by different algorithms. In 10D, 15D, and 20D cases, the average value and standard deviation of the proposed TFSSA in unimodal function are better than other comparison algorithms.

(2) Basic function (F2-F4): From the F2-F4 in tables \ref{TD1},  \ref{TD2}, and \ref{TD3}, it can be seen that all the mean values of TFSSA are better than other algorithms in 10D and 20D. Only in 10D are the average values of the two primary function optimization results (F2 and F3) obtained by TFSSA better than those of other algorithms. CSO obtains the optimal mean values of F4. Regarding standard deviation, F2 and F3 in D = 10 and F3 in D = 15 get the best results. Therefore, compared with other algorithms, the proposed TFSSA performs better on basic functions. The above results show that with the increase of dimension, the proposed TFSSA performs better in simple single peak problems and other multimodal problems.

(3) Mixed function (F5-F7): It can be seen from Table \ref{TD1}, Table \ref{TD2} and Table \ref{TD3} that TFSSA has obtained 2 (F6 and F7), 1 (F6) and 1 (F6) optimal mean in the mixed function under 10D, 15D, and 20D respectively. SHADE has obtained an F5 optimal mean in D = 10. CSO has obtained F5 and F7 optimal means at D = 15 and D = 20, which indicates that the stability of TFSSA decreases slightly with the increase in the dimension of the solution. In particular, TFSSA has apparent advantages in the F6 function and is the best in all dimensions compared with the comparison algorithm.

(4) composition function (F8-F10): Table \ref{TD1}, Table \ref{TD2}, and Table \ref{TD3} illustrate the outcomes of solving F8-F10 using different strategies. It can be observed that in the cases of 10D, 15D, and 20D, TFSSA produces the best means on one (F8), two (F10 and F8), and one (F9) composite function, respectively. SSA best obtains F9 and F10 and DE in D = 10, F8 and F9 in D = 15, and F9 and F10 in D = 20 by PSO SHADE. So the composite function has multiple local optimal values and complex properties. The final findings demonstrate that TFSSA can strike a balance between exploration and development.

To sum up, TFSSA obtained 7(D=10), 5(D=15), and 6(D=20) optimal mean values and 0(D=10), 2(D=15), and 1(D=20) suboptimal mean values of 10 functions in all dimensions, which means that the algorithm's accuracy is less affected by the change of dimensions. Under 10D, SHADE obtains an optimal mean value and three optimal standard deviations, SSA and DE obtain an optimal mean value, respectively, and CSO and DE obtain an optimal standard deviation, respectively. CSO challenged the proposed TFSSA performance under the 15D condition, which obtained three optimal mean values and one optimal standard deviation. ABC and DE obtained one optimal mean value and one standard deviation, respectively, while SHADE obtained three optimal standard deviations. At 20D, PSO and CSO obtained 1 and 2 optimal mean values, and PSO, DE, and SHADE obtained 1, 1, and 4 optimal standard deviations, respectively. According to NFL theory, one algorithm can hardly effectively solve all optimization problems. Therefore, the proposed TFSSA algorithm can not obtain the best results on all classical test functions. However, compared with other algorithms, the comprehensive comparison of the best results obtained by the proposed TFSSA algorithm is still ideal, showing the TFSSA algorithm's superiority. The results show that the proposed TFSSA is more suitable for solving 10D problems, and its performance for optimizing test functions with higher dimensions (such as 20D) is slightly poor.

\subsubsection{Algorithm stability analysis}
From the Wilcoxon test results in tables \ref{TD1},  \ref{TD2} and \ref{TD3}, it is observed that TFSSA has significantly better performance than ABC, PSO, CSO, DE, OFA, and SHADE in more than half of the functions. Compared with the performance of DE and SHADE, the performance of TFSSA on six functions is significantly improved when D = 15, but the performance on four functions is significantly reduced. In other words, the performance of TFSSA is much better than that of DE and SHADE at 15D. Compared with the performance of CSO, when D = 10, the performance of TFSSA on seven functions is significantly improved, but the performance on two functions is significantly reduced. When D = 15, the performance of TFSSA on six functions is significantly improved, but the performance on two functions is significantly reduced. When D = 20, the performance of TFSSA on four functions is significantly improved, but the performance on six functions is significantly reduced. To some extent, it also demonstrates that the stability of TFSSA at 10D is more significant than that at 15D and greater than that at 20D.
\subsubsection{Convergence rate analysis}
This section describes the convergence rates obtained by different algorithms when solving the CEC2020 test function.
The convergence rate of the optimal global solution is an important index to test the performance of EA. Figures \ref{D1 1-9}, \ref{D2 1-9}, and \ref{D3 1-9} respectively show the convergence rates obtained by different algorithms for solving test functions under CEC 2020. The abscissa is the number of function evaluations, and the ordinate is the minimum value obtained when each algorithm runs independently.

It can be seen that TFSSA converges faster than other comparison algorithms in functions F1, F4, F5, F6, F7, F8, and F10 (D=10), in functions F1, F4, F5, F7, F8, and F10 (D=15), and functions F4, F5, F7 and F10 (D=20), especially in the early evolutionary stage of these classical test functions. In addition, TFSSA has no apparent advantages in other functions. Except for F8 (D=10), F9 (D=10, D=15), and F10 (D=10), TFSSA's convergence speed is slightly worse than other algorithms, but its accuracy is still the best. At this time, it can be seen that the TFSSA algorithm has a relatively strong exploration ability in the late stage, which means that the proposed TFSSA algorithm can maintain relatively high population diversity and avoid premature convergence. TFSSA showed the best convergence rate performance for most of the test functions in the entire optimization process. Therefore, it can be concluded that the proposed TFSSA has a very promising exploration capability in most test functions.

\subsubsection{Runtime analysis}
Table \ref{T-time} shows the running time of TFSSA and the comparison algorithm. It can be seen from table \ref{T-time} that the running time of TFSSA on all test functions is slightly longer than that of most comparison algorithms. The main reasons for the above phenomena are as follows: (1). After roulette strategy calculation, $r $ and $and $ are compared, and the best individual is mutated. This stage takes more time than the original SSA. (2). The optimal individual needs to reorder the fitness function values after the chaotic mutation of $\psi$-Tent. Sorting is time-consuming, so this stage is also one of the main reasons for the increase in running time. Although the running time of TFSSA in this study is slightly higher than most of these comparison algorithms, in the end, considering the performance improvement, these additional running times are ignored to some extent.
\begin{table}[htbp]
	\centering
	\caption{Running time of different algorithms on CEC2020 F1-F9 D=10,15,20.}\label{T-time}%
	\begin{tabular}{cccccccccc}
		\toprule
		Problem    & Dimension    & ABC   & PSO   & CSO   & DE    & SSA   & OFA   & SHADE & \multicolumn{1}{p{4.19em}}{TFSSA} \\
		\midrule
		F1    & \multirow{10}[2]{*}{D=10} & 1.17E-01 & 1.23E-01 & 1.90E-01 & 1.32E-01 & 1.94E-01 & 1.03E-01 & 1.78E-01 & 1.41E-01 \\
		F2    &       & 1.36E-01 & 1.44E-01 & 2.10E-01 & 1.48E-01 & 2.17E-01 & 1.05E-01 & 2.13E-01 & 1.64E-01 \\
		F3    &       & 1.30E-01 & 1.36E-01 & 1.99E-01 & 1.41E-01 & 1.96E-01 & 1.03E-01 & 2.02E-01 & 1.46E-01 \\
		F4    &       & 1.21E-01 & 1.30E-01 & 1.95E-01 & 1.31E-01 & 2.02E-01 & 1.00E-01 & 1.98E-01 & 1.36E-01 \\
		F5    &       & 1.72E-01 & 1.26E-01 & 2.09E-01 & 1.44E-01 & 2.15E-01 & 1.20E-01 & 2.13E-01 & 1.53E-01 \\
		F6    &       & 1.48E-01 & 1.30E-01 & 2.31E-01 & 1.52E-01 & 2.00E-01 & 1.19E-01 & 1.96E-01 & 1.60E-01 \\
		F7    &       & 1.42E-01 & 1.54E-01 & 2.31E-01 & 1.95E-01 & 2.28E-01 & 1.09E-01 & 2.09E-01 & 1.96E-01 \\
		F8    &       & 2.02E-01 & 1.68E-01 & 2.56E-01 & 1.67E-01 & 2.95E-01 & 1.51E-01 & 2.38E-01 & 1.67E-01 \\
		F9    &       & 2.60E-01 & 1.76E-01 & 3.91E-01 & 2.21E-01 & 3.12E-01 & 1.66E-01 & 2.81E-01 & 2.15E-01 \\
		F10   &       & 2.60E-01 & 1.88E-01 & 2.65E-01 & 1.92E-01 & 2.57E-01 & 1.46E-01 & 2.38E-01 & 2.15E-01 \\
		\midrule
		F1    & \multirow{10}[2]{*}{D=15} & 2.73E-01 & 1.74E-01 & 2.16E-01 & 1.80E-01 & 2.51E-01 & 1.39E-01 & 1.98E-01 & 1.93E-01 \\
		F2    &       & 1.72E-01 & 1.46E-01 & 2.14E-01 & 2.32E-01 & 2.48E-01 & 1.30E-01 & 2.08E-01 & 1.97E-01 \\
		F3    &       & 1.41E-01 & 1.44E-01 & 1.81E-01 & 1.71E-01 & 2.43E-01 & 1.23E-01 & 1.88E-01 & 1.98E-01 \\
		F4    &       & 1.36E-01 & 1.31E-01 & 1.97E-01 & 1.55E-01 & 2.21E-01 & 1.27E-01 & 1.86E-01 & 1.78E-01 \\
		F5    &       & 1.88E-01 & 1.58E-01 & 2.21E-01 & 1.95E-01 & 2.54E-01 & 1.27E-01 & 2.13E-01 & 2.04E-01 \\
		F6    &       & 1.61E-01 & 1.64E-01 & 2.22E-01 & 1.83E-01 & 2.53E-01 & 1.20E-01 & 2.13E-01 & 1.92E-01 \\
		F7    &       & 1.81E-01 & 1.62E-01 & 2.26E-01 & 1.89E-01 & 2.55E-01 & 1.34E-01 & 2.11E-01 & 2.04E-01 \\
		F8    &       & 1.88E-01 & 1.79E-01 & 2.88E-01 & 2.00E-01 & 2.75E-01 & 1.57E-01 & 2.27E-01 & 2.32E-01 \\
		F9    &       & 2.01E-01 & 2.21E-01 & 2.80E-01 & 2.62E-01 & 2.98E-01 & 1.75E-01 & 2.67E-01 & 2.52E-01 \\
		F10   &       & 2.12E-01 & 2.12E-01 & 3.32E-01 & 2.12E-01 & 3.06E-01 & 1.88E-01 & 2.75E-01 & 2.35E-01 \\
		\midrule
		F1    & \multirow{10}[2]{*}{D=20} & 2.88E-01 & 1.62E-01 & 1.97E-01 & 1.80E-01 & 2.70E-01 & 1.26E-01 & 2.01E-01 & 2.22E-01 \\
		F2    &       & 1.82E-01 & 1.67E-01 & 2.24E-01 & 2.81E-01 & 2.96E-01 & 1.50E-01 & 2.25E-01 & 2.29E-01 \\
		F3    &       & 1.43E-01 & 1.64E-01 & 1.96E-01 & 1.97E-01 & 3.04E-01 & 1.29E-01 & 2.11E-01 & 2.28E-01 \\
		F4    &       & 1.51E-01 & 1.35E-01 & 2.04E-01 & 1.97E-01 & 2.59E-01 & 1.29E-01 & 2.11E-01 & 2.06E-01 \\
		F5    &       & 1.77E-01 & 1.80E-01 & 2.24E-01 & 2.04E-01 & 2.80E-01 & 1.48E-01 & 2.27E-01 & 2.32E-01 \\
		F6    &       & 1.71E-01 & 1.81E-01 & 2.09E-01 & 1.95E-01 & 2.78E-01 & 1.41E-01 & 2.13E-01 & 2.25E-01 \\
		F7    &       & 1.91E-01 & 1.73E-01 & 2.24E-01 & 2.00E-01 & 2.96E-01 & 1.62E-01 & 2.29E-01 & 2.14E-01 \\
		F8    &       & 2.23E-01 & 2.02E-01 & 2.73E-01 & 2.50E-01 & 3.09E-01 & 1.99E-01 & 2.59E-01 & 2.51E-01 \\
		F9    &       & 2.27E-01 & 2.60E-01 & 3.28E-01 & 2.99E-01 & 3.28E-01 & 2.27E-01 & 3.09E-01 & 2.90E-01 \\
		F10   &       & 2.50E-01 & 2.43E-01 & 3.14E-01 & 2.50E-01 & 3.45E-01 & 1.85E-01 & 2.92E-01 & 2.70E-01 \\
		\bottomrule
	\end{tabular}%
\end{table}%
\subsubsection{Computational complexity analysis}
In this section, the proposed TFSSA mainly include the following stages:
\begin{itemize}
    \item [(1)] initialization of sparrows.
    \item [(2)] fitness evaluation.
    \item [(3)] fitness ranking.
    \item [(4)] position update. 
\end{itemize}

The computational complexity of stage (1) in SSA is \textit{O(D $\times$ N)}, the computational complexity of stage (2) and (3) is \textit{O(D $\times$ N)}, and the computational complexity of stage (4) is \textit{O((PD + SD + FD)$\times$ T\_max $\times$ D}, which is \textit{O(N $\times$ T\_max $\times$ D)}, hence the total computational complexity of SSA is \textit{O(N)}. 
The computational cost of the TFSSA is primarily different from that of the stage (4) in comparison to SSA. TFSSA has a computational complexity of \textit{O(T\_max $\times$ D $\times$ 3N)} when it comes to the sparrow's location updating phase. To summarize, the proposed TFSSA and classical SSA have a computational complexity of \textit{O(N)}. Furthermore, the space complexity of both algorithms is \textit{O(D $\times$ N)} and is mostly influenced by population size \textit{N} and search space dimension \textit{D}. 
Among them, \textit{N} represents the number of sparrows in a flock, \textit{D} represents the dimension of the function, \textit{T\_max} represents the maximum number of iterations, \textit{PD} represents the number of leaders, \textit{FD} represents the number of followers, and \textit{SD} represents the number of patrollers.
\begin{sidewaystable}[thp]
		\centering	
	\caption{Mean, standard deviation and Wilcoxon rank sum test results of different algorithms on CEC2020 (D=10)}
	\label{TD1}
	\begin{tabular}{ccccccccc}
	\hline	& \begin{tabular}[c]{@{}c@{}}ABC\\ Mean\\ (Std)\end{tabular} & \begin{tabular}[c]{@{}c@{}}PSO\\ Mean\\ (Std)\end{tabular} & \begin{tabular}[c]{@{}c@{}}CSO\\ Mean\\ (Std)\end{tabular} & \begin{tabular}[c]{@{}c@{}}DE\\ Mean\\ (Std)\end{tabular} & \begin{tabular}[c]{@{}c@{}}SSA\\ Mean\\ (Std)\end{tabular} & \begin{tabular}[c]{@{}c@{}}OFA\\ Mean\\ (Std)\end{tabular} & \begin{tabular}[c]{@{}c@{}}SHADE\\ Mean\\ (Std)\end{tabular} & \begin{tabular}[c]{@{}c@{}}TFSSA\\ Mean\\ (Std)\end{tabular} \\\hline
	CEC2020\_F1 & \begin{tabular}[c]{@{}c@{}}2.7683e+4 \\ (7.29e+4) +\end{tabular} & \begin{tabular}[c]{@{}c@{}}3.6043e+3 \\ (3.38e+3) +\end{tabular} & \begin{tabular}[c]{@{}c@{}}1.9567e+3 \\ (9.27e+2) =\end{tabular} & \begin{tabular}[c]{@{}c@{}}3.9673e+4 \\ (2.09e+4) -\end{tabular} & \begin{tabular}[c]{@{}c@{}}4.2342e+3 \\ (4.82e+3) =\end{tabular} & \begin{tabular}[c]{@{}c@{}}2.7980e+5\\  (1.54e+5) +\end{tabular} & \begin{tabular}[c]{@{}c@{}}4.7054e+3 \\ (4.24e+3)+\end{tabular} & \textbf{\begin{tabular}[c]{@{}c@{}}5.7683e+2 \\ (2.31e+2)\end{tabular}} \\
	CEC2020\_F2 & \begin{tabular}[c]{@{}c@{}}1.4226e+3\\  (1.63e+2) +\end{tabular} & \begin{tabular}[c]{@{}c@{}}1.4462e+3\\  (1.54e+2) -\end{tabular} & \begin{tabular}[c]{@{}c@{}}1.1631e+3 \\ (6.93e+1) +\end{tabular} & \begin{tabular}[c]{@{}c@{}}1.3938e+3\\  (8.74e+1) +\end{tabular} & \begin{tabular}[c]{@{}c@{}}1.3095e+3 \\ (1.19e+2) +\end{tabular} & \begin{tabular}[c]{@{}c@{}}1.4722e+3 \\ (1.58e+2) +\end{tabular} & \begin{tabular}[c]{@{}c@{}}1.2221e+3 \\ (5.64e+1) +\end{tabular} & \textbf{\begin{tabular}[c]{@{}c@{}}1.1508e+3\\  (6.11e+0)\end{tabular}} \\
	CEC2020\_F3 & \begin{tabular}[c]{@{}c@{}}7.1526e+2\\  (3.11e+0) =\end{tabular} & \begin{tabular}[c]{@{}c@{}}7.1286e+2\\  (5.03e+0) +\end{tabular} & \begin{tabular}[c]{@{}c@{}}7.0711e+2 \\ (1.44e+0) +\end{tabular} & \begin{tabular}[c]{@{}c@{}}7.1502e+2 \\ (2.83e+0) +\end{tabular} & \begin{tabular}[c]{@{}c@{}}7.1267e+2 \\ (5.42e+0) +\end{tabular} & \begin{tabular}[c]{@{}c@{}}7.2197e+2 \\ (5.19e+0) =\end{tabular} & \begin{tabular}[c]{@{}c@{}}7.1070e+2 \\ (1.63e+0) +\end{tabular} & \textbf{\begin{tabular}[c]{@{}c@{}}7.0620e+2\\  (8.06e-1)\end{tabular}} \\
	CEC2020\_F4 & \begin{tabular}[c]{@{}c@{}}1.9009e+3\\  (2.90e-1) +\end{tabular} & \begin{tabular}[c]{@{}c@{}}1.9008e+3 \\ (7.54e-1) +\end{tabular} & \begin{tabular}[c]{@{}c@{}}1.9003e+3 \\ (9.82e-2) +\end{tabular} & \begin{tabular}[c]{@{}c@{}}1.9008e+3 \\ (2.76e-1) +\end{tabular} & \begin{tabular}[c]{@{}c@{}}1.9005e+3\\  (2.74e-1) +\end{tabular} & \begin{tabular}[c]{@{}c@{}}1.9028e+3 \\ (1.01e+0) -\end{tabular} & \begin{tabular}[c]{@{}c@{}}1.9006e+3\\  (1.17e-1) -\end{tabular} & \textbf{\begin{tabular}[c]{@{}c@{}}1.9003e+3\\  (1.60e-1)\end{tabular}} \\
	CEC2020\_F5 & \begin{tabular}[c]{@{}c@{}}1.7150e+3 \\ (1.76e+1) =\end{tabular} & \begin{tabular}[c]{@{}c@{}}8.9157e+3\\  (6.58e+3) -\end{tabular} & \begin{tabular}[c]{@{}c@{}}1.7254e+3 \\ (2.73e+1) +\end{tabular} & \begin{tabular}[c]{@{}c@{}}1.7314e+3 \\ (1.28e+1) =\end{tabular} & \begin{tabular}[c]{@{}c@{}}1.7552e+3 \\ (7.78e+1) =\end{tabular} & \begin{tabular}[c]{@{}c@{}}1.7356e+3\\  (1.49e+1) =\end{tabular} & \textbf{\begin{tabular}[c]{@{}c@{}}1.7065e+3 \\ (2.65e+0) +\end{tabular}} & \begin{tabular}[c]{@{}c@{}}1.7595e+3 \\ (6.06e+1)\end{tabular} \\
	CEC2020\_F6 & \begin{tabular}[c]{@{}c@{}}1.6044e+3 \\ (4.07e+0) +\end{tabular} & \begin{tabular}[c]{@{}c@{}}1.6336e+3 \\ (4.65e+1) -\end{tabular} & \begin{tabular}[c]{@{}c@{}}1.6037e+3 \\ (6.16e+0) -\end{tabular} & \begin{tabular}[c]{@{}c@{}}1.6033e+3 \\ (1.43e+0) -\end{tabular} & \begin{tabular}[c]{@{}c@{}}1.6077e+3 \\ (1.40e+1) +\end{tabular} & \begin{tabular}[c]{@{}c@{}}1.6110e+3 \\ (1.10e+1) -\end{tabular} & \begin{tabular}[c]{@{}c@{}}1.6011e+3\\  \textbf{(2.08e-1)} -\end{tabular} & \begin{tabular}[c]{@{}c@{}}1.6001e+3\\  (1.46e-1)\end{tabular} \\
	CEC2020\_F7 & \begin{tabular}[c]{@{}c@{}}\textbf{2.1003e+3}\\  (3.11e-1) +\end{tabular} & \begin{tabular}[c]{@{}c@{}}2.1070e+3\\  (1.23e+1) -\end{tabular} & \begin{tabular}[c]{@{}c@{}}2.1007e+3\\  (3.31e-1) -\end{tabular} & \begin{tabular}[c]{@{}c@{}}2.1008e+3 \\ (1.46e-1) +\end{tabular} & \begin{tabular}[c]{@{}c@{}}2.1036e+3 \\ (1.02e+1) +\end{tabular} & \begin{tabular}[c]{@{}c@{}}2.1024e+3\\  (1.39e+0) =\end{tabular} & \begin{tabular}[c]{@{}c@{}}2.1001e+3\\  \textbf{(2.87e-2) +}\end{tabular} & \begin{tabular}[c]{@{}c@{}}\textbf{2.1000e+3} \\ (1.22e-2)\end{tabular} \\
	CEC2020\_F8 & \begin{tabular}[c]{@{}c@{}}2.2175e+3 \\ (3.40e+1) =\end{tabular} & \begin{tabular}[c]{@{}c@{}}2.2490e+3\\  (4.61e+1) =\end{tabular} & \begin{tabular}[c]{@{}c@{}}2.2326e+3 \\ (4.56e+1) +\end{tabular} & \begin{tabular}[c]{@{}c@{}}2.2200e+3\\  (3.11e+0) =\end{tabular} & \begin{tabular}[c]{@{}c@{}}2.2456e+3 \\ (4.84e+1) =\end{tabular} & \begin{tabular}[c]{@{}c@{}}2.2586e+3 \\ (3.28e+1) +\end{tabular} & \begin{tabular}[c]{@{}c@{}}2.2396e+3 \\ (5.28e+1) =\end{tabular} & \textbf{\begin{tabular}[c]{@{}c@{}}2.2016e+3\\  (1.22e+0)\end{tabular}} \\
	CEC2020\_F9 & \begin{tabular}[c]{@{}c@{}}2.5879e+3\\  (7.23e+1) +\end{tabular} & \begin{tabular}[c]{@{}c@{}}2.5694e+3 \\ (1.13e+2) +\end{tabular} & \begin{tabular}[c]{@{}c@{}}2.5163e+3\\  (5.33e+1) +\end{tabular} & \begin{tabular}[c]{@{}c@{}}2.5227e+3\\  \textbf{(6.61e+0)} +\end{tabular} & \begin{tabular}[c]{@{}c@{}}\textbf{2.5095e+3} \\ (2.47e+1) =\end{tabular} & \begin{tabular}[c]{@{}c@{}}2.5232e+3 \\ (6.69e+0) +\end{tabular} & \begin{tabular}[c]{@{}c@{}}2.5112e+3 \\ (4.74e+1) =\end{tabular} & \begin{tabular}[c]{@{}c@{}}2.5485e+3 \\ (8.11e+1)\end{tabular} \\
	CEC2020\_F10 & \begin{tabular}[c]{@{}c@{}}2.8474e+3\\  (2.18e-2) =\end{tabular} & \begin{tabular}[c]{@{}c@{}}2.8515e+3\\  (1.08e+1) +\end{tabular} & \begin{tabular}[c]{@{}c@{}}2.8474e+3 \\ (5.47e-3) +\end{tabular} & \begin{tabular}[c]{@{}c@{}}\textbf{2.7394e+3} \\ (6.38e+1) +\end{tabular} & \begin{tabular}[c]{@{}c@{}}2.8433e+3 \\ (1.48e+1) =\end{tabular} & \begin{tabular}[c]{@{}c@{}}2.8518e+3 \\ (2.56e+0) -\end{tabular} & \begin{tabular}[c]{@{}c@{}}2.8474e+3 \\ (7.79e-2) +\end{tabular} & \begin{tabular}[c]{@{}c@{}}2.8475e+3 \\ \textbf{(9.47e-2)}\end{tabular} \\
	+/-/= & 6/0/4 & 5/4/1 & 7/2/1 & 6/2/2 & 5/0/5 & 4/3/3 & 6/2/2 &  \\\hline 
\end{tabular}
\end{sidewaystable}


\begin{sidewaystable}[thp]
	\centering	
	\caption{Mean, standard deviation and Wilcoxon rank sum test results of different algorithms on CEC2020 (D=15)}
	\label{TD2}
	\begin{tabular}{ccccccccc}
	\hline	& \begin{tabular}[c]{@{}c@{}}ABC\\ Mean\\ (Std)\end{tabular} & \begin{tabular}[c]{@{}c@{}}PSO\\ Mean\\ (Std)\end{tabular} & \begin{tabular}[c]{@{}c@{}}CSO\\ Mean\\ (Std)\end{tabular} & \begin{tabular}[c]{@{}c@{}}DE\\ Mean\\ (Std)\end{tabular} & \begin{tabular}[c]{@{}c@{}}SSA\\ Mean\\ (Std)\end{tabular} & \begin{tabular}[c]{@{}c@{}}OFA\\ Mean\\ (Std)\end{tabular} & \begin{tabular}[c]{@{}c@{}}SHADE\\ Mean\\ (Std)\end{tabular} & \begin{tabular}[c]{@{}c@{}}TFSSA\\ Mean\\ (Std)\end{tabular} \\\hline
	CEC2020\_F1 & \begin{tabular}[c]{@{}c@{}}4.4019e+6 \\ (2.51e+6) +\end{tabular} & \begin{tabular}[c]{@{}c@{}}2.9111e+8\\  (2.59e+8) +\end{tabular} & \begin{tabular}[c]{@{}c@{}}4.0135e+5\\  (8.96e+5) +\end{tabular} & \begin{tabular}[c]{@{}c@{}}7.9228e+7 \\ (3.10e+7) +\end{tabular} & \begin{tabular}[c]{@{}c@{}}4.8618e+7 \\ (5.11e+7) +\end{tabular} & \begin{tabular}[c]{@{}c@{}}2.1070e+8 \\ (9.23e+7) -\end{tabular} & \begin{tabular}[c]{@{}c@{}}5.8744e+5 \\ (2.12e+5) -\end{tabular} & \textbf{\begin{tabular}[c]{@{}c@{}}1.1879e+5\\  (9.47e+4)\end{tabular}} \\
	CEC2020\_F2 & \begin{tabular}[c]{@{}c@{}}2.9428e+3 \\ (2.08e+2) +\end{tabular} & \begin{tabular}[c]{@{}c@{}}2.1909e+3 \\ (4.08e+2) -\end{tabular} & \begin{tabular}[c]{@{}c@{}}1.5976e+3\\  (2.67e+2) =\end{tabular} & \begin{tabular}[c]{@{}c@{}}2.6368e+3 \\ \textbf{(1.67e+2)} +\end{tabular} & \begin{tabular}[c]{@{}c@{}}1.6696e+3\\  (2.04e+2) -\end{tabular} & \begin{tabular}[c]{@{}c@{}}2.7231e+3 \\ (1.67e+2) +\end{tabular} & \begin{tabular}[c]{@{}c@{}}2.2149e+3 \\ (1.80e+2) +\end{tabular} & \begin{tabular}[c]{@{}c@{}}\textbf{1.4524e+3} \\ (2.29e+2)\end{tabular} \\
	CEC2020\_F3 & \begin{tabular}[c]{@{}c@{}}7.6011e+2 \\ (7.84e+0) -\end{tabular} & \begin{tabular}[c]{@{}c@{}}7.4882e+2\\  (1.57e+1) +\end{tabular} & \begin{tabular}[c]{@{}c@{}}7.2060e+2 \\ (4.90e+0) +\end{tabular} & \begin{tabular}[c]{@{}c@{}}7.7524e+2\\  (9.53e+0) -\end{tabular} & \begin{tabular}[c]{@{}c@{}}7.5645e+2 \\ (1.44e+1) +\end{tabular} & \begin{tabular}[c]{@{}c@{}}7.8067e+2 \\ (9.54e+0) +\end{tabular} & \begin{tabular}[c]{@{}c@{}}7.4497e+2\\  (5.67e+0) +\end{tabular} & \textbf{\begin{tabular}[c]{@{}c@{}}7.1830e+2\\  (3.48e+0)\end{tabular}} \\
	CEC2020\_F4 & \begin{tabular}[c]{@{}c@{}}1.9048e+3 \\ \textbf{(6.95e-1)} -\end{tabular} & \begin{tabular}[c]{@{}c@{}}2.3811e+3\\  (1.25e+3) +\end{tabular} & \begin{tabular}[c]{@{}c@{}}1.9013e+3 \\ (5.85e-1) =\end{tabular} & \begin{tabular}[c]{@{}c@{}}1.9063e+3 \\ (9.44e-1) -\end{tabular} & \begin{tabular}[c]{@{}c@{}}1.9290e+3 \\ (1.10e+2) -\end{tabular} & \begin{tabular}[c]{@{}c@{}}\textbf{1.9394e+3} \\ (2.93e+1) +\end{tabular} & \begin{tabular}[c]{@{}c@{}}1.9032e+3 \\ (4.50e-1) +\end{tabular} & \begin{tabular}[c]{@{}c@{}}1.9016e+3 \\ (5.06e-1)\end{tabular} \\
	CEC2020\_F5 & \begin{tabular}[c]{@{}c@{}}2.7081e+5 \\ (2.35e+5) =\end{tabular} & \begin{tabular}[c]{@{}c@{}}1.9609e+5\\  (2.33e+5) =\end{tabular} & \textbf{\begin{tabular}[c]{@{}c@{}}2.1201e+3\\  (1.57e+2) +\end{tabular}} & \begin{tabular}[c]{@{}c@{}}3.1328e+3 \\ (3.25e+2) +\end{tabular} & \begin{tabular}[c]{@{}c@{}}3.2132e+5\\  (4.63e+5) +\end{tabular} & \begin{tabular}[c]{@{}c@{}}5.6867e+4 \\ (3.80e+4) =\end{tabular} & \begin{tabular}[c]{@{}c@{}}2.5813e+3 \\ (2.04e+2) +\end{tabular} & \begin{tabular}[c]{@{}c@{}}3.0341e+5\\  (5.41e+5)\end{tabular} \\
	CEC2020\_F6 & \begin{tabular}[c]{@{}c@{}}1.7540e+3 \\ (8.24e+1) +\end{tabular} & \begin{tabular}[c]{@{}c@{}}1.9243e+3\\  (1.23e+2) -\end{tabular} & \begin{tabular}[c]{@{}c@{}}1.6886e+3 \\ (7.38e+1) -\end{tabular} & \begin{tabular}[c]{@{}c@{}}1.8042e+3 \\ (5.76e+1) +\end{tabular} & \begin{tabular}[c]{@{}c@{}}1.7443e+3 \\ (1.02e+2) -\end{tabular} & \begin{tabular}[c]{@{}c@{}}1.8783e+3 \\ (6.44e+1) -\end{tabular} & \begin{tabular}[c]{@{}c@{}}1.6480e+3 \\ \textbf{(3.32e+1)} -\end{tabular} & \begin{tabular}[c]{@{}c@{}}1.6370e+3 \\ (5.21e+1)\end{tabular} \\
	CEC2020\_F7 & \begin{tabular}[c]{@{}c@{}}2.7754e+4\\  (4.28e+4) =\end{tabular} & \begin{tabular}[c]{@{}c@{}}1.1125e+4 \\ (8.70e+3) =\end{tabular} & \begin{tabular}[c]{@{}c@{}}\textbf{2.3156e+3}\\  (1.34e+2) +\end{tabular} & \begin{tabular}[c]{@{}c@{}}2.5866e+3 \\ (1.85e+2) +\end{tabular} & \begin{tabular}[c]{@{}c@{}}1.8706e+4\\  (2.32e+4) +\end{tabular} & \begin{tabular}[c]{@{}c@{}}1.3730e+4\\  (8.65e+3) =\end{tabular} & \begin{tabular}[c]{@{}c@{}}2.3187e+3 \\ \textbf{(8.84e+1)} +\end{tabular} & \begin{tabular}[c]{@{}c@{}}4.8971e+4\\  (1.05e+5)\end{tabular} \\
	CEC2020\_F8 & \begin{tabular}[c]{@{}c@{}}\textbf{2.3084e+3}\\  (1.33e+1) +\end{tabular} & \begin{tabular}[c]{@{}c@{}}2.3514e+3 \\ (3.41e+1) +\end{tabular} & \begin{tabular}[c]{@{}c@{}}2.3118e+3 \\ (2.18e+0) +\end{tabular} & \begin{tabular}[c]{@{}c@{}}2.3251e+3 \\ (1.75e+1) -\end{tabular} & \begin{tabular}[c]{@{}c@{}}2.3326e+3\\  (4.36e+1) +\end{tabular} & \begin{tabular}[c]{@{}c@{}}2.3514e+3 \\ (1.35e+1) +\end{tabular} & \begin{tabular}[c]{@{}c@{}}2.3101e+3 \\ \textbf{(4.05e-2)} -\end{tabular} & \begin{tabular}[c]{@{}c@{}}2.3101e+3\\  (3.76e-2)\end{tabular} \\
	CEC2020\_F9 & \begin{tabular}[c]{@{}c@{}}2.7803e+3 \\ (9.44e+0) +\end{tabular} & \begin{tabular}[c]{@{}c@{}}2.7573e+3\\  (8.56e+1) -\end{tabular} & \begin{tabular}[c]{@{}c@{}}2.7211e+3 \\ (6.03e+1) =\end{tabular} & \textbf{\begin{tabular}[c]{@{}c@{}}2.6728e+3 \\ (4.05e+1) +\end{tabular}} & \begin{tabular}[c]{@{}c@{}}2.7205e+3\\  (9.73e+1) +\end{tabular} & \begin{tabular}[c]{@{}c@{}}2.7540e+3 \\ (6.15e+1) =\end{tabular} & \begin{tabular}[c]{@{}c@{}}2.7454e+3\\  (5.71e+1) -\end{tabular} & \begin{tabular}[c]{@{}c@{}}2.7188e+3 \\ (8.71e+1)\end{tabular} \\
	CEC2020\_F10 & \begin{tabular}[c]{@{}c@{}}2.9523e+3 \\ (2.17e+1) +\end{tabular} & \begin{tabular}[c]{@{}c@{}}2.9583e+3\\  (3.36e+1) +\end{tabular} & \begin{tabular}[c]{@{}c@{}}2.9215e+3\\  (2.21e+1) +\end{tabular} & \begin{tabular}[c]{@{}c@{}}2.9531e+3 \\ (7.78e+0) -\end{tabular} & \begin{tabular}[c]{@{}c@{}}2.9476e+3\\  (3.08e+1) +\end{tabular} & \begin{tabular}[c]{@{}c@{}}2.9844e+3\\  (1.96e+1) +\end{tabular} & \begin{tabular}[c]{@{}c@{}}2.9298e+3 \\ (2.21e+1) +\end{tabular} & \textbf{\begin{tabular}[c]{@{}c@{}}2.9160e+3 \\ (1.31e+0)\end{tabular}} \\
	+/-/= & 6/2/2 & 5/3/2 & 6/1/3 & 6/4/0 & 7/3/0 & 5/2/3 & 6/4/0 & \\\hline
	\end{tabular}
\end{sidewaystable}

\begin{sidewaystable}[thp]
	\centering	
	\caption{Mean, standard deviation and Wilcoxon rank sum test results of different algorithms on CEC2020 (D=20)}
	\label{TD3}
	\begin{tabular}{ccccccccc}
		\hline	& \begin{tabular}[c]{@{}c@{}}ABC\\ Mean\\ (Std)\end{tabular} & \begin{tabular}[c]{@{}c@{}}PSO\\ Mean\\ (Std)\end{tabular} & \begin{tabular}[c]{@{}c@{}}CSO\\ Mean\\ (Std)\end{tabular} & \begin{tabular}[c]{@{}c@{}}DE\\ Mean\\ (Std)\end{tabular} & \begin{tabular}[c]{@{}c@{}}SSA\\ Mean\\ (Std)\end{tabular} & \begin{tabular}[c]{@{}c@{}}OFA\\ Mean\\ (Std)\end{tabular} & \begin{tabular}[c]{@{}c@{}}SHADE\\ Mean\\ (Std)\end{tabular} & \begin{tabular}[c]{@{}c@{}}TFSSA\\ Mean\\ (Std)\end{tabular} \\\hline
		CEC2020\_F1 & \begin{tabular}[c]{@{}c@{}}8.2541e+8 \\ (2.51e+8) +\end{tabular} & \begin{tabular}[c]{@{}c@{}}6.1650e+9 \\ (2.74e+9) +\end{tabular} & \begin{tabular}[c]{@{}c@{}}2.5690e+9\\  (1.70e+9) +\end{tabular} & \begin{tabular}[c]{@{}c@{}}2.4253e+9\\  (6.48e+8) +\end{tabular} & \begin{tabular}[c]{@{}c@{}}1.8891e+9 \\ (1.08e+9) -\end{tabular} & \begin{tabular}[c]{@{}c@{}}3.2793e+9 \\ (9.17e+8) +\end{tabular} & \begin{tabular}[c]{@{}c@{}}8.1623e+6\\  (2.73e+6) +\end{tabular} & \textbf{\begin{tabular}[c]{@{}c@{}}3.2993e+6\\  (1.14e+6)\end{tabular}} \\
		CEC2020\_F2 & \begin{tabular}[c]{@{}c@{}}5.7675e+3\\  (3.20e+2) +\end{tabular} & \begin{tabular}[c]{@{}c@{}}3.9708e+3 \\ (4.21e+2) -\end{tabular} & \begin{tabular}[c]{@{}c@{}}3.7089e+3\\  (4.74e+2) -\end{tabular} & \begin{tabular}[c]{@{}c@{}}5.5763e+3 \\ \textbf{(1.98e+2)} =\end{tabular} & \begin{tabular}[c]{@{}c@{}}2.7448e+3 \\ (3.21e+2) +\end{tabular} & \begin{tabular}[c]{@{}c@{}}5.6418e+3 \\ (2.89e+2) +\end{tabular} & \begin{tabular}[c]{@{}c@{}}4.4956e+3\\  (2.84e+2) -\end{tabular} & \begin{tabular}[c]{@{}c@{}}1.5610e+3\\  (2.27e+2)\end{tabular} \\
		CEC2020\_F3 & \begin{tabular}[c]{@{}c@{}}9.5314e+2\\  (2.25e+1) -\end{tabular} & \begin{tabular}[c]{@{}c@{}}9.2851e+2 \\ (3.56e+1) +\end{tabular} & \begin{tabular}[c]{@{}c@{}}8.1250e+2 \\ (1.74e+1) -\end{tabular} & \begin{tabular}[c]{@{}c@{}}9.6884e+2\\  (1.69e+1) +\end{tabular} & \begin{tabular}[c]{@{}c@{}}9.5392e+2 \\ (7.45e+1) +\end{tabular} & \begin{tabular}[c]{@{}c@{}}9.7358e+2 \\ (2.29e+1) -\end{tabular} & \begin{tabular}[c]{@{}c@{}}8.2764e+2 \\ \textbf{(7.51e+0)} +\end{tabular} & \begin{tabular}[c]{@{}c@{}}\textbf{7.4445e+2} \\ (8.55e+0)\end{tabular} \\
		CEC2020\_F4 & \begin{tabular}[c]{@{}c@{}}2.0108e+3 \\ (5.50e+1) -\end{tabular} & \begin{tabular}[c]{@{}c@{}}1.8210e+4\\  (3.72e+4) -\end{tabular} & \begin{tabular}[c]{@{}c@{}}4.9532e+3 \\ (7.09e+3) -\end{tabular} & \begin{tabular}[c]{@{}c@{}}2.4562e+3 \\ (4.35e+2) -\end{tabular} & \begin{tabular}[c]{@{}c@{}}3.8546e+3\\  (3.45e+3) -\end{tabular} & \begin{tabular}[c]{@{}c@{}}3.3717e+3 \\ (9.53e+2) +\end{tabular} & \begin{tabular}[c]{@{}c@{}}1.9107e+3 \\ \textbf{(1.03e+0)} -\end{tabular} & \begin{tabular}[c]{@{}c@{}}\textbf{1.9041e+3} \\ (1.07e+0)\end{tabular} \\
		CEC2020\_F5 & \begin{tabular}[c]{@{}c@{}}5.8713e+6 \\ (3.06e+6) +\end{tabular} & \begin{tabular}[c]{@{}c@{}}1.9239e+6 \\ (1.82e+6) =\end{tabular} & \begin{tabular}[c]{@{}c@{}}\textbf{2.6514e+4} \\ (2.36e+4) +\end{tabular} & \begin{tabular}[c]{@{}c@{}}2.8432e+5 \\ (8.82e+4) +\end{tabular} & \begin{tabular}[c]{@{}c@{}}1.6705e+6\\  (1.85e+6) =\end{tabular} & \begin{tabular}[c]{@{}c@{}}1.5941e+6 \\ (8.94e+5) =\end{tabular} & \begin{tabular}[c]{@{}c@{}}4.6806e+4 \\ \textbf{(2.20e+4)} +\end{tabular} & \begin{tabular}[c]{@{}c@{}}1.1683e+6 \\ (1.04e+6)\end{tabular} \\
		CEC2020\_F6 & \begin{tabular}[c]{@{}c@{}}2.1056e+3\\  (1.34e+2) -\end{tabular} & \begin{tabular}[c]{@{}c@{}}2.4993e+3 \\ (2.56e+2) +\end{tabular} & \begin{tabular}[c]{@{}c@{}}2.0628e+3 \\ (1.78e+2) -\end{tabular} & \begin{tabular}[c]{@{}c@{}}2.4975e+3 \\ (1.14e+2) -\end{tabular} & \begin{tabular}[c]{@{}c@{}}1.9359e+3 \\ (1.10e+2) -\end{tabular} & \begin{tabular}[c]{@{}c@{}}2.6072e+3 \\ (2.13e+2) -\end{tabular} & \begin{tabular}[c]{@{}c@{}}1.8690e+3 \\ (6.26e+1) -\end{tabular} & \textbf{\begin{tabular}[c]{@{}c@{}}1.6400e+3\\  (5.22e+1)\end{tabular}} \\
		CEC2020\_F7 & \begin{tabular}[c]{@{}c@{}}9.5712e+5 \\ (7.41e+5) =\end{tabular} & \begin{tabular}[c]{@{}c@{}}6.8662e+5\\  (1.23e+6) =\end{tabular} & \begin{tabular}[c]{@{}c@{}}\textbf{8.0781e+3} \\ (7.77e+3) +\end{tabular} & \begin{tabular}[c]{@{}c@{}}3.1638e+4\\  (1.46e+4) +\end{tabular} & \begin{tabular}[c]{@{}c@{}}7.7198e+5 \\ (7.98e+5) =\end{tabular} & \begin{tabular}[c]{@{}c@{}}4.9021e+5 \\ (2.77e+5) =\end{tabular} & \begin{tabular}[c]{@{}c@{}}8.3634e+3\\  \textbf{(2.10e+3)} +\end{tabular} & \begin{tabular}[c]{@{}c@{}}8.0619e+5 \\ (9.61e+5)\end{tabular} \\
		CEC2020\_F8 & \begin{tabular}[c]{@{}c@{}}2.5147e+3 \\ (4.22e+1) -\end{tabular} & \begin{tabular}[c]{@{}c@{}}3.7116e+3 \\ (9.12e+2) -\end{tabular} & \begin{tabular}[c]{@{}c@{}}2.6301e+3 \\ (1.80e+2) -\end{tabular} & \begin{tabular}[c]{@{}c@{}}3.5211e+3 \\ (3.88e+2) +\end{tabular} & \begin{tabular}[c]{@{}c@{}}3.1946e+3 \\ (9.62e+2) -\end{tabular} & \begin{tabular}[c]{@{}c@{}}3.1260e+3 \\ (2.32e+2) -\end{tabular} & \begin{tabular}[c]{@{}c@{}}2.3229e+3 \\ (1.01e+1) -\end{tabular} & \textbf{\begin{tabular}[c]{@{}c@{}}2.3127e+3 \\ (5.46e-1)\end{tabular}} \\
		CEC2020\_F9 & \begin{tabular}[c]{@{}c@{}}2.9444e+3\\  (1.10e+1) +\end{tabular} & \textbf{\begin{tabular}[c]{@{}c@{}}2.8361e+3 \\ (9.71e+0) +\end{tabular}} & \begin{tabular}[c]{@{}c@{}}2.9084e+3\\  (2.37e+1) +\end{tabular} & \begin{tabular}[c]{@{}c@{}}2.9937e+3 \\ (2.12e+1) =\end{tabular} & \begin{tabular}[c]{@{}c@{}}2.9290e+3 \\ (3.21e+1) +\end{tabular} & \begin{tabular}[c]{@{}c@{}}3.0541e+3\\  (3.02e+1) +\end{tabular} & \begin{tabular}[c]{@{}c@{}}2.9137e+3 \\ (9.92e+0) +\end{tabular} & \begin{tabular}[c]{@{}c@{}}3.1352e+3\\  (1.02e+2)\end{tabular} \\
		CEC2020\_F10 & \begin{tabular}[c]{@{}c@{}}3.0418e+3 \\ (4.50e+1) +\end{tabular} & \begin{tabular}[c]{@{}c@{}}3.3440e+3 \\ (1.50e+2) +\end{tabular} & \begin{tabular}[c]{@{}c@{}}3.0693e+3 \\ (5.39e+1) -\end{tabular} & \begin{tabular}[c]{@{}c@{}}3.1881e+3 \\ (7.56e+1) -\end{tabular} & \begin{tabular}[c]{@{}c@{}}3.0981e+3\\  (7.90e+1) +\end{tabular} & \begin{tabular}[c]{@{}c@{}}3.2489e+3\\  (8.23e+1) +\end{tabular} & \begin{tabular}[c]{@{}c@{}}\textbf{2.9166e+3} \\ (1.40e+0) +\end{tabular} & \begin{tabular}[c]{@{}c@{}}2.9607e+3 \\ \textbf{(3.37e+1)}\end{tabular} \\
		+/-/= & 5/4/1 & 5/3/2 & 4/6/0 & 5/3/2 & 4/4/2 & 5/3/2 & 6/4/0 &   \\\hline 
	\end{tabular}
\end{sidewaystable}
\subsection{Performance of proposed model}

\subsubsection{Description of data}
The utility and strength of our suggested strategy will be thoroughly investigated by selecting features from well-known datasets. Twenty-one datasets are from the UCI machine learning repository \cite{asuncion2007uci} and can be accessed online\renewcommand{\thefootnote}{1}
\footnote{\url{https://www.openml.org/search}}. Table \ref{T1} gives a brief summary of the datasets used. The amount of features (\#Feat), samples (\#SMP), classes (\#CL), and the area to which each dataset belongs are all provided for each dataset.
\subsubsection{Parameter configuration}
Several top-of-the-line and most recent FS techniques are contrasted with the suggested approach, which is summarized as follows: 
\begin{itemize}
    \item Genetic Algorithm (GA) \cite{holland1992genetic}.
    \item Dragonfly Algorithm (DA) \cite{mirjalili2016dragonfly}.
    \item Ant Lion Optimizer (ALO) \cite{mirjalili2015ant}.
    \item Sparrow Search Algorithm (SSA) \cite{xue2020novel}.
    \item Sine Cosine Algorithm (SCA) \cite{mirjalili2016sca}.
    \item Particle Swarm Optimizer (PSO) \cite{kennedy1995particle}.
    \item binary Butterfly Optimization Algorithm (bBOA) \cite{arora2019binary}.
    \item Brain Storm Optimizer (BSO) \cite{shi2011brain}.
    \item Improved Sparrow Search Algorithm (ISSA) \cite{yuan2021dmppt}.
    \item Grey Wolf Optimizer (GWO) \cite{mirjalili2014grey}.
\end{itemize}


\begin{figure}[ht]
    \centering
    \includegraphics[width=0.48\textwidth]{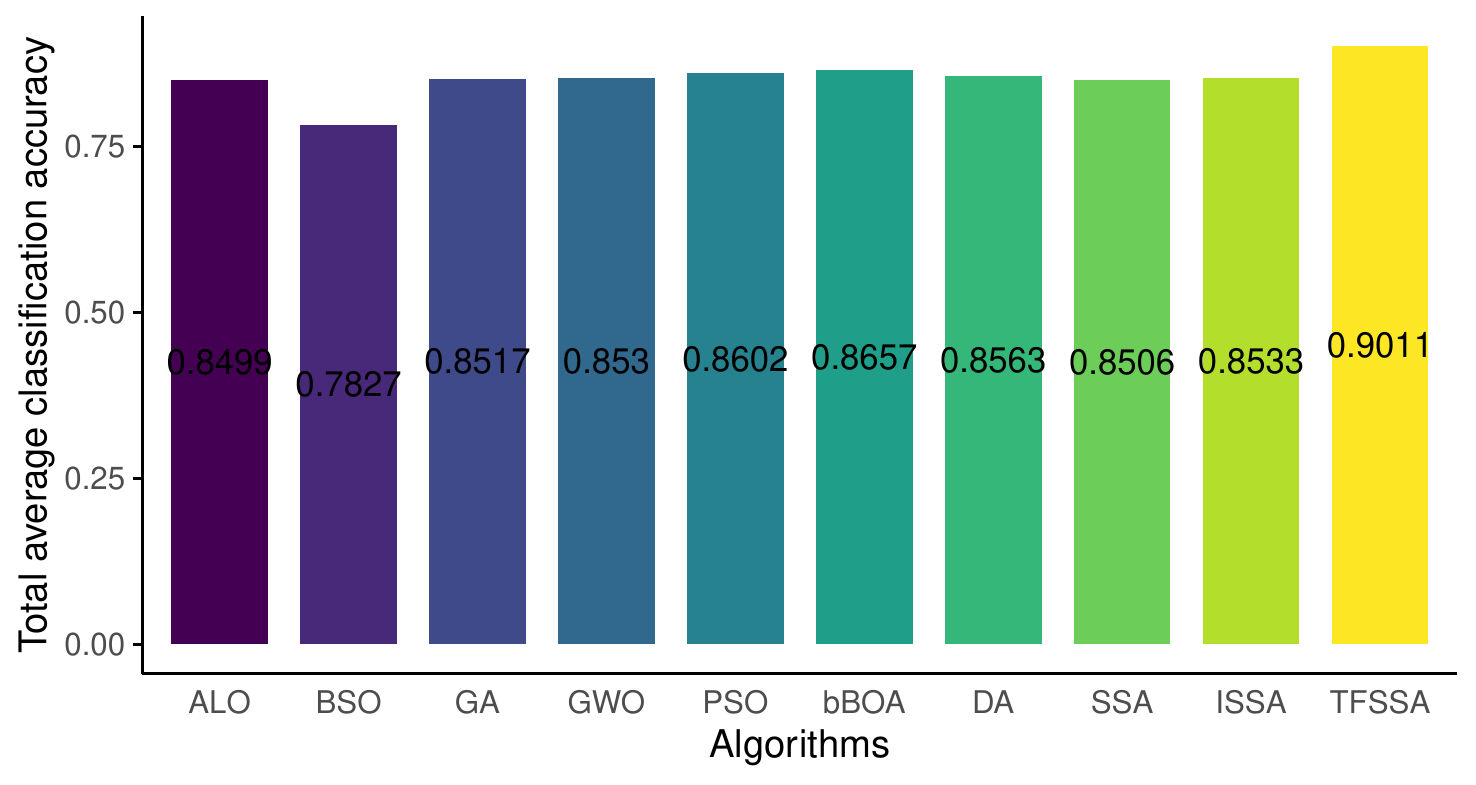}
    \caption{The average classification accuracy selected by the algorithms.}
    \label{fig:avg acc}
\end{figure}

Each algorithm is run 20 times with a different seed in each run.
For all subsequent tests, the maximum number of repetitions is set at 100. In the population, there are 7 search agents. In addition, 10-fold cross-validation is used. Table \ref{T2} shows the global and algorithm-specific parameter settings.
To ensure a fair comparison of the algorithms, the parameters of the algorithms are gathered from the literature. The main purpose of this research is to evaluate the performance of numerous FS methods to the proposed methodology. The K-NN classifier is a popular wrapper approach for FS.  When K=5, the method produces superior results. 10-fold cross-validation is recommended for real-world datasets as in \cite{kohavi1995study} and \cite{hastie2009elements}. 

 \subsubsection{Evaluation criteria}

Individual datasets are separated into three equal parts at random: training, testing, and validation datasets. The data is partitioned 20 times to verify that the results are stable and statistically significant. The following assurances are captured from the validation data for each run:

 \begin{table*}[]
\begin{center}
\small
\caption{Experiment parameter configuration.}\label{T2}
\begin{tabular}{ll} 
\hline Parameter description & Value $(\mathrm{s})$ \\
\hline a parameter in Tent chaos & $0.7$ \\
$\alpha$ parameter in Lévy flights & $1.5$\\
$\lambda$ parameter in $Fitness$ & $0.99$\\
$\mu$ parameter in $Fitness$ & $0.01$\\
Count of runs(M) & 20 \\
The amount of search agents & 7 \\
The amount of T\_max & 100 \\
Problem Dimensions & No. of features in each datasets \\
K for cross-validation & 10 \\
Search field & $\{0,1\}$ \\
GA crossover ratio & $0.9$ \\
GA mutation ratio & $0.1$ \\
Selection strategy in GA & Roulette wheel \\
A factors in WOA & {$[0, 2]$} \\
Acceleration factors in PSO & {$[2,2]$} \\
Inertia index(w) in PSO & {$[0.9,0.6]$} \\
A factors in GWO & $\{0,2\}$ \\
Mutation rate $\mathrm{r}$ in ALO & {$[0, 0.9]$} \\
Parameter(a) in bBOA & $0.1$ \\
Parameter(c) in bBOA & {$[0.01, 0.25]$} \\
The amount of clusters in BSO & 5 \\
\hline
\end{tabular}
\end{center}
\end{table*}

\begin{table*}[]
 \centering
    \small
\caption{Comparison of the classification accuracy of each algorithm.}\label{Tacc}

    \begin{threeparttable}
   \begin{tabular}{llllllllllll}
\hline No. & Datasets & ALO & BSO & GA & GWO & PSO & bBOA & DA & SSA & ISSA & TFSSA \\\hline
1 & BreastCO & 0.9591 & 0.9200 & 0.9597 & 0.9603 & 0.9609 & 0.9286 & 0.9626 & 0.9600 & 0.9611 & \textbf{0.9668} \\
2 & BreastCWD & 0.9392 & 0.9020 & 0.9488 & 0.9375 & 0.9385 & 0.9396 & 0.9385 & 0.9347 & 0.9396 & \textbf{0.9718} \\
3 & Clean-1 & 0.8465 & 0.8261 & 0.8697 & 0.8580 & 0.8549 & 0.8562 & 0.8541 & 0.8431 & 0.8585 & \textbf{0.8923} \\
4 & Clean-2 & 0.9496 & 0.9391 & 0.9423 & 0.9463 & 0.9465 & 0.9480 & 0.9487 & 0.9462 & 0.9510 & \textbf{0.9667} \\
5 & CongressVR & 0.9370 & 0.8547 & 0.9413 & 0.9327 & 0.9235 & 0.9280 & 0.9318 & 0.9321 & 0.9349 & \textbf{0.9521} \\
6 & Exactly-1 & 0.7061 & 0.6021 & 0.7306 & 0.7249 & 0.7471 & \textbf{0.8531} & 0.7481 & 0.7091 & 0.7197 & 0.8524 \\
7 & Exactly-2 & 0.6980 & 0.6345 & 0.6940 & 0.6929 & 0.6959 & 0.6527 & 0.7007 & 0.6985 & 0.6977 & \textbf{0.7472} \\
8 & StatlogH & 0.7773 & 0.6948 & 0.7867 & 0.7768 & 0.7788 & 0.7583 & 0.7773 & 0.7595 & 0.7842 & \textbf{0.8127} \\
9 & IonosphereVS & 0.8595 & 0.8538 & 0.8938 & 0.8682 & 0.8485 & 0.8639 & 0.8708 & 0.8890 & 0.8826 & \textbf{0.9042} \\
10 & KrvskpEW & 0.9006 & 0.7603 & 0.9215 & 0.9143 & 0.9200 & 0.8580 & 0.9269 & 0.8929 & 0.8980 & \textbf{0.9360} \\
11 & Lymphography & 0.7863 & 0.6931 & 0.8164 & 0.7629 & 0.7906 & 0.8613 & 0.7793 & 0.7736 & 0.7880 & \textbf{0.8667} \\
12 & M-of-n & 0.8184 & 0.7033 & 0.7988 & 0.8272 & 0.8425 & 0.8689 & 0.8293 & 0.8361 & 0.8549 & \textbf{0.9020} \\
13 & Penglung & 0.8072 & 0.7676 & 0.6721 & 0.8341 & 0.8140 & 0.8482 & 0.8268 & 0.8331 & 0.7951 & \textbf{0.8745} \\
14 & Semeion & 0.9584 & 0.9461 & 0.9557 & 0.9471 & 0.9476 & 0.9480 & 0.9521 & 0.9449 & 0.9504 & \textbf{0.9729} \\
15 & SonarMR & 0.8487 & 0.7936 & \textbf{0.8750} & 0.8622 & 0.8667 & 0.8614 & 0.8506 & 0.8449 & 0.8506 & 0.8634 \\
16 & Spectheart & 0.7881 & 0.7507 & 0.8097 & 0.7846 & 0.7841 & 0.7643 & 0.8000 & 0.7826 & 0.7871 & \textbf{0.8443} \\
17 & 3T Endgame & 0.7587 & 0.6601 & 0.7609 & 0.7537 & 0.8622 & 0.8667 & 0.7564 & 0.7557 & 0.7546 & \textbf{0.8983} \\
18 & Vote & 0.9258 & 0.8413 & 0.9333 & 0.9196 & 0.9258 & 0.9618 & 0.9227 & 0.9196 & 0.9200 & \textbf{0.9695} \\
19 & WaveformV2 & 0.7066 & 0.6150 & 0.6921 & 0.7096 & 0.7192 & 0.7827 & 0.7154 & 0.7091 & 0.7044 & \textbf{0.7929} \\
20 & Wine & 0.9543 & 0.8652 & 0.9536 & 0.9476 & 0.9521 & 0.9474 & 0.9551 & 0.9506 & 0.9566 & \textbf{0.9843} \\
21 & Zoology & 0.9216 & 0.8131 & 0.9294 & \textbf{0.9525} & 0.9451 & 0.8827 & 0.9359 & 0.9476 & 0.9307 & \textbf{0.9525} \\
 & \textbf{AVG.} & 0.8499 & 0.7827 & 0.8517 & 0.8530 & 0.8602 & 0.8657 & 0.8563 & 0.8506 & 0.8533 & 0.9011\\\hline

\end{tabular}
\begin{tablenotes} 
		\item The \textbf{bolded} values represent the best outcomes.
     \end{tablenotes} 
\end{threeparttable}
\end{table*}

\begin{table*}
\caption{Comparison of the selected average No. of features(AVG.NOF.) and the selected feature ratio(AVG\_Ri) of each algorithm.}\label{Tfea}
\centering
\small
\begin{threeparttable}
\scalebox{0.8}{
\begin{tabular}{llllllllllll}
\hline \multicolumn{1}{c}{\multirow{3}{*}{No.}} & \multicolumn{1}{c}{\multirow{3}{*}{Dataset}} & \multicolumn{1}{c}{ALO} & \multicolumn{1}{c}{BSO} & \multicolumn{1}{c}{GA} & \multicolumn{1}{c}{GWO} & \multicolumn{1}{c}{PSO} & \multicolumn{1}{c}{bBOA} & \multicolumn{1}{c}{DA} & \multicolumn{1}{c}{SSA} & \multicolumn{1}{c}{ISSA} & \multicolumn{1}{c}{TFSSA} \\
\cmidrule(l){3-3}\cmidrule(l){4-4}\cmidrule(l){5-5}\cmidrule(l){6-6}\cmidrule(l){7-7}\cmidrule(l){8-8}\cmidrule(l){9-9}\cmidrule(l){10-10}\cmidrule(l){11-11}\cmidrule(l){12-12}
\multicolumn{1}{c}{} & \multicolumn{1}{c}{} & \multicolumn{1}{c}{AVG.NOF.} & \multicolumn{1}{c}{AVG.NOF.} & \multicolumn{1}{c}{AVG.NOF.} & \multicolumn{1}{c}{AVG.NOF.} & \multicolumn{1}{c}{AVG.NOF.} & \multicolumn{1}{c}{AVG.NOF.} & \multicolumn{1}{c}{AVG.NOF.} & \multicolumn{1}{c}{AVG.NOF.} & \multicolumn{1}{c}{AVG.NOF.} & \multicolumn{1}{c}{AVG.NOF.} \\
\multicolumn{1}{c}{} & \multicolumn{1}{c}{} & \multicolumn{1}{c}{(AVG\_Ri.)} & \multicolumn{1}{c}{(AVG\_Ri.)} & \multicolumn{1}{c}{(AVG\_Ri.)} & \multicolumn{1}{c}{(AVG\_Ri.)} & \multicolumn{1}{c}{(AVG\_Ri.)} & \multicolumn{1}{c}{(AVG\_Ri.)} & \multicolumn{1}{c}{(AVG\_Ri.)} & \multicolumn{1}{c}{(AVG\_Ri.)} & \multicolumn{1}{c}{(AVG\_Ri.)} & \multicolumn{1}{c}{(AVG\_Ri.)}\\
  \hline

1 &
  BreastCO &
  \begin{tabular}[c]{@{}c@{}}7.00\\ (0.778)\end{tabular} &
  \begin{tabular}[c]{@{}c@{}}6.40\\ (0.711)\end{tabular} &
  \begin{tabular}[c]{@{}c@{}}6.10\\ (0.678)\end{tabular} &
  \begin{tabular}[c]{@{}c@{}}6.90\\ (0.767)\end{tabular} &
  \begin{tabular}[c]{@{}c@{}}5.70\\ (0.633)\end{tabular} &
  \begin{tabular}[c]{@{}c@{}}5.60\\ (0.622)\end{tabular} &
  \begin{tabular}[c]{@{}c@{}}6.27\\ (0.697)\end{tabular} &
  \begin{tabular}[c]{@{}c@{}}7.20\\ (0.800)\end{tabular} &
  \begin{tabular}[c]{@{}c@{}}5.70\\ (0.633)\end{tabular} &
  \textbf{\begin{tabular}[c]{@{}c@{}}4.40\\ (0.489)\end{tabular}} \\
2 &
  BreastCWD &
  \begin{tabular}[c]{@{}c@{}}24.27\\ (0.809)\end{tabular} &
  \begin{tabular}[c]{@{}c@{}}13.73\\ (0.458)\end{tabular} &
  \begin{tabular}[c]{@{}c@{}}12.20\\ (0.407)\end{tabular} &
  \begin{tabular}[c]{@{}c@{}}19.00\\ (0.633)\end{tabular} &
  \begin{tabular}[c]{@{}c@{}}18.33\\ (0.611)\end{tabular} &
  \begin{tabular}[c]{@{}c@{}}16.80\\ (0.560)\end{tabular} &
  \begin{tabular}[c]{@{}c@{}}20.00\\ (0.667)\end{tabular} &
  \begin{tabular}[c]{@{}c@{}}18.27\\ (0.609)\end{tabular} &
  \begin{tabular}[c]{@{}c@{}}20.47\\ (0.682)\end{tabular} &
  \textbf{\begin{tabular}[c]{@{}c@{}}8.40\\ (0.280)\end{tabular}} \\
3 &
  Clean-1 &
  \begin{tabular}[c]{@{}c@{}}132.00\\ (0.795)\end{tabular} &
  \begin{tabular}[c]{@{}c@{}}98.73\\ (0.595)\end{tabular} &
  \begin{tabular}[c]{@{}c@{}}98.90\\ (0.596)\end{tabular} &
  \begin{tabular}[c]{@{}c@{}}109.60\\ (0.660)\end{tabular} &
  \begin{tabular}[c]{@{}c@{}}104.93\\ (0.632)\end{tabular} &
  \begin{tabular}[c]{@{}c@{}}91.80\\ (0.553)\end{tabular} &
  \begin{tabular}[c]{@{}c@{}}109.67\\ (0.661)\end{tabular} &
  \begin{tabular}[c]{@{}c@{}}94.87\\ (0.572)\end{tabular} &
  \textbf{\begin{tabular}[c]{@{}c@{}}90.20\\ (0.543)\end{tabular}} &
  \begin{tabular}[c]{@{}c@{}}90.27\\ (0.544)\end{tabular} \\
4 &
  Clean-2 &
  \begin{tabular}[c]{@{}c@{}}95.00\\ (0.572)\end{tabular} &
  \begin{tabular}[c]{@{}c@{}}101.00\\ (0.608)\end{tabular} &
  \begin{tabular}[c]{@{}c@{}}94.10\\ (0.567)\end{tabular} &
  \begin{tabular}[c]{@{}c@{}}106.00\\ (0.639)\end{tabular} &
  \begin{tabular}[c]{@{}c@{}}109.40\\ (0.659)\end{tabular} &
  \begin{tabular}[c]{@{}c@{}}92.40\\ (0.557)\end{tabular} &
  \begin{tabular}[c]{@{}c@{}}100.40\\ (0.605)\end{tabular} &
  \begin{tabular}[c]{@{}c@{}}90.40\\ (0.545)\end{tabular} &
  \begin{tabular}[c]{@{}c@{}}92.40\\ (0.557)\end{tabular} &
  \textbf{\begin{tabular}[c]{@{}c@{}}90.28\\ (0.544)\end{tabular}} \\
5 &
  CongressVR &
  \begin{tabular}[c]{@{}c@{}}9.87\\ (0.617)\end{tabular} &
  \begin{tabular}[c]{@{}c@{}}7.53\\ (0.471)\end{tabular} &
  \begin{tabular}[c]{@{}c@{}}7.10\\ (0.444)\end{tabular} &
  \begin{tabular}[c]{@{}c@{}}9.80\\ (0.613)\end{tabular} &
  \begin{tabular}[c]{@{}c@{}}10.80\\ (0.675)\end{tabular} &
  \begin{tabular}[c]{@{}c@{}}6.40\\ (0.400)\end{tabular} &
  \begin{tabular}[c]{@{}c@{}}10.87\\ (0.679)\end{tabular} &
  \begin{tabular}[c]{@{}c@{}}8.40\\ (0.525)\end{tabular} &
  \begin{tabular}[c]{@{}c@{}}9.00\\ (0.563)\end{tabular} &
  \textbf{\begin{tabular}[c]{@{}c@{}}6.15\\ (0.384)\end{tabular}} \\
6 &
  Exactly-1 &
  \begin{tabular}[c]{@{}c@{}}12.87\\ (0.990)\end{tabular} &
  \begin{tabular}[c]{@{}c@{}}7.73\\ (0.595)\end{tabular} &
  \begin{tabular}[c]{@{}c@{}}8.10\\ (0.623)\end{tabular} &
  \begin{tabular}[c]{@{}c@{}}12.07\\ (0.928)\end{tabular} &
  \begin{tabular}[c]{@{}c@{}}9.00\\ (0.692)\end{tabular} &
  \begin{tabular}[c]{@{}c@{}}7.60\\ (0.585)\end{tabular} &
  \begin{tabular}[c]{@{}c@{}}10.53\\ (0.810)\end{tabular} &
  \begin{tabular}[c]{@{}c@{}}12.80\\ (0.985)\end{tabular} &
  \begin{tabular}[c]{@{}c@{}}10.47\\ (0.805)\end{tabular} &
  \textbf{\begin{tabular}[c]{@{}c@{}}6.48\\ (0.498)\end{tabular}} \\
7 &
  Exactly-2 &
  \begin{tabular}[c]{@{}c@{}}8.40\\ (0.646)\end{tabular} &
  \begin{tabular}[c]{@{}c@{}}6.27\\ (0.482)\end{tabular} &
  \begin{tabular}[c]{@{}c@{}}7.10\\ (0.546)\end{tabular} &
  \begin{tabular}[c]{@{}c@{}}7.53\\ (0.579)\end{tabular} &
  \begin{tabular}[c]{@{}c@{}}9.40\\ (0.723)\end{tabular} &
  \begin{tabular}[c]{@{}c@{}}4.80\\ (0.369)\end{tabular} &
  \begin{tabular}[c]{@{}c@{}}8.67\\ (0.667)\end{tabular} &
  \begin{tabular}[c]{@{}c@{}}6.27\\ (0.482)\end{tabular} &
  \begin{tabular}[c]{@{}c@{}}9.00\\ (0.692)\end{tabular} &
  \textbf{\begin{tabular}[c]{@{}c@{}}4.62\\ (0.355)\end{tabular}} \\
8 &
  StatlogH &
  \begin{tabular}[c]{@{}c@{}}10.40\\ (0.800)\end{tabular} &
  \begin{tabular}[c]{@{}c@{}}6.60\\ (0.508)\end{tabular} &
  \begin{tabular}[c]{@{}c@{}}6.60\\ (0.508)\end{tabular} &
  \begin{tabular}[c]{@{}c@{}}8.80\\ (0.677)\end{tabular} &
  \begin{tabular}[c]{@{}c@{}}9.07\\ (0.698)\end{tabular} &
  \begin{tabular}[c]{@{}c@{}}5.80\\ (0.446)\end{tabular} &
  \begin{tabular}[c]{@{}c@{}}9.60\\ (0.738)\end{tabular} &
  \begin{tabular}[c]{@{}c@{}}7.47\\ (0.575)\end{tabular} &
  \begin{tabular}[c]{@{}c@{}}8.47\\ (0.652)\end{tabular} &
  \textbf{\begin{tabular}[c]{@{}c@{}}4.86\\ (0.374)\end{tabular}} \\
9 &
  IonosphereVS &
  \begin{tabular}[c]{@{}c@{}}20.13\\ (0.592)\end{tabular} &
  \begin{tabular}[c]{@{}c@{}}15.93\\ (0.469)\end{tabular} &
  \textbf{\begin{tabular}[c]{@{}c@{}}13.50\\ (0.397)\end{tabular}} &
  \begin{tabular}[c]{@{}c@{}}17.33\\ (0.510)\end{tabular} &
  \begin{tabular}[c]{@{}c@{}}19.20\\ (0.565)\end{tabular} &
  \begin{tabular}[c]{@{}c@{}}16.20\\ (0.476)\end{tabular} &
  \begin{tabular}[c]{@{}c@{}}18.00\\ (0.529)\end{tabular} &
  \begin{tabular}[c]{@{}c@{}}19.67\\ (0.579)\end{tabular} &
  \begin{tabular}[c]{@{}c@{}}19.07\\ (0.561)\end{tabular} &
  \begin{tabular}[c]{@{}c@{}}17.14\\ (0.504)\end{tabular} \\
10 &
  KrvskpEW &
  \begin{tabular}[c]{@{}c@{}}35.80\\ (0.994)\end{tabular} &
  \begin{tabular}[c]{@{}c@{}}17.80\\ (0.494)\end{tabular} &
  \begin{tabular}[c]{@{}c@{}}18.00\\ (0.500)\end{tabular} &
  \begin{tabular}[c]{@{}c@{}}31.60\\ (0.878)\end{tabular} &
  \begin{tabular}[c]{@{}c@{}}25.60\\ (0.711)\end{tabular} &
  \begin{tabular}[c]{@{}c@{}}17.60\\ (0.489)\end{tabular} &
  \begin{tabular}[c]{@{}c@{}}28.60\\ (0.794)\end{tabular} &
  \begin{tabular}[c]{@{}c@{}}29.40\\ (0.817)\end{tabular} &
  \begin{tabular}[c]{@{}c@{}}20.80\\ (0.578)\end{tabular} &
  \textbf{\begin{tabular}[c]{@{}c@{}}16.91\\ (0.470)\end{tabular}} \\
11 &
  Lymphography &
  \begin{tabular}[c]{@{}c@{}}13.33\\ (0.741)\end{tabular} &
  \begin{tabular}[c]{@{}c@{}}9.47\\ (0.526)\end{tabular} &
  \begin{tabular}[c]{@{}c@{}}8.90\\ (0.494)\end{tabular} &
  \begin{tabular}[c]{@{}c@{}}11.80\\ (0.656)\end{tabular} &
  \begin{tabular}[c]{@{}c@{}}11.73\\ (0.652)\end{tabular} &
  \textbf{\begin{tabular}[c]{@{}c@{}}8.40\\ (0.467)\end{tabular}} &
  \begin{tabular}[c]{@{}c@{}}12.53\\ (0.696)\end{tabular} &
  \begin{tabular}[c]{@{}c@{}}12.20\\ (0.678)\end{tabular} &
  \begin{tabular}[c]{@{}c@{}}8.87\\ (0.493)\end{tabular} &
  \begin{tabular}[c]{@{}c@{}}9.17\\ (0.509)\end{tabular} \\
12 &
  M-of-n &
  \begin{tabular}[c]{@{}c@{}}11.27\\ (0.867)\end{tabular} &
  \begin{tabular}[c]{@{}c@{}}6.90\\ (0.531)\end{tabular} &
  \begin{tabular}[c]{@{}c@{}}7.68\\ (0.591)\end{tabular} &
  \begin{tabular}[c]{@{}c@{}}11.27\\ (0.867)\end{tabular} &
  \begin{tabular}[c]{@{}c@{}}10.87\\ (0.836)\end{tabular} &
  \begin{tabular}[c]{@{}c@{}}6.80\\ (0.523)\end{tabular} &
  \begin{tabular}[c]{@{}c@{}}12.13\\ (0.933)\end{tabular} &
  \begin{tabular}[c]{@{}c@{}}12.33\\ (0.948)\end{tabular} &
  \begin{tabular}[c]{@{}c@{}}10.67\\ (0.821)\end{tabular} &
  \textbf{\begin{tabular}[c]{@{}c@{}}6.30\\ (0.485)\end{tabular}} \\
13 &
  Penglung &
  \begin{tabular}[c]{@{}c@{}}172.07\\ (0.529)\end{tabular} &
  \begin{tabular}[c]{@{}c@{}}160.60\\ (0.494)\end{tabular} &
  \textbf{\begin{tabular}[c]{@{}c@{}}153.00\\ (0.471)\end{tabular}} &
  \begin{tabular}[c]{@{}c@{}}162.80\\ (0.501)\end{tabular} &
  \begin{tabular}[c]{@{}c@{}}183.33\\ (0.564)\end{tabular} &
  \begin{tabular}[c]{@{}c@{}}172.00\\ (0.529)\end{tabular} &
  \begin{tabular}[c]{@{}c@{}}175.20\\ (0.539)\end{tabular} &
  \begin{tabular}[c]{@{}c@{}}162.33\\ (0.499)\end{tabular} &
  \begin{tabular}[c]{@{}c@{}}182.67\\ (0.562)\end{tabular} &
  \begin{tabular}[c]{@{}c@{}}161.42\\ (0.497)\end{tabular} \\
14 &
  Semeion &
  \begin{tabular}[c]{@{}c@{}}187.80\\ (0.709)\end{tabular} &
  \begin{tabular}[c]{@{}c@{}}162.00\\ (0.611)\end{tabular} &
  \begin{tabular}[c]{@{}c@{}}149.40\\ (0.564)\end{tabular} &
  \begin{tabular}[c]{@{}c@{}}203.60\\ (0.768)\end{tabular} &
  \begin{tabular}[c]{@{}c@{}}171.60\\ (0.648)\end{tabular} &
  \begin{tabular}[c]{@{}c@{}}143.20\\ (0.540)\end{tabular} &
  \begin{tabular}[c]{@{}c@{}}193.00\\ (0.728)\end{tabular} &
  \begin{tabular}[c]{@{}c@{}}161.80\\ (0.611)\end{tabular} &
  \begin{tabular}[c]{@{}c@{}}194.40\\ (0.734)\end{tabular} &
  \textbf{\begin{tabular}[c]{@{}c@{}}142.38\\ (0.537)\end{tabular}} \\
15 &
  SonarMR &
  \begin{tabular}[c]{@{}c@{}}48.00\\ (0.800)\end{tabular} &
  \begin{tabular}[c]{@{}c@{}}30.60\\ (0.510)\end{tabular} &
  \begin{tabular}[c]{@{}c@{}}30.30\\ (0.505)\end{tabular} &
  \begin{tabular}[c]{@{}c@{}}41.60\\ (0.693)\end{tabular} &
  \begin{tabular}[c]{@{}c@{}}37.60\\ (0.627)\end{tabular} &
  \begin{tabular}[c]{@{}c@{}}32.80\\ (0.547)\end{tabular} &
  \begin{tabular}[c]{@{}c@{}}29.40\\ (0.490)\end{tabular} &
  \begin{tabular}[c]{@{}c@{}}34.13\\ (0.569)\end{tabular} &
  \begin{tabular}[c]{@{}c@{}}37.13\\ (0.619)\end{tabular} &
  \textbf{\begin{tabular}[c]{@{}c@{}}22.36\\ (0.373)\end{tabular}} \\
16 &
  Spectheart &
  \begin{tabular}[c]{@{}c@{}}13.87\\ (0.630)\end{tabular} &
  \begin{tabular}[c]{@{}c@{}}10.87\\ (0.494)\end{tabular} &
  \textbf{\begin{tabular}[c]{@{}c@{}}7.00\\ (0.318)\end{tabular}} &
  \begin{tabular}[c]{@{}c@{}}13.20\\ (0.600)\end{tabular} &
  \begin{tabular}[c]{@{}c@{}}12.07\\ (0.549)\end{tabular} &
  \begin{tabular}[c]{@{}c@{}}10.80\\ (0.491)\end{tabular} &
  \begin{tabular}[c]{@{}c@{}}14.67\\ (0.667)\end{tabular} &
  \begin{tabular}[c]{@{}c@{}}11.33\\ (0.515)\end{tabular} &
  \begin{tabular}[c]{@{}c@{}}9.60\\ (0.436)\end{tabular} &
  \begin{tabular}[c]{@{}c@{}}8.60\\ (0.391)\end{tabular} \\
17 &
  3T Endgame &
  \begin{tabular}[c]{@{}c@{}}8.80\\ (0.978)\end{tabular} &
  \begin{tabular}[c]{@{}c@{}}5.88\\ (0.653)\end{tabular} &
  \begin{tabular}[c]{@{}c@{}}5.80\\ (0.644)\end{tabular} &
  \begin{tabular}[c]{@{}c@{}}7.53\\ (0.837)\end{tabular} &
  \begin{tabular}[c]{@{}c@{}}6.73\\ (0.748)\end{tabular} &
  \begin{tabular}[c]{@{}c@{}}5.60\\ (0.622)\end{tabular} &
  \begin{tabular}[c]{@{}c@{}}7.20\\ (0.800)\end{tabular} &
  \begin{tabular}[c]{@{}c@{}}8.07\\ (0.897)\end{tabular} &
  \begin{tabular}[c]{@{}c@{}}7.47\\ (0.830)\end{tabular} &
  \textbf{\begin{tabular}[c]{@{}c@{}}5.29\\ (0.588)\end{tabular}} \\
18 &
  Vote &
  \begin{tabular}[c]{@{}c@{}}8.40\\ (0.525)\end{tabular} &
  \begin{tabular}[c]{@{}c@{}}7.87\\ (0.492)\end{tabular} &
  \begin{tabular}[c]{@{}c@{}}5.80\\ (0.363)\end{tabular} &
  \begin{tabular}[c]{@{}c@{}}8.47\\ (0.529)\end{tabular} &
  \begin{tabular}[c]{@{}c@{}}9.33\\ (0.583)\end{tabular} &
  \textbf{\begin{tabular}[c]{@{}c@{}}5.20\\ (0.325)\end{tabular}} &
  \begin{tabular}[c]{@{}c@{}}8.87\\ (0.554)\end{tabular} &
  \begin{tabular}[c]{@{}c@{}}8.53\\ (0.533)\end{tabular} &
  \begin{tabular}[c]{@{}c@{}}9.60\\ (0.600)\end{tabular} &
  \begin{tabular}[c]{@{}c@{}}8.67\\ (0.542)\end{tabular} \\
19 &
  WaveformV2 &
  \begin{tabular}[c]{@{}c@{}}39.60\\ (0.990)\end{tabular} &
  \begin{tabular}[c]{@{}c@{}}29.00\\ (0.725)\end{tabular} &
  \begin{tabular}[c]{@{}c@{}}30.40\\ (0.760)\end{tabular} &
  \begin{tabular}[c]{@{}c@{}}36.60\\ (0.915)\end{tabular} &
  \begin{tabular}[c]{@{}c@{}}35.80\\ (0.895)\end{tabular} &
  \begin{tabular}[c]{@{}c@{}}25.00\\ (0.625)\end{tabular} &
  \begin{tabular}[c]{@{}c@{}}36.00\\ (0.900)\end{tabular} &
  \begin{tabular}[c]{@{}c@{}}37.20\\ (0.930)\end{tabular} &
  \begin{tabular}[c]{@{}c@{}}34.40\\ (0.860)\end{tabular} &
  \textbf{\begin{tabular}[c]{@{}c@{}}24.56\\ (0.614)\end{tabular}} \\
20 &
  Wine &
  \begin{tabular}[c]{@{}c@{}}11.07\\ (0.852)\end{tabular} &
  \begin{tabular}[c]{@{}c@{}}6.67\\ (0.513)\end{tabular} &
  \begin{tabular}[c]{@{}c@{}}6.73\\ (0.518)\end{tabular} &
  \begin{tabular}[c]{@{}c@{}}10.73\\ (0.825)\end{tabular} &
  \begin{tabular}[c]{@{}c@{}}10.07\\ (0.775)\end{tabular} &
  \textbf{\begin{tabular}[c]{@{}c@{}}6.20\\ (0.477)\end{tabular}} &
  \begin{tabular}[c]{@{}c@{}}9.53\\ (0.733)\end{tabular} &
  \begin{tabular}[c]{@{}c@{}}9.07\\ (0.698)\end{tabular} &
  \begin{tabular}[c]{@{}c@{}}9.40\\ (0.723)\end{tabular} &
  \begin{tabular}[c]{@{}c@{}}6.34\\ (0.488)\end{tabular} \\
21 &
  Zoology &
  \begin{tabular}[c]{@{}c@{}}11.67\\ (0.729)\end{tabular} &
  \begin{tabular}[c]{@{}c@{}}7.67\\ (0.479)\end{tabular} &
  \begin{tabular}[c]{@{}c@{}}5.35\\ (0.334)\end{tabular} &
  \begin{tabular}[c]{@{}c@{}}12.40\\ (0.775)\end{tabular} &
  \begin{tabular}[c]{@{}c@{}}11.80\\ (0.738)\end{tabular} &
  \textbf{\begin{tabular}[c]{@{}c@{}}5.20\\ (0.325)\end{tabular}} &
  \begin{tabular}[c]{@{}c@{}}11.47\\ (0.717)\end{tabular} &
  \begin{tabular}[c]{@{}c@{}}11.93\\ (0.746)\end{tabular} &
  \begin{tabular}[c]{@{}c@{}}9.60\\ (0.600)\end{tabular} &
  \begin{tabular}[c]{@{}c@{}}5.78\\ (0.361)\end{tabular} \\
 &
  \textbf{AVG.} &
  \begin{tabular}[c]{@{}c@{}}41.98\\ (0.759)\end{tabular} &
  \begin{tabular}[c]{@{}c@{}}34.25\\ (0.544)\end{tabular} &
  \begin{tabular}[c]{@{}c@{}}32.48\\ (0.516)\end{tabular} &
  \begin{tabular}[c]{@{}c@{}}40.41\\ (0.707)\end{tabular} &
  \begin{tabular}[c]{@{}c@{}}39.16\\ (0.677)\end{tabular} &
  \begin{tabular}[c]{@{}c@{}}32.68\\ (0.501)\end{tabular} &
  \begin{tabular}[c]{@{}c@{}}39.65\\ (0.695)\end{tabular} &
  \begin{tabular}[c]{@{}c@{}}36.37\\ (0.672)\end{tabular} &
  \begin{tabular}[c]{@{}c@{}}38.07\\ (0.645)\end{tabular} &
  \begin{tabular}[c]{@{}c@{}}30.97\\ (0.468)\end{tabular}\\\hline
\end{tabular}}
\begin{tablenotes} 
		\item The \textbf{bolded} values represent the best outcomes.
     \end{tablenotes} 
\end{threeparttable}
\end{table*}

\begin{table}[]
    \centering
    \begin{threeparttable}
     \small
\caption{Comparison of the average fitness measure of each algorithm.}\label{Tmean}
   \begin{tabular}{llllllllllll}
\hline No. & Dataset & ALO & BSO & GA & GWO & PSO & bBOA & DA & SSA & ISSA & TFSSA\\
\hline1   & BreastCO & 0.048 & 0.084 & 0.046 & 0.047 & 0.045 & 0.040          & 0.041 & 0.046 & 0.048 & \textbf{0.032} \\
2   & BreastCWD     & 0.068 & 0.102 & 0.055 & 0.068 & 0.068 & \textbf{0.042} & 0.068 & 0.067 & 0.071 & 0.045          \\
3   & Clean-1       & 0.160 & 0.177 & 0.134 & 0.147 & 0.150 & 0.113          & 0.151 & 0.147 & 0.160 & \textbf{0.108} \\
4   & Clean-2       & 0.055 & 0.065 & 0.062 & 0.060 & 0.060 & 0.051          & 0.057 & 0.054 & 0.058 & \textbf{0.041} \\
5   & CongressVR   & 0.069 & 0.149 & 0.063 & 0.073 & 0.082 & 0.045          & 0.074 & 0.042 & 0.072 & \textbf{0.035} \\
6   & Exactly-1      & 0.301 & 0.400 & 0.270 & 0.282 & 0.257 & \textbf{0.040} & 0.257 & 0.286 & 0.298 & 0.229          \\
7   & Exactly-2     & 0.305 & 0.367 & 0.308 & 0.310 & 0.308 & 0.260          & 0.303 & 0.306 & 0.303 & \textbf{0.240} \\
8   & StatlogH      & 0.228 & 0.307 & 0.216 & 0.228 & 0.226 & \textbf{0.180} & 0.228 & 0.220 & 0.244 & 0.185          \\
9   & IonosphereVS & 0.145 & 0.149 & 0.109 & 0.136 & 0.124 & 0.096          & 0.133 & 0.122 & 0.116 & \textbf{0.081} \\
10  & KrvskpEW     & 0.108 & 0.242 & 0.083 & 0.094 & 0.086 & 0.054          & 0.080 & 0.140 & 0.116 & \textbf{0.044} \\
11  & Lymphography & 0.219 & 0.309 & 0.187 & 0.241 & 0.214 & 0.189          & 0.225 & 0.216 & 0.231 & \textbf{0.109} \\
12  & M-of-n       & 0.188 & 0.299 & 0.205 & 0.180 & 0.164 & 0.027          & 0.178 & 0.152 & 0.172 & \textbf{0.024} \\
13  & Penglung   & 0.196 & 0.235 & 0.129 & 0.169 & 0.190 & 0.118          & 0.177 & 0.209 & 0.170 & \textbf{0.106} \\
14  & Semeion      & 0.045 & 0.049 & 0.039 & 0.050 & 0.059 & 0.036          & 0.055 & 0.057 & 0.052 & \textbf{0.021} \\
15  & SonarMR      & 0.158 & 0.209 & 0.128 & 0.143 & 0.138 & 0.086          & 0.155 & 0.154 & 0.159 & \textbf{0.079} \\
16  & Spectheart      & 0.216 & 0.252 & 0.192 & 0.219 & 0.219 & 0.160          & 0.205 & 0.217 & 0.220 & \textbf{0.120} \\
17  & 3T Endgame  & 0.249 & 0.342 & 0.243 & 0.252 & 0.253 & \textbf{0.205} & 0.249 & 0.251 & 0.251 & 0.219          \\
18  & Vote         & 0.079 & 0.162 & 0.070 & 0.085 & 0.079 & 0.044          & 0.082 & 0.085 & 0.085 & \textbf{0.037} \\
19  & WaveformV2   & 0.300 & 0.386 & 0.319 & 0.297 & 0.287 & 0.265          & 0.291 & 0.301 & 0.298 & \textbf{0.254} \\
20  & Wine       & 0.054 & 0.139 & 0.051 & 0.060 & 0.055 & \textbf{0.023} & 0.052 & 0.050 & 0.056 & \textbf{0.023} \\
21  & Zoology          & 0.085 & 0.190 & 0.073 & 0.055 & 0.062 & 0.034          & 0.071 & 0.075 & 0.059 & \textbf{0.021} \\
    & \textbf{AVG.}      & 0.156 & 0.220 & 0.142 & 0.152 & 0.149 & 0.100          & 0.149 & 0.152 & 0.154 & 0.098\\ \hline 
  \end{tabular}
\begin{tablenotes} 
		\item The \textbf{bolded} values represent the best outcomes.
     \end{tablenotes} 
\end{threeparttable}
  
\end{table}

\begin{table*}[]
    \centering
    \begin{threeparttable}
    \small
    \caption{Comparison of the best fitness measure of each algorithm.} \label{Tbest}
    \begin{tabular}{llllllllllll}
\hline No. & Dataset & ALO & BS0 & GA & GWO & PSO & bBOA & DA & SSA  &  ISSA & TFSSA\\
\hline 1   & BreastCO & 0.038 & 0.046 & 0.040 & 0.038 & 0.039 & 0.024          & 0.031 & 0.038 & 0.038 & \textbf{0.022} \\
2   & BreastCWD     & 0.066 & 0.065 & 0.048 & 0.048 & 0.049 & 0.032          & 0.051 & 0.052 & 0.059 & \textbf{0.029} \\
3   & Clean-1       & 0.118 & 0.130 & 0.122 & 0.117 & 0.100 & 0.088          & 0.118 & 0.100 & 0.122 & \textbf{0.074} \\
4   & Clean-2       & 0.049 & 0.058 & 0.062 & 0.056 & 0.058 & 0.037          & 0.050 & 0.052 & 0.054 & \textbf{0.033} \\
5   & CongressVR   & 0.044 & 0.076 & 0.054 & 0.042 & 0.048 & 0.030          & 0.035 & 0.045 & 0.041 & \textbf{0.026} \\
6   & Exactly-1      & 0.267 & 0.328 & 0.015 & 0.173 & 0.138 & \textbf{0.005} & 0.155 & 0.089 & 0.229 & 0.224          \\
7   & Exactly-2     & 0.252 & 0.296 & 0.295 & 0.279 & 0.275 & 0.225          & 0.238 & 0.237 & 0.270 & \textbf{0.221} \\
8   & StatlogH      & 0.172 & 0.206 & 0.202 & 0.189 & 0.178 & 0.138          & 0.159 & 0.163 & 0.194 & \textbf{0.134} \\
9   & IonosphereVS & 0.111 & 0.101 & 0.099 & 0.088 & 0.081 & 0.060          & 0.104 & 0.092 & 0.078 & \textbf{0.056} \\
10  & KruskpLW     & 0.093 & 0.133 & 0.063 & 0.090 & 0.052 & 0.036          & 0.062 & 0.084 & 0.111 & \textbf{0.032} \\
11  & Lymphography & 0.165 & 0.220 & 0.168 & 0.193 & 0.179 & 0.183          & 0.166 & 0.168 & 0.169 & \textbf{0.064} \\
12  & M-of-n       & 0.160 & 0.170 & 0.140 & 0.128 & 0.064 & 0.005          & 0.157 & 0.101 & 0.035 & \textbf{0.003} \\
13  & Penglung   & 0.085 & 0.085 & 0.137 & 0.085 & 0.086 & 0.033          & 0.035 & 0.062 & 0.112 & \textbf{0.029} \\
14  & Semeion      & 0.041 & 0.046 & 0.033 & 0.044 & 0.042 & 0.029          & 0.040 & 0.047 & 0.045 & \textbf{0.020} \\
15  & SonarMR      & 0.128 & 0.139 & 0.109 & 0.090 & 0.091 & 0.072          & 0.113 & 0.081 & 0.129 & \textbf{0.069} \\
16  & Spectheart      & 0.144 & 0.198 & 0.170 & 0.149 & 0.166 & 0.122          & 0.142 & 0.159 & 0.173 & \textbf{0.118} \\
17  & 3T Endgame  & 0.213 & 0.252 & 0.232 & 0.223 & 0.204 & 0.195          & 0.217 & 0.219 & 0.213 & \textbf{0.183} \\
18  & Vote         & 0.043 & 0.065 & 0.061 & 0.051 & 0.039 & 0.016          & 0.051 & 0.060 & 0.050 & \textbf{0.012} \\
19  & WaveformLW   & 0.294 & 0.338 & 0.312 & 0.283 & 0.271 & 0.254          & 0.278 & 0.291 & 0.291 & \textbf{0.250} \\
20  & Wine       & 0.029 & 0.061 & 0.038 & 0.019 & 0.028 & 0.005          & 0.031 & 0.028 & 0.016 & \textbf{0.003} \\
21  & Zoology          & 0.026 & 0.025 & 0.061 & 0.007 & 0.008 & 0.002          & 0.007 & 0.009 & 0.026 & \textbf{0.002}\\  
& \textbf{AVG.} & 0.121 & 0.145 & 0.117 & 0.114 & 0.105 & 0.076          & 0.107 & 0.104 & 0.117 & 0.076 \\\hline
\end{tabular}
 \begin{tablenotes} 
		\item The \textbf{bolded} values represent the best outcomes.
     \end{tablenotes} 
    \end{threeparttable}
       
\end{table*}

\begin{table*}[]
    \centering
    \small
    \begin{threeparttable}
    \caption{Comparison of the worst fitness measure of each algorithm.} \label{Tworst}
    \begin{tabular}{llllllllllll}
\hline No. & Dataset & ALO & BSO & GA & GWO & PSO & bBOA & DA & SSA & ISSA & TFSSA\\
\hline 1 & BreastCO & 0.059 & 0.196 & 0.051 & 0.054 & 0.06 & 0.041 & 0.059 & 0.056 & 0.058 & \textbf{0.036} \\
2 & BreastCWD & 0.083 & 0.144 & 0.063 & 0.085 & 0.078 & 0.049 & 0.09 & 0.095 & 0.088 & \textbf{0.049} \\
3 & Clean-1 & 0.193 & 0.208 & \textbf{0.143} & 0.187 & 0.186 & \textbf{0.138} & 0.178 & 0.214 & 0.200 & 0.153 \\
4 & Clean-2 & 0.060 & 0.073 & 0.071 & 0.063 & 0.061 & 0.068 & 0.059 & 0.057 & 0.064 & \textbf{0.043} \\
5 & CongressVR & 0.110 & 0.267 & 0.083 & 0.110 & 0.149 & 0.058 & 0.107 & 0.096 & 0.120 & \textbf{0.053} \\
6 & Exactly-1 & 0.343 & 0.448 & 0.378 & 0.344 & 0.384 & \textbf{0.115} & 0.319 & 0.375 & 0.335 & 0.285 \\
7 & Exactly-2 & 0.355 & 0.517 & 0.331 & 0.333 & 0.335 & 0.291 & 0.330 & 0.337 & 0.363 & \textbf{0.287} \\
8 & StatlogH & 0.289 & 0.378 & 0.261 & 0.256 & 0.288 & 0.195 & 0.284 & 0.277 & 0.299 & \textbf{0.191} \\
9 & IonosphereVS & 0.168 & 0.195 & 0.134 & 0.179 & 0.163 & 0.118 & 0.157 & 0.155 & 0.157 & \textbf{0.114} \\
10 & KruskpLW & 0.118 & 0.344 & 0.150 & 0.096 & 0.164 & 0.064 & 0.097 & 0.176 & 0.121 & \textbf{0.060} \\
11 & Lymphography & 0.251 & 0.378 & 0.220 & 0.299 & 0.276 & {0.194} & 0.303 & 0.261 & 0.299 & \textbf{0.146} \\
12 & M-of-n & 0.224 & 0.391 & 0.288 & 0.235 & 0.287 & \textbf{0.110} & 0.210 & 0.236 & 0.212 & 0.206 \\
13 & Penglung & 0.300 & 0.460 & 0.190 & 0.246 & 0.328 & 0.169 & 0.326 & 0.379 & 0.273 & \textbf{0.153} \\
14 & Semeion & 0.049 & 0.056 & 0.043 & 0.064 & 0.072 & 0.049 & 0.070 & 0.077 & 0.065 & \textbf{0.025} \\
15 & SonarMR & 0.216 & 0.253 & 0.156 & 0.218 & 0.187 & 0.109 & 0.217 & 0.198 & 0.214 & \textbf{0.103} \\
16 & Spectheart & 0.271 & 0.322 & 0.218 & 0.265 & 0.265 & 0.209 & 0.252 & 0.271 & 0.262 & \textbf{0.201} \\
17 & 3T Endgame & 0.275 & 0.436 & 0.255 & 0.309 & 0.331 & \textbf{0.216} & 0.293 & 0.307 & 0.293 & 0.225 \\
18 & Vote & 0.113 & 0.256 & 0.088 & 0.169 & 0.119 & 0.057 & 0.124 & 0.138 & 0.118 & \textbf{0.056} \\
19 & WaveformV2 & 0.304 & 0.434 & 0.319 & 0.316 & 0.303 & 0.265 & 0.299 & 0.313 & 0.305 & \textbf{0.259} \\
20 & Wine & 0.075 & 0.303 & 0.082 & 0.142 & 0.075 & 0.028 & 0.086 & 0.077 & 0.076 & \textbf{0.026} \\
21 & Zoology & 0.158 & 0.430 & 0.101 & 0.203 & 0.182 & 0.048 & 0.125 & 0.181 & 0.107 & \textbf{0.039} \\
\multicolumn{1}{l}{} & \textbf{AVG.} & \multicolumn{1}{l}{0.191} & \multicolumn{1}{l}{0.309} & \multicolumn{1}{l}{0.173} & \multicolumn{1}{l}{0.199} & \multicolumn{1}{l}{0.204} & \multicolumn{1}{l}{0.123} & \multicolumn{1}{l}{0.190} & \multicolumn{1}{l}{0.204} & \multicolumn{1}{l}{0.192} & \multicolumn{1}{l}{0.129} \\ \hline
\end{tabular}
\begin{tablenotes} 
		\item The \textbf{bolded} values represent the best outcomes.
     \end{tablenotes} 
    \end{threeparttable}
    
\end{table*}

\begin{table*}[]
    \centering
    \small
    \begin{threeparttable}
\caption{Comparison of the standard deviation fitness measure of each algorithm.} \label{Tstd}
 \begin{tabular}{llllllllllll}
\hline No. & Dataset & ALO & BSO & GA & GWO & PSO & bBOA & DA & SSA & ISSA & TFSSA\\
\hline 1  & BreastCO & 0.008 & 0.023 & {0.005} & 0.011          & 0.008          & 0.006          & 0.011 & 0.013 & 0.009 & \textbf{0.005} \\
2  & BreastCWD     & 0.007 & 0.044 & {0.003} & 0.006          & 0.006          & {0.003} & 0.010 & 0.005 & 0.005 & \textbf{0.003} \\
3  & Clean-1       & 0.017 & 0.048 & \textbf{0.008} & 0.019          & 0.026          & 0.010          & 0.021 & 0.016 & 0.020 & 0.018          \\
4  & Clean-2       & 0.023 & 0.036 & 0.136          & 0.051          & 0.067          & {0.012} & 0.038 & 0.066 & 0.025 & \textbf{0.010} \\
5  & CongressVR   & 0.030 & 0.066 & {0.014} & 0.013          & 0.019          & 0.020          & 0.022 & 0.023 & 0.025 & \textbf{0.011} \\
6  & Exactly-1      & 0.031 & 0.051 & 0.020          & 0.023          & 0.029          & \textbf{0.011} & 0.028 & 0.035 & 0.028 & 0.020          \\
7  & Exactly-2     & 0.019 & 0.023 & \textbf{0.012} & 0.022          & 0.023          & 0.059          & 0.015 & 0.019 & 0.021 & 0.054          \\
8  & StatlogH      & 0.009 & 0.094 & 0.033          & \textbf{0.002} & 0.045          & 0.008          & 0.014 & 0.037 & 0.005 & 0.011          \\
9  & IonosphereVS & 0.026 & 0.049 & 0.016          & 0.030          & 0.028          & {0.014} & 0.041 & 0.032 & 0.042 & \textbf{0.010} \\
10 & KrvskpEW     & 0.020 & 0.074 & 0.054          & 0.030          & 0.058          & {0.033} & 0.018 & 0.035 & 0.049 & \textbf{0.031} \\
11 & Lymphography & 0.025 & 0.037 & \textbf{0.013} & 0.040          & 0.029          & 0.018          & 0.023 & 0.030 & 0.021 & 0.018          \\
12 & M-of-n       & 0.035 & 0.036 & 0.016          & 0.031          & 0.027          & 0.035          & 0.030 & 0.031 & 0.025 & \textbf{0.015} \\
13 & Penglung   & 0.020 & 0.053 & \textbf{0.006} & 0.026          & 0.034          & 0.007          & 0.023 & 0.025 & 0.021 & 0.018          \\
14 & Semeion      & 0.019 & 0.057 & {0.009} & 0.029          & 0.019          & 0.010          & 0.019 & 0.024 & 0.020 & \textbf{0.008} \\
15 & SonarMR      & 0.004 & 0.088 & 0.003          & 0.012          & 0.012          & {0.001} & 0.009 & 0.009 & 0.006 & \textbf{0.001} \\
16 & Spectheart      & 0.012 & 0.067 & 0.011          & 0.033          & 0.012          & \textbf{0.010} & 0.013 & 0.014 & 0.015 & 0.012          \\
17 & 3T Endgame  & 0.035 & 0.128 & \textbf{0.015} & 0.047          & 0.050          & 0.044          & 0.037 & 0.047 & 0.028 & 0.041          \\
18 & Vote         & 0.021 & 0.026 & \textbf{0.009} & 0.022          & 0.028          & 0.016          & 0.018 & 0.029 & 0.022 & 0.015          \\
19 & WaveformV2   & 0.004 & 0.005 & 0.004          & 0.002          & {0.001} & {0.001} & 0.003 & 0.002 & 0.004 & \textbf{0.001} \\
20 & Wine       & 0.072 & 0.102 & \textbf{0.018} & 0.046          & 0.077          & 0.056          & 0.077 & 0.093 & 0.052 & \textbf{0.018} \\
21 & Zoology          & 0.007 & 0.002 & \textbf{0.001} & 0.005          & 0.002          & 0.004          & 0.003 & 0.006 & 0.004 & 0.002 \\
\hline
\end{tabular}
 \begin{tablenotes} 
		\item The \textbf{bolded} values represent the best outcomes.
\end{tablenotes} 
    \end{threeparttable}
          
\end{table*}

\begin{table*}[]
\caption{Comparison of the average running time of each algorithm.}\label{Ttime}
    \centering
    \small
\begin{tabular}{llllllllllll}
\hline No. & Dataset & ALO & BSO & GA & GWO & PSO & bBOA & DA & SSA & ISSA & TFSSA \\\hline
1 & BreastCO & 4.90 & 3.02 & 2.34 & 3.45 & 2.39 & 2.48 & 2.36 & 2.32 & 2.45 & \textbf{2.27} \\
2 & BreastCWD & 2.87 & 2.85 & 2.37 & 3.61 & 2.43 & 2.82 & 2.41 & 2.36 & 2.35 & \textbf{2.32} \\
3 & Clean-1 & 5.31 & 3.58 & 4.07 & \textbf{3.39} & 3.66 & 4.05 & 3.61 & 3.50 & 3.54 & 3.84 \\
4 & Clean-2 & 310.83 & 223.69 & 279.32 & \textbf{158.67} & 227.16 & 171.84 & 223.70 & 173.07 & 182.94 & 159.32 \\
5 & CongressVR & 2.88 & 3.33 & 2.81 & 3.32 & 2.64 & 3.03 & 2.59 & 2.61 & 2.72 & \textbf{2.59} \\
6 & Exactly-1 & 3.92 & 4.58 & \textbf{3.85} & 4.04 & 4.53 & 4.06 & 4.63 & 5.18 & 4.65 & 4.96 \\
7 & Exactly-2 & 4.22 & 4.62 & \textbf{4.15} & 4.52 & 4.82 & 4.60 & 4.88 & 4.20 & 4.22 & 4.20 \\
8 & StatlogH & 2.69 & 2.96 & \textbf{2.46} & 3.09 & 2.49 & 2.78 & 2.52 & 2.62 & 2.47 & 2.50 \\
9 & IonosphereVS & 3.14 & 3.10 & 2.58 & 3.25 & 2.64 & 2.96 & 2.60 & 2.54 & 2.57 & \textbf{2.47} \\
10 & KrvskpEW & 18.16 & 11.56 & 10.42 & \textbf{9.53} & 17.16 & 13.84 & 15.89 & 13.89 & 13.03 & 12.07 \\
11 & Lymphography & 2.68 & 2.94 & 2.46 & 2.98 & 3.04 & 2.82 & \textbf{2.38} & 2.87 & 2.91 & 2.69 \\
12 & M-of-n & 4.08 & 4.07 & 3.53 & \textbf{3.39} & 4.56 & 4.04 & 3.74 & 4.26 & 4.14 & 4.19 \\
13 & Penglung & 7.65 & 3.10 & 2.51 & 2.49 & 2.56 & 4.13 & 2.55 & 2.50 & 2.50 & \textbf{2.45} \\
14 & Semeion & 28.41 & 14.33 & \textbf{13.10} & 31.67 & 24.51 & 19.92 & 24.06 & 21.82 & 19.21 & 15.45 \\
15 & SonarMR & 3.30 & 2.93 & \textbf{2.39} & 2.97 & 2.62 & 2.92 & 2.72 & 2.75 & 2.59 & 2.45 \\
16 & Spectheart & 2.88 & 2.96 & 2.45 & 3.00 & 2.40 & 2.80 & 2.38 & 2.40 & \textbf{2.38} & 2.29 \\
17 & 3T Endgame & 4.36 & 3.99 & \textbf{3.28} & 4.38 & 4.49 & 3.91 & 4.38 & 4.25 & 4.10 & 4.45 \\
18 & Vote & 2.89 & 3.26 & 2.62 & 3.25 & 2.57 & 2.82 & 2.60 & 2.53 & 2.62 & \textbf{2.47} \\
19 & WaveformV2 & 40.51 & 25.03 & 23.48 & \textbf{20.63} & 27.09 & 35.56 & 43.72 & 34.14 & 36.64 & 21.26 \\
20 & Wine & 2.68 & 2.92 & 2.46 & 3.13 & 2.45 & 2.68 & 2.43 & \textbf{2.43} & 2.47 & 2.52 \\
21 & Zoology & 2.79 & 4.85 & 2.33 & 3.25 & 2.24 & 2.66 & 2.30 & 2.21 & 2.19 & \textbf{2.15} \\
\hline
\end{tabular}
      
\end{table*}

\begin{itemize}
    \item[$\bullet$]\textit{Classification average accuracy (CAA)} is a metric that indicates how accurate the classifier is given the provided feature set. Eq.(\ref{eq.AvgPerf}) can be used to get the classification average accuracy. 
    \begin{equation}
    AvgPerf = \frac{1}{N} \sum_{i=1}^{N}\frac{1}{M} \sum_{j=1}^{M}\label{eq.AvgPerf} \operatorname{Match}\left(C_{i}, L_{i}\right),
    \end{equation}
    where $M$ denotes the amount of times the optimizer is run to pick the feature subset, $N$ denotes the amount of points in the test set, $C_{i}$ denotes the output label of the classifier for data point $i, L_{i}$ denotes the data point $i$'s reference class label. If the two input labels are identical, the match function returns 1 if they are. Otherwise, it returns 0. 
\end{itemize}
\begin{itemize}
    \item[$\bullet$]\textit{Statistical best (SB)} is the optimistic(minimum) fitness value obtained after each feature selection method runs $M$ times. As shown in Eq. (\ref{eq.Best}).
    \begin{equation}
   Best =\min _{i=1}^{M} g_{\text {* }}^{i}\label{eq.Best},
    \end{equation}
    where $g_{*}^{i}$ indicates the best result determined after $i$ times of operation.
\end{itemize}
\begin{itemize}
    \item [$\bullet$]\textit{Statistical worst (SW)} in contrast to \textit{Statistical best (SB)}. Worst is the pessimistic result, which can be expressed as Eq. (\ref{eq.Worst}).
    \begin{equation}
    Worst =\max _{i=1}^{M} g_{*}^{i}\label{eq.Worst}.
    \end{equation}
\end{itemize}
\begin{itemize}
    \item[$\bullet$]\textit{Statistical mean (SM)} is the average value of the solution obtained by running under the condition of $M$ times. As shown in Eq. (\ref{eq.Mean}).
    \begin{equation}
    Mean =\frac{1}{M} \sum_{i=1}^{M} g_{*}^{i}\label{eq.Mean}.
    \end{equation}
   
\end{itemize}
\begin{itemize}
    \item[$\bullet$]\textit{Statistical Std (SSD)} is a representation of the variation in the obtained minimum(best) solutions for $M$ different runs of a stochastic optimizer. Std is a stability and robustness metric for optimizers; if Std is small, the optimizer always converges to the same solution; on the contrary, the optimizer produces numerous random outcomes. As shown in Eq. (\ref{eq.Std}).
    \begin{equation}
    Std=\sqrt{\frac{1}{M-1} \sum\left(g_{*}^{i}-\text {$Mean$}\right)^{2}}\label{eq.Std}.
    \end{equation}
\end{itemize}
\begin{itemize}
    \item[$\bullet$]\textit{Selection average size (SAS)} represents the average amount of features selected. As shown in Eq. (\ref{eq.AVGSelectionSZ}).
    \begin{equation}
    AVGSelectionSZ =\frac{1}{M} \sum_{i=1}^{M} \frac{\operatorname{size}\left(g_{*}^{i}\right)}{Di}\label{eq.AVGSelectionSZ},
    \end{equation}
    where $\operatorname{size}(x)$ is the amount of on values for the vector $x$, and $Di$ is the dimension of each dataset.
\end{itemize}

\begin{table*}[]

\caption{p-values of the Wilcoxon test of TFSSA vs. others.}\label{Tp-values}
\centering
\small

\begin{threeparttable}
\scalebox{0.85}{
  \begin{tabular}{llllllllllll}
 \hline 
  & Dataset & ALO & BSO & GA & GWO & PSO & bBOA & DA & SSA & ISSA & TFSSA \\\hline
1 & BreastCO & 1.09E-02 & 5.06E-03 & 5.03E-03 & 6.91E-03 & 1.09E-02 & 1.14E-01 & \uline{6.87E-03} & 5.06E-03 & 7.63E-03 & 6.11E-04 \\
2 & BreastCWD & 6.48E-04 & 6.23E-04 & 6.32E-04 & 6.49E-04 & 6.43E-04 & 6.58E-04 & 7.12E-04 & 6.58E-04 & 6.45E-04 & 7.62E-03 \\
3 & Clean-1 & 6.58E-04 & 9.85E-04 & 2.15E-03 & 8.01E-04 & 2.15E-03 & 8.03E-04 & 1.79E-03 & 6.53E-04 & 4.51E-03 & \uline{ 7.21E-01} \\
4 & Clean-2 & 1.03E-02 & 6.23E-04 & 6.23E-04 & 6.23E-04 & 6.23E-04 & 2.07E-03 & 6.23E-04 & 6.23E-04 & 6.23E-04 & 6.11E-04 \\
5 & CongressVR & 9.85E-04 & 6.58E-04 & 1.19E-03 & 1.21E-03 & 6.58E-04 & 4.51E-03 & 2.16E-03 & 1.47E-03 & 4.51E-03 & \uline{  3.44E-01} \\
6 & Exactly-1 & 6.52E-04 & 6.58E-04 & 6.42E-04 & 6.48E-04 & 6.58E-04 & 6.48E-05 & 6.58E-04 & 6.52E-04 & 6.58E-04 & 5.39E-03 \\
7 & Exactly-2 & 9.87E-04 & 6.58E-04 & 6.47E-04 & 6.58E-04 & 8.05E-04 & 1.21E-03 & 8.05E-04 & 8.05E-04 & 8.98E-03 & 8.86E-03 \\
8 & StatlogH & 9.87E-04 & 6.53E-04 & 6.47E-04 & 9.79E-04 & 6.53E-04 & 8.05E-04 & 6.41E-03 & 6.53E-04 & 7.59E-03 & 8.86E-03 \\
9 & IonosphereVS & 3.09E-02 & 2.31E-02 & \uline{  8.20E-01} & \uline{ 7.83E-02} & \uline{ 3.34E-01} & \uline{ 6.09E-02} & \uline{ 3.07E-01} & \uline{ 5.32E-01} & \uline{ 4.60E-01} & \uline{ 5.23E-02} \\
10 & KrvskpEW & 4.35E-02 & 4.31E-02 & \uline{ 7.96E-02} & 4.31E-02 & \uline{ 7.96E-02} & 4.31E-02 & 4.31E-02 & 4.31E-02 & 4.31E-02 & 4.19E-02 \\
11 & Lymphography & 6.58E-04 & 6.58E-04 & 6.47E-04 & 6.58E-04 & 6.58E-04 & 6.58E-04 & 6.52E-04 & 6.58E-04 & 6.58E-04 & 6.11E-04 \\
12 & M-of-n & 6.58E-04 & 6.58E-04 & 6.53E-04 & 6.53E-04 & 6.58E-04 & 6.58E-04 & 6.58E-04 & 6.53E-04 & 6.58E-04 & 6.43E-04 \\
13 & Penglung & 1.71E-02 & 3.14E-03 & \uline{ 1.40E-01} & \uline{ 8.83E-02} & 3.09E-02 & 7.83E-02 & \uline{ 3.56E-02} & 2.68E-02 & \uline{ 4.95E-01} & \uline{ 3.02E-01} \\
14 & semeion & \uline{ 1.38E-01} & 4.31E-02 & \uline{ 7.96E-02} & 4.31E-02 & 4.31E-02 & 4.31E-02 & 4.31E-02 & 4.31E-02 & 4.31E-02 & \uline{5.31E-02} \\
15 & SonarMR & 6.58E-04 & 6.58E-04 & 6.53E-04 & 6.58E-04 & 1.79E-03 & 6.58E-04 & 6.58E-04 & 6.53E-04 & 6.58E-04 & 6.43E-04 \\
16 & Spectheart & 1.25E-02 & 6.58E-04 & 4.48E-03 & 6.58E-04 & 6.58E-04 & 5.37E-03 & 1.79E-03 & 6.58E-04 & 1.46E-02 & 4.59E-02 \\
17 & 3T Endgame & 6.50E-04 & 6.58E-04 & 6.53E-04 & 6.58E-04 & 8.03E-04 & 6.58E-04 & 6.58E-04 & 6.53E-04 & 8.05E-04 & 1.46E-02 \\
18 & Vote & 9.85E-04 & 6.53E-04 & 6.47E-04 & 9.87E-04 & 6.53E-04 & 6.58E-04 & 6.58E-04 & 6.58E-04 & 6.53E-04 & 2.30E-02 \\
19 & WaveformV2 & 4.31E-02 & 4.31E-02 & 4.31E-02 & 4.31E-02 & 4.31E-02 & 4.31E-02 & 4.31E-02 & 4.31E-02 & \uline{ 2.23E-01} & \uline{5.24E-02} \\
20 & Wine & 6.48E-04 & 6.53E-04 & 6.50E-04 & 3.77E-03 & 8.03E-04 & 6.52E-04 & 8.01E-04 & 6.48E-04 & 1.19E-03 & 4.19E-03 \\
21 & Zoology & 3.56E-02 & 3.14E-03 & 4.67E-02 & \uline{ 6.09E-01} & \uline{ 3.07E-01} & \uline{ 1.40E-01} & \uline{ 6.91E-02} & \uline{ 3.63E-01} & \uline{ 9.55E-01} & \uline{ 2.60E-01}\\\hline
\end{tabular}
}
\begin{tablenotes} 
		\item The p$\geq0.05$ are underlined.
\end{tablenotes} 
    \end{threeparttable}
\end{table*}

\begin{itemize}
    \item[$\bullet$]\textit{Wilcoxon rank-sum test (WRST)} \cite{wilcoxon1992individual} is a nonparametric statistical test used to see if the proposed techniques' outcomes are statistically different from those of other algorithms\cite{derrac2011practical}. This statistical test generates a p-value parameter, which is used to compare the significance levels of two algorithms.
\end{itemize}


\subsubsection{Comparison of the proposed algorithm and other FS methods.}
In this section, the performance of the best strategy, TFSSA, is compared to that of five top-of-the-line approaches (BSO, ALO, PSO, GWO, and GA) and four recent high-performance methods (bBOA, DA, SSA, and ISSA) that have been widely used to address the FS problem in the literature. Some performance indicators used to evaluate the algorithm's performance include classification average accuracy (CAA), selected average feature number, selected average feature rate, statistical best (SB) fitness, statistical worst (SW) fitness, statistical mean (SM) fitness, statistical standard deviation (SSD), calculation time, and Wilcoxon rank-sum test (WRST). 

In Table \ref{Tacc}, the classification average accuracy achieved by each algorithm is compared. TFSSA is preferred over other algorithms in most datasets except Exactly-1 and SonarMR. Furthermore, Figure \ref{fig:avg acc} shows the overall average classification accuracy selected by different algorithms on all datasets. We can see that the proposed algorithm ranks first with a classification accuracy of 0.9011. This result confirms that the proposed TFSSA can effectively explore the solution search space and find the optimal feature subset with the highest classification accuracy.

Table \ref{Tfea} compares the average number and average ratio of features selected by different algorithms. Both tables show that TFSSA outperforms the other algorithms in the 13 datasets. Although the number of features selected by TFSSA is not optimal in the other eight datasets, it is not significantly different from other outperformed methods. Figure \ref{fig:feat & ratio} shows the population average number of features and ratios chosen by the algorithm. It can be seen that the average number of features and ratios selected by TFSSA in all datasets rank first with 30.97 and 0.468, respectively. Although the advantage is not apparent, it can also be proved that TFSSA outperforms other algorithms in most datasets to ensure high classification accuracy because we want to focus on both the CAA and selected average feature number.

\begin{figure}[ht]
    \centering
    \includegraphics[width=0.48\textwidth]{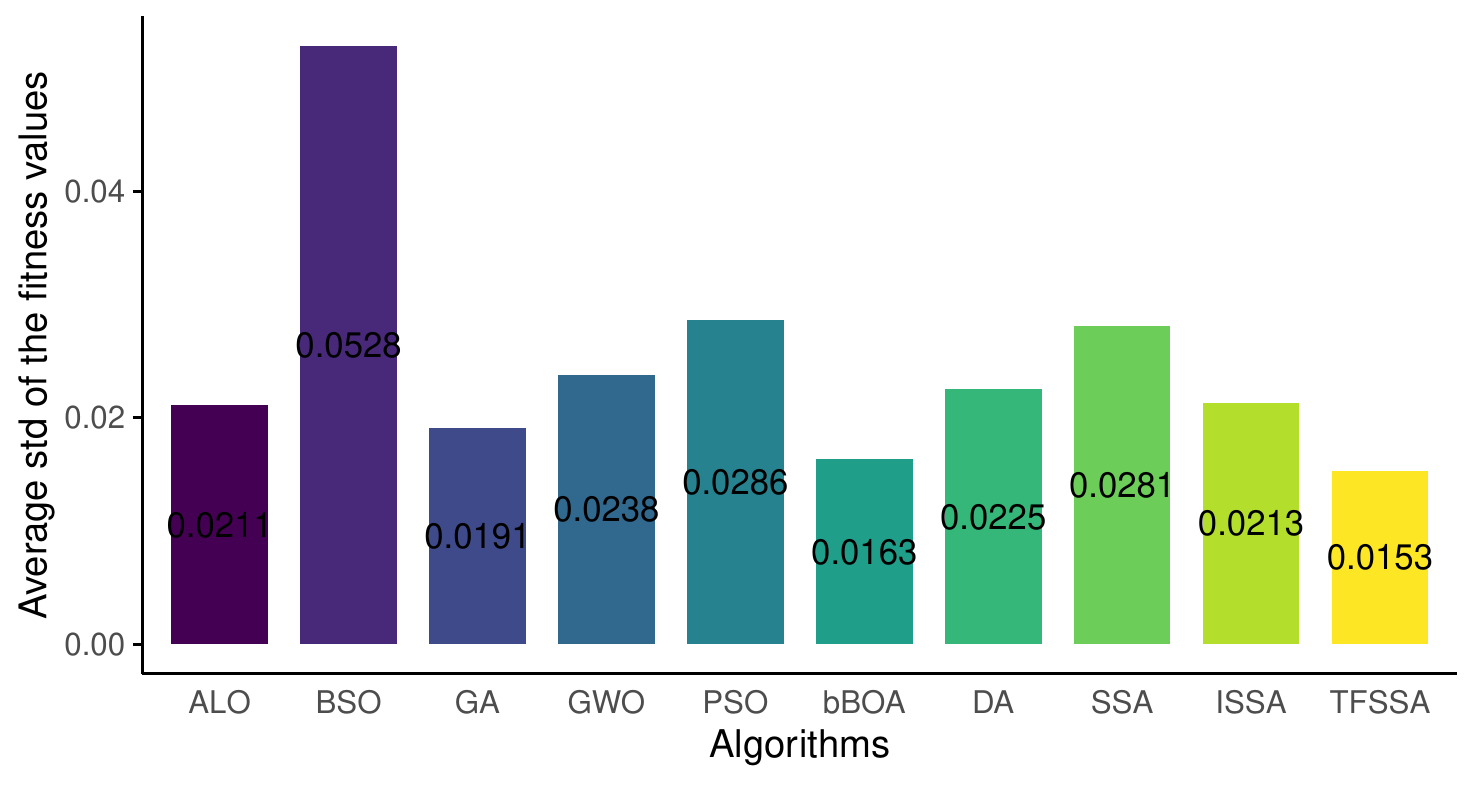}
    \caption{Comparison of total average standard deviation for mean fitness values among algorithms.}
    \label{fig:std}
\end{figure}

As a result, the number of selected attributes affected by the classification accuracy value is often slightly less than the fitness value. Table \ref{Tmean}-\ref{Tstd} presents the statistical measures (SB, SW, SM, and SSD) obtained by different runs of the algorithm on each dataset. TFSSA has a lower fitness value compared to other algorithms, checking the results. Among them, the average fitness value of TFSSA maintains a leading edge in 17 datasets, and bBOA outperforms different algorithms in 5 datasets. The overall average fitness value of TFSSA is 0.098, ranking first. The best fitness value of TFSSA maintains the lead in all datasets except Exactly-1, and its overall best fitness value is 0.076, ranking first. The worst fitness value of TFSSA outperforms other algorithms in 17 datasets, bBOA outperforms different algorithms in 4 datasets, and GA outperforms other algorithms in dataset Clean-1. From Table \ref{Tstd}, it is observed that the standard deviation of TFSSA outperforms different algorithms in 11 datasets, the standard deviation of GA outperforms different algorithms in 8 datasets, and the standard deviation of bBOA outperforms other algorithms in 2 datasets Outperforming different algorithms, the standard deviation of GWO outperforms other algorithms in the dataset StatlogH. 

The average execution time of each method in the experiment is shown in Table \ref{Ttime}. Because almost all optimization algorithms employ the same amount of iterations, computation time can be used to compare algorithm performance. We get the following observations from Table \ref{Ttime}. The ten EA, have intimate performances in terms of the time consumption for all 21 datasets. An EA-based feature selection technique, as we all know, requires a classifier to evaluate an individual. The time it takes the classifier to evaluate a set of features and samples is usually proportional to the number of features and samples.Therefore, for datasets with many features or/and samples (such as WaveformV2, Clean-2, and Semeion.), the fitness function is the most time-consuming part of EA-based feature selection algorithms. The ten EA-based algorithms used in the trials all had the same maximum number of evaluations as their termination conditions, which resulted in identical time consumption. Among them, TFSSA has the best computing time on seven datasets. In comparison, GWO performs better on five datasets, and GA performs better than other optimizers on six datasets, DA, SSA, and ISSA, each with better performance on 1 dataset. 

In addition, Table \ref{Tp-values} shows the Wilcoxon rank-sum test p-values at the 5\% significance level for the Wilcoxon rank-sum test. A p-value of less than 0.05 implies that the null hypothesis of no meaningful difference at the 5\% level is rejected. The p-values in Table \ref{Tp-values} confirm that the results of the proposed method TFSSA (especially in 12 data sets, the performance is outstanding, including BreastCWD, Clean-2, Exactly-1, Exactly-2, StatlogH, Lymphography, M-of-n, SonarMR, Spectheart, 3T Endgame, Vote and Wine.) are significantly different from those of classical and top-of-the-line algorithms on most datasets.

Overall, the Tables \ref{Tacc}-\ref{Tstd} results show that TFSSA is capable of balancing exploration and exploitation in the optimization search process. It can be seen that TFSSA outperforms other algorithms in both small and large datasets. This experiment employed four major data sets: Clean-2(No.4),  krvskpEW(No.10), Penglung(No.13), and semeion(No.14). The proposed method outperforms all other algorithms in CAA, selected average feature number, selected average feature rate, measures of fitness (SB, SW, SM, and SSD), and the Wilcoxon rank-sum test.

We can conclude from all of these experiments that employing the Lévy flights strategy and improved tent chaos improves the robustness and performance of the proposed algorithm. This method solves FS difficulties by combining global search algorithms (which are good for exploration) and local search algorithms (which are good for development). It's critical to establish a balance between exploration and production in the FS problem to avoid a large number of local solutions and discover an accurate approximation of the optimal solution. This is the primary reason behind TFSSA's improved performance when compared to the comparative algorithm used in this study. TFSSA has the fewest amount of features and the highest accuracy among the ten approaches. However, as compared to the other methods utilized in this study, TFSSA takes greater calculation time. Another drawback of the suggested random packing-based FS strategy is the imprecision with which the optimization results can be repeated. The algorithm's subset of features selected for different applications has been noted, which may mislead users when determining which subset to evaluate.

To avoid overfitting issues, K-fold cross-validation is a great approach. A k-NN classifier is used in the suggested technique, which can swiftly learn from training data and produce high-quality results. It's a common wrapper-based strategy. When moving to a different classifier, such as support vector machines or random forests, the runtime may rise. As a result, diverse classifiers must be handled with caution, particularly in real-world applications.
\begin{table*}[]
\centering
    \caption{COVID-19 dataset description.} \label{T-covid-19}
\begin{tabular}{llll}\hline
Dateset  & No. Features & No. Instances & Area    \\
\hline COVID-19 & 15     & 1085  & Medical \\\hline
\end{tabular}
\end{table*}
\section{Real-world dataset instances}\label{Realworld}
The COVID-19 is an infectious disease caused by severe acute respiratory syndrome coronavirus type 2 (SARS-CoV-2), which has led to an epidemic that has continued to this day and has become one of the epidemics with the largest number of deaths in human history \cite{sayed2020nature}. The first known patient with the disease was diagnosed in Wuhan, Hubei Province, the people's Republic of China at the end of 2019 (although the disease is likely to have infected humans before). Since then, the disease has been detected all over the world and is still spreading. At the same time, mankind hopes to defeat the virus through various technologies, so it has once again started a protracted war against the virus. According to research, artificial intelligence (AI) has become a weapon with great potential to fight SARS-CoV-2 \cite{chen2020diagnostic}.
\begin{figure*}
    \centering
    \includegraphics[width=0.85\textwidth]{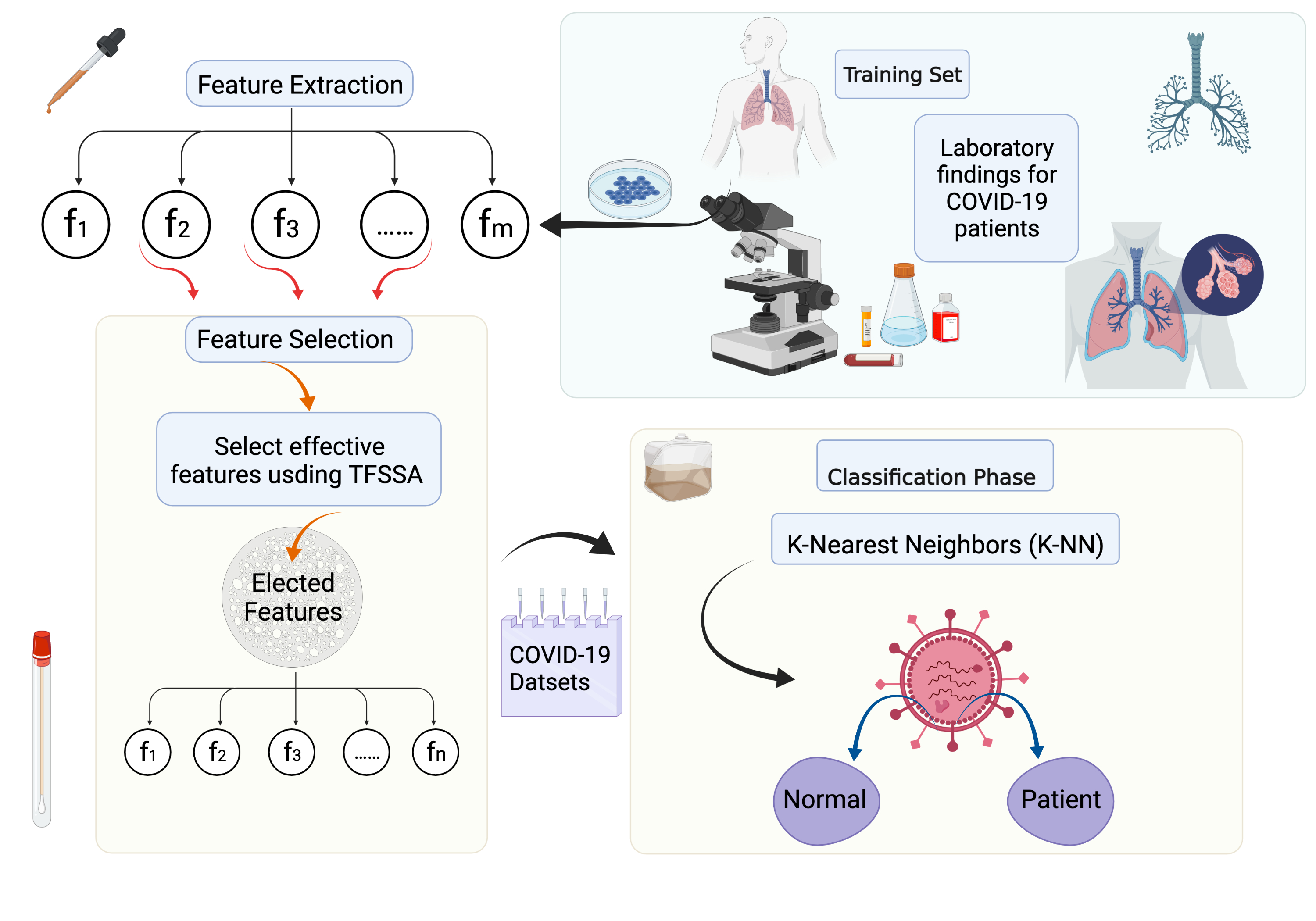}
    \caption{The proposed TFSSA classification strategy in COVID-19.}
    \label{fig:bio}
\end{figure*}

In this section, the proposed TFSSA is employed for 2019 Coronavirus Disease patient health prediction, as shown in \figref{fig:bio}. The dataset of COVID-19 patients\renewcommand{\thefootnote}{2}
\footnote{\url{https://github.com/yyy24601/Covid-19-Patient-Health-Analytics}} was gathered completely in the \cite{iwendi2020covid}. Table \tabref{T-covid-19} and \tabref{T-covid} gives a summary of the real-world datasets used. The purpose of this study was to predict illness and health based on a given variable. First, the "normal" and "patient" patient data sets are removed from the COVID-19 master data set. The remaining 15 attributes are then translated into numerical numbers. Finally, the data is separated into two groups: training set and testing set.

\begin{figure}[htbp]
\centering
\begin{minipage}[t]{0.48\textwidth}
\centering
\includegraphics[width=8cm]{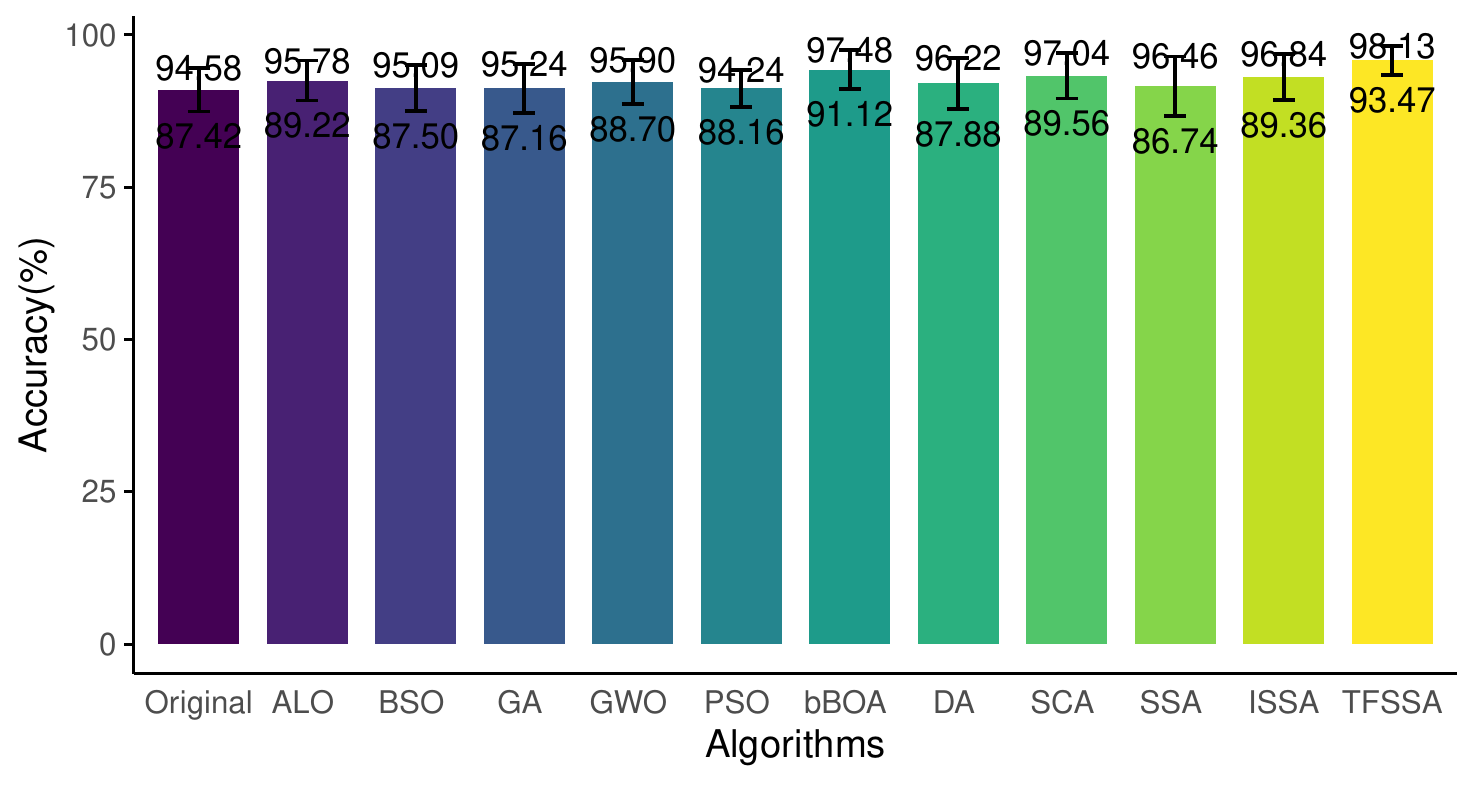}
\end{minipage}
\begin{minipage}[t]{0.48\textwidth}
\centering
\includegraphics[width=8cm]{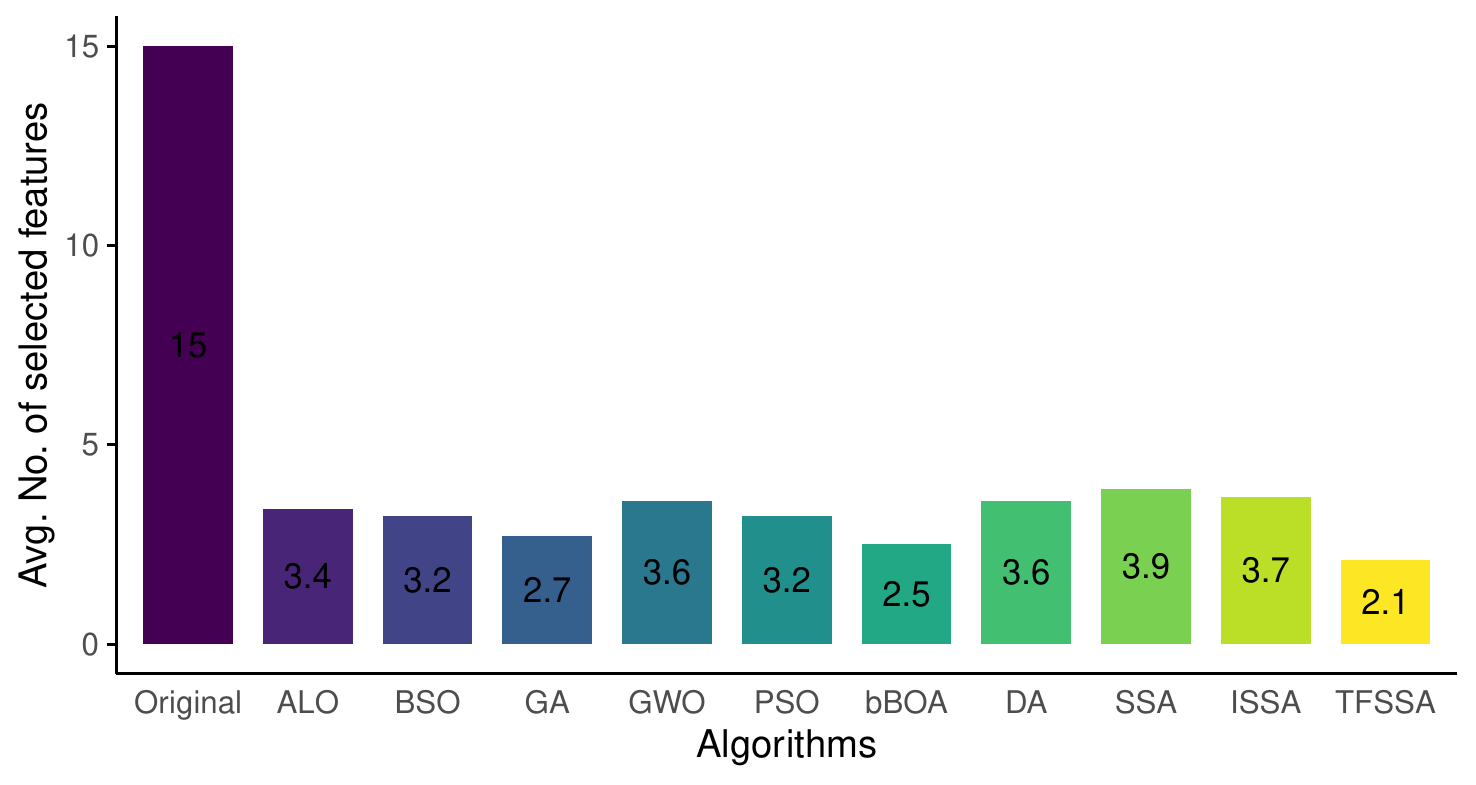}
\end{minipage}
\caption{Accuracy rating and feature size of the proposed TFSSA on the COVID-19 dataset.}\label{fig:Acc+No.feat-covid19}
\end{figure}

As can be seen from \figref{fig:Acc+No.feat-covid19}, TFSSA achieves the highest average classification accuracy of 93.47\% and the lowest average feature selection number of 2.1. On the other hand, the results reveal that for TFSSA inpatient health prediction, around three features were sufficient. The most popular features were id, age, and nationality, according to the results. Furthermore, the data suggest that the TFSSA algorithm has never chosen symptom\_4, symptom\_5, or symptom\_6. Also, to validate TFSSA's classification performance, we try to remove symptoms 4, 5, and 6, and the difference is minor when compared to previous experimental findings. As a result, these features are unable to appropriately detect the data pattern in the patient health prediction process. The performance of TFSSA is observed after eliminating these characteristics, and the accuracy is barely affected. To continue studying the performance of TFSSA, we also remove the original feature (ID) from the data set, the experiment revealed that AAC is about 91.3\%. The researchers said that in the future, more abundant, detailed, and comprehensive clinical features should be collected to more accurately predict the health status of patients. 

\section{Conclusions}\label{Conclusions}
This paper proposes a TFSSA that combines a Tent chaotic map and Lévy flights to solve the FS problem in packing mode. Combining Tent chaotic map and Lévy flights significantly improves the algorithm's performance in FS. In this paper, twenty-one datasets from the UC Irvine Machine Learning Repository are utilized to validate the suggested method's performance. In addition, the model is also applied to the diagnosis and prediction of covid-19. Nine criteria are reported to evaluate each technique: classification accuracy, average selection size, average selection rate, measures of fitness (SB, SW, SM, and SSD), computation time, and Wilcoxon rank-sum test. Comparing the proposed method with five top-of-the-line methods (BSO, ALO, PSO, GWO, and GA) and the four latest high-performance methods (bBOA, DA, SSA, and ISSA), TFSSA achieves the goal of lowering the number of features and boosting the model's accuracy by removing as many irrelevant and redundant features as possible. Our results show that the proposed TFSSA is superior to other algorithms in most data sets. Therefore, TFSSA can find the best feature subset and obtain high accuracy when applied to various FS tasks.

\begin{table}[]
    \centering
    \caption{The description of the 2019 Coronavirus Disease dataset.} \label{T-covid}
    \begin{tabular}{lll}
\hline No. & Features & Feature description  \\
\hline 1 & code(id) & Patients' identification numbers \\
2 & location & The place where patients are situated  \\
3 & country & The country from which the patients come  \\
4 & gender & The patients' gender \\
5 & age & How old patients are  \\
6 & sym\_on & When people first show symptoms \\
7 & hosp\_vis & The date patients visit hospital \\
8 & vis\_wuhan & Whether or not the patients visited Wuhan, CN\\
9 & from\_wuhan & Whether or not the patients from Wuhan, CN \\
10 & symptom\_1 & One of the symptoms encountered by patients \\
11 & symptom\_2 & One of the symptoms encountered by patients \\
12 & symptom\_3 & One of the symptoms encountered by patients \\
13 & symptom\_4 & One of the symptoms encountered by patients \\
14 & symptom\_5 & One of the symptoms encountered by patients \\
15 & symptom\_6 & One of the symptoms encountered by patients \\
\hline
\end{tabular}
\end{table}

Some findings for future research can be summarized as follows. TFSSA can be applied to various clinical applications and engineering problems, such as novel coronavirus diagnosis, dynamic electrocardiogram detection, electromyography pattern recognition, power quality diagnosis, and optimized deep neural networks. Exploring new search strategies may be a popular trend in the future. Parts of other species' behavioral mechanisms in biology, theorems and qualities in mathematics, and so on can be incorporated into EA, new search techniques created, and basic calculation formulas modified. Because combining several techniques to increase algorithmic exploration may be a trend, readers are invited to research the best combination of strategies for optimal performance. In some ways, the multi-strategy fusion algorithm outperforms the original, and the more efficient hybrid algorithm will also demonstrate the "population evolution" law. We also believe that merging notions like quantum computing with our suggested approach will improve results. In terms of comprehensive ability, while improving the convergence ability of the algorithm, we should also consider reducing the system overhead of the algorithm, which is a direction that needs further research in the future. EA generally have prominent biological and social behavior characteristics and weak mathematical theoretical support in theoretical analysis. It is necessary to strengthen the academic mathematical research of algorithms further. Furthermore, multiple advanced initialization procedures can be employed in TFSSA to improve speed.


\noindent

\bibliography{TFSSA}



\end{document}